\documentclass{article}

\usepackage[main,final]{neurips_2026}

\usepackage[utf8]{inputenc}
\usepackage[T1]{fontenc}
\usepackage{float}
\usepackage{array}
\usepackage{tabularx}
\usepackage[hidelinks]{hyperref}
\hypersetup{pdftitle={Learning to Read Out: Unembedding Dynamics in Language Model Pretraining}, pdfauthor={Matteo He, William F. Shen, Alex Iacob, Andrej Jovanovic, Xinchi Qiu, Nicholas D. Lane}}
\usepackage{url}
\usepackage{booktabs}
\usepackage{longtable}
\usepackage{amsfonts}
\usepackage{amsmath}
\usepackage{amssymb}
\usepackage{microtype}
\usepackage{xcolor}
\usepackage{graphicx}
\graphicspath{{figures/}}
\usepackage{tikz}
\usetikzlibrary{arrows.meta,positioning}
\usepackage{enumitem}
\usepackage{placeins}
\usepackage{titletoc}
\usepackage[capitalise,nameinlink,noabbrev]{cleveref}

\newcommand{\WU}{W_U}
\newcommand{\WE}{W_E}
\newcommand{\R}{\mathbb{R}}
\newcommand{\dmodel}{d_{\mathrm{model}}}
\newcommand{\dsae}{d_{\mathrm{SAE}}}

\title{
  Learning to Read Out: Unembedding Dynamics in Language Model Pretraining
}

\author{
  \textbf{Matteo He$^{1}$ \quad William F. Shen$^{1}$ \quad Alex Iacob$^{1,2}$} \\[4pt]
  \textbf{Andrej Jovanovic$^{1}$ \quad Xinchi Qiu$^{1,3}$ \quad Nicholas D. Lane$^{1,2}$}
}

\begin{document}
\maketitle
\makeatletter
\begingroup
\renewcommand{\thefootnote}{}
\long\def\@makefntext#1{\parindent 1em\noindent\hb@xt@1.8em{\hss$\m@th^{1}$}#1}
\footnotetext{Department of Computer Science and Technology, University of Cambridge.
  $^{2}$Flower Labs. $^{3}$Isomorphic Labs.\\
  Correspondence: \href{mailto:mh2274@cantab.ac.uk}{\texttt{mh2274@cantab.ac.uk}}.
  Code: \url{https://github.com/hematteo/learning-to-read-out}.\\
  Dictionaries and checkpoints: \url{https://huggingface.co/hematteo}.}
\endgroup
\makeatother
\begin{abstract}
When does a language model acquire a capability? Developmental studies usually answer this by tracking when models start producing the correct token. But that prediction passes through the unembedding matrix, or readout, which is itself still being learned during pretraining. We show that a model's hidden states can encode task information before its own readout expresses it, so timelines read from outputs conflate what the model encodes with what its readout expresses. To track the readout directly, we introduce parameter-trajectory crosscoding, which fits one sparse dictionary to every token's readout row at every checkpoint. Unlike crosscoders on activations, it needs only checkpoint weights. We use it to locate a narrow early window, 0.4--1.1\% of the way through Pythia's pretraining, in which the readout reorganizes and WordNet word categories become decodable from its rows. By then, linear probes on the hidden states already decode task information that the model's own outputs do not yet show. To test whether such gaps depend on how fast the readout learns, we pretrain small models from scratch that differ only in the readout's learning rate. We find that raising it makes the readout develop earlier and the gap close sooner, with little effect on final loss. Together, these results imply that early in training, where many developmental transitions are dated, a behavioral date can reflect when the readout learns to express task information and not when the model learns it.
\end{abstract}

\section{Introduction}
\label{sec:introduction}

Developmental studies of language models ask when a model learns a capability, and they often date it through the model's outputs. Surprisal curves across checkpoints date when individual words are acquired \citep{chang2022word}, and translation accuracy dates a shift from copying to translation \citep{korner2026copy}. The same study traces this shift across layers with the logit lens \citep{nostalgebraist2020}, which reads hidden states through the unembedding.

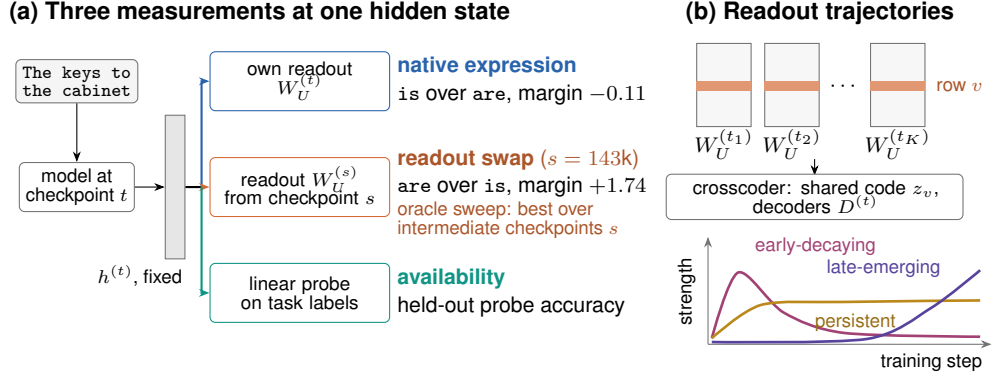
\begin{figure}[t]
\centering
\definecolor{ovnative}{HTML}{2863AB}
\definecolor{ovswap}{HTML}{D46A34}
\definecolor{ovprobe}{HTML}{0F9485}
\definecolor{ovearly}{HTML}{A14174}
\definecolor{ovpersist}{HTML}{BB881A}
\definecolor{ovlate}{HTML}{57429C}
\begin{tikzpicture}[
    font=\sffamily\footnotesize,
    >={Stealth[length=4pt,width=3pt]},
    box/.style={draw=black!55, rounded corners=2pt, align=center, inner sep=2.5pt, fill=white},
    mbox/.style={box, text width=2.2cm, font=\sffamily\scriptsize, minimum height=0.78cm},
    route/.style={->, line width=0.8pt},
    note/.style={font=\sffamily\footnotesize, align=left, anchor=west, inner sep=1pt},
    small/.style={font=\sffamily\scriptsize, align=left, anchor=west, inner sep=1pt},
    head/.style={font=\sffamily\small\bfseries, anchor=west, inner sep=0pt},
  ]
  \node[head] at (0,2.35) {(a) Three measurements at one hidden state};
  \node[box, fill=black!4, font=\scriptsize] (prompt) at (0.9,1.45) {\ttfamily The keys to\\[-1pt]\ttfamily the cabinet};
  \node[box, font=\sffamily\scriptsize] (model) at (0.9,0.05) {model at\\[-1pt] checkpoint $t$};
  \draw[->] (prompt) -- (model);
  \node[draw=black!55, fill=black!10, minimum width=0.26cm, minimum height=1.9cm, inner sep=0pt] (h) at (2.2,0.05) {};
  \node[font=\sffamily\scriptsize, align=right, anchor=north east, inner sep=1pt] at ([yshift=-1pt]h.south east) {$h^{(t)}$, fixed};
  \draw[->] (model) -- (h);
  \coordinate (fork) at (2.55,0.05);
  \draw[line width=0.8pt] (h.east) -- (fork);
  \node[mbox, draw=ovnative] (own) at (3.85,1.45) {own readout\\[-1pt] $W_U^{(t)}$};
  \draw[route, ovnative] (fork) |- (own.west);
  \node[note, text=ovnative] at (5.1,1.64) {\textbf{native expression}};
  \node[note] at (5.1,1.27) {\texttt{is} over \texttt{are}, margin $-0.11$};
  \node[mbox, draw=ovswap] (swap) at (3.85,0.05) {readout $W_U^{(s)}$\\[-1pt] from checkpoint $s$};
  \draw[route, ovswap] (fork) -- (swap.west);
  \node[note, text=ovswap!80!black] at (5.1,0.42) {\textbf{readout swap} ($s=143$k)};
  \node[note] at (5.1,0.05) {\texttt{are} over \texttt{is}, margin $+1.74$};
  \node[small, text=ovswap!80!black] at (5.1,-0.38) {oracle sweep: best over\\[-1pt] intermediate checkpoints $s$};
  \node[mbox, draw=ovprobe] (probe) at (3.85,-1.35) {linear probe\\[-1pt] on task labels};
  \draw[route, ovprobe] (fork) |- (probe.west);
  \node[note, text=ovprobe] at (5.1,-1.16) {\textbf{availability}};
  \node[note] at (5.1,-1.53) {held-out probe accuracy};
  \begin{scope}[xshift=8.95cm]
    \node[head] at (0,2.35) {(b) Readout trajectories};
    \foreach \x/\lab in {0.15/t_1,1.05/t_2,2.45/t_K} {
      \draw[black!55, fill=black!3] (\x,1.95) rectangle ++(0.72,-1.1);
      \fill[ovswap!70] (\x,1.45) rectangle ++(0.72,-0.13);
      \node[font=\footnotesize] at (\x+0.36,0.62) {$W_U^{(\lab)}$};
    }
    \node at (2.11,1.38) {$\cdots$};
    \node[font=\sffamily\scriptsize, anchor=west, text=ovswap!80!black] at (3.2,1.39) {row $v$};
    \draw[->] (1.71,0.42) -- (1.71,0.2);
    \node[box, font=\sffamily\scriptsize, text width=3.75cm] (cc) at (1.71,-0.08) {crosscoder: shared code $z_v$,\\[-1pt] decoders $D^{(t)}$};
    \begin{scope}[shift={(0.3,-2.03)}]
      \draw[->, black!55] (0,0) -- (3.75,0) node[font=\sffamily\scriptsize, anchor=north east, black] {training step};
      \draw[->, black!55] (0,0) -- (0,1.42);
      \node[font=\sffamily\scriptsize, anchor=south, inner sep=1pt, rotate=90] at (-0.12,0.7) {strength};
      \draw[line width=1pt, ovearly] plot[smooth] coordinates {(0.05,0.08) (0.4,0.95) (1.0,0.42) (1.8,0.15) (3.6,0.1)};
      \draw[line width=1pt, ovpersist] plot[smooth] coordinates {(0.05,0.08) (0.7,0.52) (1.8,0.56) (3.6,0.58)};
      \draw[line width=1pt, ovlate] plot[smooth] coordinates {(0.05,0.03) (2.1,0.07) (2.95,0.48) (3.6,0.98)};
      \node[font=\sffamily\scriptsize, text=ovearly, anchor=west] at (0.5,1.3) {early-decaying};
      \node[font=\sffamily\scriptsize, text=ovpersist!80!black, anchor=north] at (1.95,0.53) {persistent};
      \node[font=\sffamily\scriptsize, text=ovlate, anchor=south east] at (3.2,0.76) {late-emerging};
    \end{scope}
  \end{scope}
\end{tikzpicture}
\caption{Overview. (a) The same Pythia-1B step-1000 hidden state prefers \texttt{is} under its own readout and \texttt{are} under the final readout, while a linear probe measures what the state encodes without passing through any readout. (b) Parameter-trajectory crosscoding fits one sparse dictionary to the trajectories of all readout rows. Feature strengths are schematic and use the profile colors of \cref{fig:main-selected-normalized-trajectories}.}
\label{fig:overview}
\end{figure}

All of these measurements read the model through its \emph{readout}, the unembedding matrix that maps the last-layer hidden state to vocabulary \emph{logits}. The readout is itself learned during pretraining. For the prompt \texttt{The keys to the cabinet}, an early hidden state of Pythia-1B \citep{biderman2023pythia} prefers the incorrect \texttt{is} under its own readout but the correct \texttt{are} under the final checkpoint's readout, although the hidden state is the same (\cref{fig:overview}a).

The source of this discrepancy is the logit computation. For a token $v$ with readout row $w_v$ and last-layer hidden state $h$, its logit is $\ell_v=h\cdot w_v$, and between two checkpoints
\begin{equation}
\Delta\ell_v
=\Delta h\cdot w_v+h\cdot\Delta w_v+\Delta h\cdot\Delta w_v.
\label{eq:logit-change}
\end{equation}
A logit change can thus come from what the hidden state encodes, from the readout, or from the two changing together, and the logits alone cannot tell these apart. Checkpoint comparisons with the logit lens \citep{wang2024grokked,korner2026copy}, vocabulary projections \citep{wiegreffe2025answer}, or attention head scores computed through the readout \citep{tigges2024llm} share this ambiguity.

The information a task depends on, such as subject number when choosing \texttt{are} over \texttt{is}, is \emph{available} once a linear probe can decode it from the hidden state, and \emph{natively expressed} once the model's own readout prefers the correct answer. Probing studies measure the first \citep{tenney2019bert}, and behavioral studies measure the second. Prior work has found related gaps between what a model encodes and what it outputs. Trained models can encode more than their outputs express \citep{orgad2025llms,gekhman2025inside}, and during training, internal structure can form before the behavior it supports appears in the outputs \citep{chen2024sudden,nanda2023progress}.

We show that the readout has its own developmental timeline, which shapes when available information becomes natively expressed. This matters most early in training, where developmental transitions are dated \citep{olsson2022induction,chen2024sudden,korner2026copy}, and where the readout reorganizes, consistent with the early organization of vocabulary embeddings \citep{papadimitriou2025vocabulary}.

To track the readout itself, we introduce \emph{parameter-trajectory crosscoding}, a crosscoder fit to readout weights rather than activations. Like a sparse autoencoder \citep{bricken2023monosemanticity,huben2024sae}, it learns a dictionary of directions, called features, and rebuilds every token's row at every checkpoint from a small fraction of them (\cref{fig:overview}b). We also decode fixed hidden states with the readouts of other checkpoints (\cref{fig:overview}a).

We make the following contributions:
\begin{enumerate}[leftmargin=1.5em,itemsep=0.15em,topsep=0.25em]
\item \textbf{Problem.} We formalize a confound in behavioral dating that also affects evaluations of early checkpoints, with diagnostics that separate hidden state and readout contributions (\cref{sec:method-diagnostics}).
\item \textbf{Method.} We introduce parameter-trajectory crosscoding, which decomposes the readout's development into sparse features that can be followed across checkpoints. Unlike crosscoders on activations \citep{bayazit2025crosscoding,ge2026crosscoder}, it needs only checkpoint weights and covers the full vocabulary (\cref{sec:method-crosscoder}).
\item \textbf{Readout dynamics.} We identify a narrow early window, 0.4--1.1\% of the way through training, in which the Pythia readout reorganizes and WordNet categories become decodable from its rows (\cref{sec:readout-lifecycle,sec:vocabulary-family-readout}).
\item \textbf{Availability and expression.} Hidden states of Pythia-1B and 6.9B encode subject number before the model's own readout or any intermediate readout expresses it (\cref{sec:readout-reorganization-decodes}).
\item \textbf{Training intervention.} We show that raising only the readout's learning rate in small controlled pretraining runs makes the readout develop earlier and native expression reach availability sooner, with little effect on final loss. Early comparisons of training recipes can therefore reflect how fast the readout develops (\cref{sec:recipe-control}).
\end{enumerate}

\section{Method}
\label{sec:method}

We track the readout's development with parameter-trajectory crosscoding (\cref{sec:method-crosscoder,sec:method-measurements}) and separate hidden state and readout contributions with probes and readout swaps (\cref{sec:method-diagnostics}).

\subsection{Parameter-Trajectory Crosscoding}
\label{sec:method-crosscoder}

Dating the readout's development requires knowing which groups of tokens change, along which directions, and when. As training proceeds, each next-token loss pulls the target's readout row toward the hidden state and pushes every other token's row away from it, so the rows of many tokens drift along shared directions \citep{gao2019representation,bis2021too}. We describe these shared motions with a sparse dictionary, which rebuilds each token's row at every checkpoint from a small fraction of its learned directions, called features. Each feature is active for only some tokens, so unlike a principal component it identifies a group of tokens as well as a direction, and a token can belong to several groups. Sparse autoencoders fitted separately per checkpoint number their features independently, and matching them to follow a feature through training is error-prone because dictionaries trained with different seeds learn partly different features \citep{paulo2025different}. We therefore adopt the crosscoder approach \citep{lindsey2024crosscoders,ge2026crosscoder} and fit one dictionary to all checkpoints at once. Each token has one code per feature that is shared across checkpoints, and each feature has one decoder direction per checkpoint. A feature's direction, norm, and active tokens can therefore change over training while its index stays fixed. We fit the dictionary to readout parameters rather than activations, so it needs only checkpoint weights, without labels or an activation corpus.

We center and scale each checkpoint's rows, so decoder norms exclude the growth of the mean row and of the readout's overall scale. For normalized rows $x_v^{(t)}$, checkpoint set $\mathcal T$, and $F=\dsae$ features, the dictionary encodes a token's rows at all checkpoints into one shared code $z_v$, gates it at each checkpoint into the features active there, $a_v^{(t)}$, and decodes these with the checkpoint's own decoder into a reconstructed row $\widehat x_v^{(t)}$,
\begin{equation}
 z_v=\sum_{t\in\mathcal T}\left(x_v^{(t)}E^{(t)}+b_E^{(t)}\right),\quad
 a_{v,f}^{(t)}=z_{v,f}\,\mathbf 1\!\left\{z_{v,f}\|D_f^{(t)}\|_2>\theta_f\right\},\quad
 \widehat x_v^{(t)}=a_v^{(t)}D^{(t)}+b_D^{(t)}.
\label{eq:crosscoder}
\end{equation}
Here $E^{(t)}\in\R^{\dmodel\times F}$ and $D^{(t)}\in\R^{F\times\dmodel}$ are encoder and decoder matrices with rows $D_f^{(t)}$, and the $b$ terms are biases. The gate is the JumpReLU variant of \citet{ge2026crosscoder}, with learned thresholds $\theta_f$ \citep{rajamanoharan2024jumping}. An active feature adds $z_{v,f}D_f^{(t)}$ to the token's reconstructed row at checkpoint $t$, and the gate compares the norm of this contribution, $z_{v,f}\|D_f^{(t)}\|_2$, with $\theta_f$. Because the code is shared, a gate on the code alone would make a feature active for a token at every checkpoint or at none. Including the decoder norm lets a feature emerge or decay over training while keeping its index. The contribution norm also stays the same when a feature's codes are multiplied by a positive constant and its decoders divided by it, a rescaling that leaves every reconstruction unchanged, so the gate does not depend on this arbitrary scale.

Training fits the encoders, decoders, biases, and thresholds by gradient descent on
\begin{equation}
\begin{aligned}
 \mathcal L&=\mathcal L_{\mathrm{rec}}+\lambda_1\mathcal L_s,\qquad
 \mathcal L_{\mathrm{rec}}=\frac{1}{V|\mathcal T|}\sum\nolimits_{v,t}\bigl\|x_v^{(t)}-\widehat x_v^{(t)}\bigr\|_2^2,\\
 \mathcal L_s&=\sum\nolimits_{f=1}^{F}\bar q_f\Bigl(1+\frac{\bar q_f}{\alpha}\Bigr),\qquad
 \bar q_f=\frac{1}{V|\mathcal T|}\sum\nolimits_{v,t}\tanh\bigl(c\,a_{v,f}^{(t)}\|D_f^{(t)}\|_2\bigr),
\end{aligned}
 \label{eq:crosscoder-objective}
\end{equation}
where $V$ is the vocabulary size and $c$, $\alpha$, and $\lambda_1$ are constants (\cref{sec:app-method-details}). Here $\mathcal L_{\mathrm{rec}}$ is the reconstruction error, and $\mathcal L_s$ is the saturating firing rate penalty of \citet{ge2026crosscoder}, which encourages each row to be rebuilt from fewer features each making a substantial contribution. Each $\tanh$ term rises from 0 toward 1 as feature $f$'s contribution to token $v$'s row at checkpoint $t$ grows, so $\bar q_f$ is a soft count of how often the feature is used, its firing rate. Because $\tanh$ levels off at 1, a large contribution costs no more than a moderate one, so the penalty does not push active contributions toward zero as an $L_1$ penalty would. The factor $1+\bar q_f/\alpha$ makes each use more expensive the more often the feature fires, which discourages broadly used features. Like the gate, the penalty depends only on contribution norms, so it is unchanged by the rescaling of codes and decoders above, and the dictionary cannot lower it by shrinking codes and inflating decoders.

A token's logit is the dot product of the hidden state with its row, so each decoder direction is also a direction in hidden-state space. A hidden state's component along this direction shifts the logits of all the feature's tokens together, in proportion to the decoder norm and each token's code. For a given hidden state, the decoder norm thus sets how far the feature moves its tokens' logits at each checkpoint, which is why we track decoder norms across checkpoints and leave them unconstrained. The rescaling above can multiply all of a feature's decoder norms by the same factor without changing the fit, so the shape of a feature's decoder norm trajectory is meaningful but its height is not. The lifecycle profiles below therefore divide each trajectory by its maximum. Normalization, training details, and a diagram of the forward pass (\cref{fig:method-cross-snapshot-schematic}) are in \cref{sec:app-method-details}.

\subsection{Measuring Readout Development}
\label{sec:method-measurements}

We describe each feature by its decoder norm trajectory $\|D_f^{(t)}\|_2$ across checkpoints. To compare when features grow and decay rather than how large they are, we divide each trajectory by its maximum and classify it with fixed rules into one \emph{lifecycle profile}, such as early-decaying, persistent, or late-emerging (\cref{sec:app-lifecycle-diagnostics}).

To locate when the readout changes most, we measure six kinds of change between adjacent checkpoints, such as how much decoder norms change and how far decoder directions rotate. The \emph{reorganization score} of a window of training steps is the geometric mean of the six measures, each summed over the window and normalized, so it is high only where several kinds of change coincide. The \emph{reorganization window} is the window whose score most exceeds circular-shift nulls, which shift each measure in time independently, keeping its own profile but breaking the alignment between measures (\cref{sec:app-temporal-localization}). For the controlled runs of \cref{sec:recipe-control}, we also report two simpler statistics. The \emph{median feature peak step} is the median over features of the checkpoint at which each decoder norm peaks, and the \emph{peak turnover step} is the checkpoint at which normalized decoder norms change most from the previous checkpoint.

\subsection{Distinguishing Hidden States and Readout Expression}
\label{sec:method-diagnostics}

We separate the hidden state's contribution to a logit change from the readout's with two measurements. First, a \emph{readout swap} holds the hidden state fixed and decodes it with another checkpoint's readout. Let $\WU^{(t)}\in\R^{V\times\dmodel}$ be the readout at training checkpoint $t$. Let $h_p^{(t)}\in\R^{\dmodel}$ be the final hidden state at prompt position $p$, the residual stream vector (the sum of the embedding and all layer outputs) after the checkpoint's final normalization, immediately before the readout. Decoding this state with the readout of checkpoint $s$ gives the logits
\begin{equation}
  \ell_p^{(t,s)}=h_p^{(t)}\left(\WU^{(s)}\right)^\top.
  \label{eq:readout-swap}
\end{equation}
Native decoding uses $s=t$, and a readout swap uses $s\neq t$ while keeping the state, including its own final normalization. A swap therefore isolates the readout term of \cref{eq:logit-change} at a fixed state.

For a correct answer $y^+$ and an incorrect alternative $y^-$ (the foil), we score the margin at the final prompt position,
\begin{equation}
 m^{(t,s)}=h_p^{(t)}\cdot\bigl(\WU^{(s)}[y^+,:]-\WU^{(s)}[y^-,:]\bigr).
 \label{eq:contrast-margin}
\end{equation}
The \emph{contrast accuracy} of readout $s$ is the fraction of items with $m^{(t,s)}>0$, and \emph{native expression} $E$ is the contrast accuracy of the checkpoint's own readout, $s=t$. The \emph{oracle readout sweep} scores the readouts of all intermediate checkpoints, excluding the native readout and the final readout, whose effect on early states we report separately (\cref{sec:readout-reorganization-decodes}), and reports the highest contrast accuracy, reached by the \emph{best swept readout}. Because the best readout is chosen after scoring, the sweep is an upper bound on the contrast accuracy of any single intermediate readout.

Second, a linear probe reads the hidden state directly, without passing it through any readout. \emph{Availability} $A$ is the probe's accuracy at predicting the task label from the hidden states that the swaps decode. Comparing $A$ with $E$ separates what a state encodes from what its own readout expresses.

\section{Experimental Setup}
\label{sec:experimental-setup}

\paragraph{Models and dictionaries.}
We study readout development in Pythia-160M, 1B, and 6.9B \citep{biderman2023pythia} and in OLMo-2-7B, which differs from Pythia in architecture, tokenizer, and training recipe \citep{olmo2024olmo2}. The dictionaries for all four models cover the full vocabulary across 32 public checkpoints (\cref{tab:full-inventory}). To test how the readout's own optimization shapes its development, we train four additional 31M-parameter models on the first 10B tokens of Pythia's preshuffled training data, in Pythia's order, from a shared initialization (\cref{sec:recipe-control}). Three runs differ only in the readout's learning rate multiplier ($0.25\times$, $1\times$, $4\times$), and the fourth keeps the $1\times$ multiplier and halves the warmup from 1430 to 715 steps. Each 31M model is trained once, and three dictionaries with one shared width and sparsity weight are fitted to 16 of its checkpoints.

\begin{table}[t]
\centering
\footnotesize
\setlength{\tabcolsep}{3pt}
\caption{Task examples and item counts for subject--verb agreement (SVA), indirect object identification (IOI), and number comparison. Underlined nouns are attractors, which stand between the subject and the verb. $n_A$ counts items in the broader task suite used for availability, and $n_L$ counts the separately built localization items before the selection and scoring split. Simple SVA is the set plotted in \cref{fig:main-sva-availability-expression}a. Templates and word lists are in \cref{sec:app-task-construction}.}
\label{tab:task-protocol}
\begin{tabularx}{\linewidth}{@{}llXll@{\hspace{2\tabcolsep}}rr@{}}
\toprule
\multicolumn{2}{@{}l}{Task} & Example prompt & Correct & Foil & $n_A$ & $n_L$ \\
\midrule
SVA & simple & The keys to the \underline{cabinet} & \texttt{are} & \texttt{is} & 320 & 320 \\
 & across PP & The judge by the \underline{street} & \texttt{appears} & \texttt{appear} & 2{,}000 & -- \\
 & subject RC & The author that admires the \underline{senators} & \texttt{likes} & \texttt{like} & 1{,}200 & -- \\
 & object RC & The author that the \underline{senators} admire & \texttt{likes} & \texttt{like} & 1{,}200 & -- \\
\addlinespace
IOI & availability & Lisa and Jane went to the store. Jane waited. Lisa gave the ring to & \texttt{Jane} & \texttt{Lisa} & 2{,}000 & -- \\
 & localization & Ryan and Adam went to the store. Ryan gave the pen to & \texttt{Adam} & \texttt{Ryan} & -- & 462 \\
\addlinespace
Number & comparison & Which is greater, 64 or 84? Answer: & \texttt{84} & \texttt{64} & 2{,}000 & -- \\
 & greater-than & 84 is greater than & \texttt{64} & \texttt{84} & -- & 2{,}000 \\
\bottomrule
\end{tabularx}
\end{table}

\paragraph{Tasks and scoring.}
We compare readouts on binary tasks, each with a correct answer token and a foil (\cref{tab:task-protocol}). Subject--verb agreement (SVA) \citep{linzen2016assessing,marvin2018targeted} contrasts the agreeing verb form with the non-agreeing one, indirect object identification (IOI) \citep{wang2022interpretability} contrasts the recipient with the giver, and number comparison contrasts the correct number with the other number in the prompt. Labels are balanced by subject number, by whether the recipient is the first or second name, or by number order, and each answer is a single token that includes its leading space. The native readout and each swapped readout are scored at the final prompt position with the margin of \cref{eq:contrast-margin}. Contrast accuracy compares only the two answers and ignores every other token in the vocabulary. With balanced labels, its chance level is $50\%$.

\paragraph{Probes and controls.}
We measure availability with linear probes on the final hidden state. For simple SVA and IOI in Pythia-6.9B, the probe is logistic regression, fit at each of the 32 checkpoints (\cref{sec:readout-reorganization-decodes}). For a broader suite of 20 task families, the probe is a single direction from shrinkage linear discriminant analysis (LDA), fit on the families with a shared binary label, and we call its accuracy $A_1$ (\cref{sec:app-availability-expression}). Prompts that share a noun, a name pair, or a number pair always fall in the same cross-validation fold. We compare each probe with three controls: a probe at step 0 shows what is decodable before training, a probe on prompt tokens alone tests whether surface features suffice, and a probe on random Gaussian features tests whether the classifier fits unrelated inputs. Probes fit to shuffled labels stay below $0.56$ on the availability sets of \cref{tab:task-protocol}. \Cref{sec:app-evaluation-details} gives fit details and maps controls to confounds.

\paragraph{Sparse localization.}
To find which readout features carry a task margin, we decode step-$1000$ Pythia-1B states with the dictionary's reconstruction of the final readout, which is how a logit lens with the final readout reads an early checkpoint. The reconstructed readout is a sum of feature contributions, so each feature contributes a separate term to the margin, and a feature's attribution is the mean of this term over items. \citet{he2026prism} explain logit lens scores this way for a single readout. In this work, the features come from a dictionary fitted across all 32 checkpoints, so each feature keeps the same index throughout training. For SVA, IOI, and greater-than completion (\cref{tab:task-protocol}), we keep only answer pairs whose rows in the final readout have similar norms, $|\log(\|w_{y^+}\|_2/\|w_{y^-}\|_2)|\leq0.5$. This filter limits the effect of row norm on the margin. We split each task's items into two equal parts at random and select the eight features with the largest attribution on the first part. On the second part, we measure contrast accuracy under two edits to the reconstruction: ablation, which removes these features, and isolation, which keeps only them. We compare these edits with the full reconstruction and with the same edits applied to random features matched on decoder norm, firing rate, and attribution sign (\cref{sec:app-contrastive-localisation}).

\paragraph{Lexical organization.}
The dictionary dates when the readout reorganizes but not what its rows learn. To check this without the dictionary, we probe the readout rows directly for WordNet supersenses, broad categories such as nouns for animals \citep{miller1995wordnet,fellbaum1998wordnet}. For each of the 35 categories with at least 20 whole-word tokens, we train a one-vs-rest logistic probe on two-thirds of the centered readout rows and report its balanced accuracy (the mean recall over the two classes) on the remaining third. Three controls test whether token metadata (frequency, length, row norm, spacing, and token class) explains this accuracy. The first probes the metadata alone, the second probes matched random token sets, and the third shuffles labels within metadata bins (\cref{sec:app-vocabulary-family-probes}).

\section{Results}
\label{sec:results}
We first ask when and how the readout's structure changes (\cref{sec:readout-lifecycle,sec:vocabulary-family-readout}), then whether hidden states encode task information before the model's own readout expresses it (\cref{sec:readout-reorganization-decodes,sec:sparse-coordinate-localization}), and finally whether the readout's learning rate sets the timing of its development (\cref{sec:recipe-control}).

We first check how well the dictionaries reconstruct the readout, since the lifecycle and localization analyses read it through them. Across Pythia-160M, 1B, and 6.9B and OLMo-2-7B, the dictionaries explain $0.808$--$0.920$ of the variance in the readout trajectories. On average, each readout row uses $1.17$--$2.26\%$ of the features at each checkpoint, and no feature is dead (\cref{fig:main-multimodel-validation}). They also reconstruct Pythia-160M checkpoints held out from fitting nearly as well as the fitted ones (\cref{sec:app-heldout-checkpoints}). They nearly match the fit of sparse autoencoders fitted separately per checkpoint with far fewer active features, and unlike those autoencoders they keep one feature index across checkpoints. In Pythia, dense PCA needs far more components to reach the same fit (\cref{sec:app-instrument-validation}).

\subsection{The Readout Reorganizes Early}
\label{sec:readout-lifecycle}

\begin{figure}[t]
    \centering
    \includegraphics[width=\linewidth]{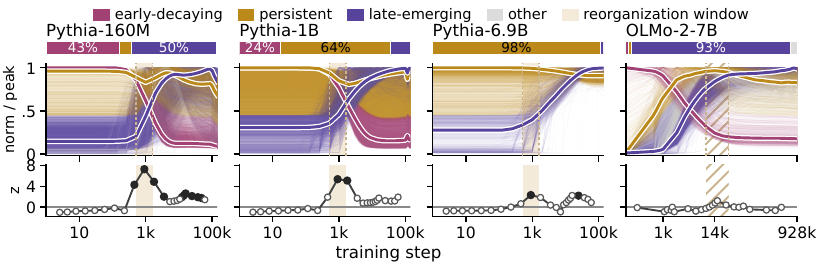}
    \caption{
        Pythia's readout reorganizes in one early window, and the mix of feature trajectories differs by model.
        Top: share of features in each lifecycle profile. Middle: each line is one active feature's decoder norm divided by its own maximum, so heights do not compare features; thick lines are profile medians. Bottom: reorganization score as a $z$-score against circular-shift nulls, with filled markers at $p\le0.05$. Bands mark the selected windows, hatched for OLMo-2-7B ($p=0.12$).
    }
    \label{fig:main-selected-normalized-trajectories}
\end{figure}

When does the readout change, and do its features share one growth pattern? In all four models, the reorganization score peaks within the first few percent of training. In Pythia, it peaks between step 512 and step 1.6k in all three models, 0.4\% to 1.1\% of the way through training, a window that contains the end of the learning rate warmup at step 1430. This window stands out against circular-shift nulls, with $p=0.005$ at 160M and 1B and $p=0.04$ at 6.9B (\cref{fig:main-selected-normalized-trajectories}, bottom; \cref{fig:app-reorganization-window-metrics,tab:app-reorganization-window-stats}). Of the 21 Pythia dictionaries we fitted, 20 select the same window and the remaining one selects the adjacent window (\cref{sec:app-lifecycle-stability}). Measures of raw readout movement that use no dictionary, such as the drift of its top subspace, also put the largest movement early, at the checkpoint pair ending at step 256 or 512, just before this window (\cref{fig:app-dense-reorg-pythia}). In OLMo-2-7B, the score peaks at a similar fraction of training, between step 9k and step 28k. This window lies 1--3\% of the way through OLMo's first training stage, the main pretraining phase before mid-training. It does not separate from the nulls ($p=0.12$, \cref{sec:app-lifecycle-diagnostics}).

Features do not share one growth pattern. We follow each feature by its decoder norm, which sets how far the feature moves its tokens' logits. In Pythia-160M and 1B, some features peak early and decay, others persist, and others emerge late, while nearly all Pythia-6.9B features persist and most OLMo-2-7B features emerge late (\cref{fig:main-selected-normalized-trajectories,fig:app-lifecycle-profile-composition}). OLMo-2-7B's readout also grows more gradually, with its total decoder norm rising through most of training. These profile shares vary with dictionary width and sparsity (\cref{sec:app-lifecycle-stability}). Early decay also depends on the checkpoints nearest initialization. A Pythia-1B dictionary fitted only from step 256 onward assigns $0.02\%$ of its features to the early-decaying profile, against $24\%$ in the main Pythia-1B dictionary (\cref{sec:app-olmo-checkpoint-window-control}).

The readout therefore changes most early on, in one window that clears the nulls in Pythia and at a similar fraction of training in OLMo-2-7B, and the mix of feature trajectories differs by model.

\subsection{Lexical Organization Develops Alongside Structural Change}
\label{sec:vocabulary-family-readout}

\begin{figure}[t]
\centering
\includegraphics[width=\linewidth]{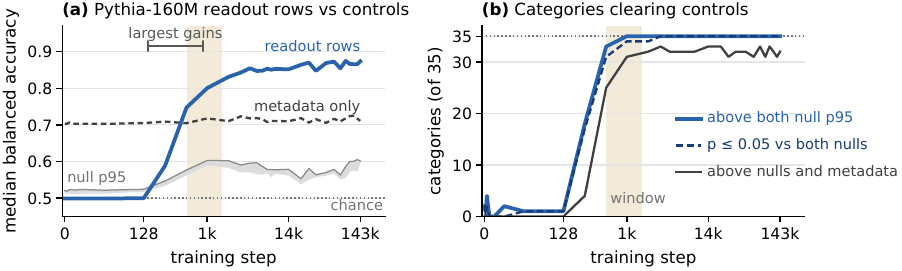}
\caption{Pythia-160M readout rows come to separate WordNet categories beyond matched controls, with the largest gains between steps $128$ and $1000$, which overlap the start of the reorganization window. (a) Median held-out balanced accuracy over the 35 categories, against a probe on token metadata alone (dashed) and the 95th percentiles (p95) of the matched-random and shuffled-label nulls (gray band). The bracket marks the largest gains and the shaded band the reorganization window. (b) Number of categories above each control.}
\label{fig:main-wordnet-matched-controls}
\end{figure}

Do the readout rows acquire recognizable lexical organization while their structure changes? In Pythia-160M, the median probe accuracy across the 35 WordNet categories rises from $0.499$ at initialization to $0.874$ at the final checkpoint. We call a category decodable once its accuracy stays at or above $\max(0.55,\mathrm{acc}_0+0.05)$ for three successive checkpoints, which counts only gains above its accuracy at initialization. All 35 categories become decodable by step $1000$, and 33 have their largest gain between steps $128$ and $1000$ (\cref{fig:main-wordnet-matched-controls}). In Pythia-1B, the median category becomes decodable at step $512$, with onsets from step $128$ to step $2000$ (\cref{sec:app-vocabulary-family-probes}).

We compare the probes with the three controls of \cref{sec:experimental-setup}. The first control, a probe on token metadata alone, predicts some categories well early in training. From step $512$ onward, median readout accuracy exceeds it. Against the second and third, by step $512$, 33 of 35 categories exceed the $95$th percentile of both the matched-random and the shuffled-label nulls, and by step $1000$ all 35 do. Splitting the tokens into training and test sets by lemma, so that variants such as \texttt{dog} and \texttt{Dog} never appear in both, leaves the early rise unchanged and lowers late accuracy, to a final median of $0.841$ with 34 of 35 categories decodable by step $1000$ (\cref{sec:app-vocabulary-family-probes}).

Lexical organization therefore emerges in the readout rows before and during the 512--1.6k window.

\subsection{Hidden State Availability Precedes Native Readout Expression}
\label{sec:readout-reorganization-decodes}

\begin{figure}[H]
\centering
\includegraphics[width=\linewidth]{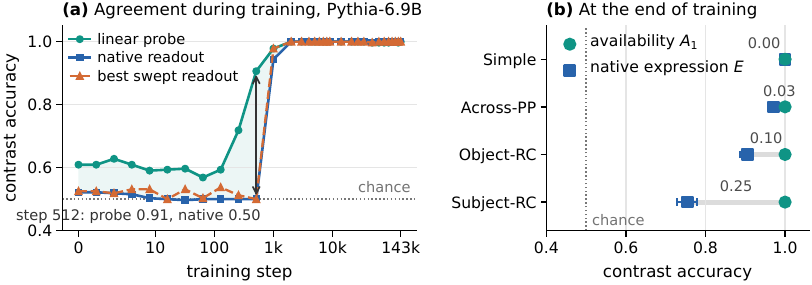}
\caption{Pythia-6.9B hidden states encode subject number before the model's own readout or any intermediate readout expresses it. (a) Accuracy on simple SVA of the logistic probe, the native readout, and the best swept readout over intermediate checkpoints; the arrow marks the gap at step 512. (b) Single-direction probe $A_1$ and native expression $E$ on the four agreement sets of \cref{tab:task-protocol} at the final checkpoint; annotations give $A_1-E$, and bars are 95\% Wilson intervals for $E$.}
\label{fig:main-sva-availability-expression}
\end{figure}

Is task information decodable from the hidden state before the model's own readout expresses it? A probe and the native readout can disagree because they read the hidden state along different directions. The probe learns its direction from the task labels, while the native readout can only use the difference between its two answer rows. If the state separates the answers along a direction that this difference does not yet point in, the information is available but not expressed. Expression begins once the hidden state or the readout changes so that the two directions line up. We track the probe, the native readout, and the oracle readout sweep on the same items at every checkpoint. On Pythia-6.9B agreement, the probe approaches ceiling by step 512 while native expression and even the best swept readout are still at chance (\cref{fig:main-sva-availability-expression}a). Probes on prompt tokens and on random Gaussian features stay near chance, so neither surface tokens nor classifier fitting explains the probe's accuracy (\cref{fig:app-contrastive-readout-lag}). The same early gap appears in every subject--verb agreement set of the broader 20-family suite whose probe starts near chance, in both Pythia-1B and Pythia-6.9B (\cref{sec:app-availability-expression}).

Can a later readout decode the current hidden state better than its own? For next-token prediction, it can. On Wikipedia text in nine languages, the step-1000 readout lowers the aggregate next-token loss of Pythia-1B step-256 and step-512 states by $1.50$ and $0.15$ nats, and every language prefers an intermediate readout, even after aligning the two readouts' frames (\cref{fig:app-readout-swap-nll-kl-grid,fig:app-aligned-readout-swap,tab:app-readout-swap-languages}). Early in training, the readout therefore lags behind the hidden states. The final readout also raises early task accuracy, from $0.800$ to $0.925$ on Pythia-1B step-1000 SVA states, as in \cref{fig:overview}a, but this gain depends on its frame and falls to $0.566$ after an orthogonal Procrustes alignment \citep{schonemann1966procrustes}, which rotates the final readout to best match the native one (\cref{sec:app-terminal-swap-alignment}).

Availability therefore precedes native expression. \Cref{sec:sparse-coordinate-localization} asks which readout directions carry these margins, and \cref{sec:recipe-control} tests whether the readout's own optimization sets their timing.

\subsection{A Few Readout Directions Carry the Agreement Margin}
\label{sec:sparse-coordinate-localization}

Which parts of the readout carry a task margin? In the localization setup of \cref{sec:experimental-setup}, eight features carry the SVA margin. Ablating them lowers reconstructed SVA accuracy from $0.900$ to $0.487$, keeping only them gives $0.938$, and ablating matched random features leaves $0.931$. These interventions are causal evidence that the reconstructed final readout's SVA margin on step-1000 states runs through these eight directions. \Cref{sec:app-contrastive-localisation} repeats the analysis for IOI and greater-than completion. This analysis decomposes the logit lens margins of \cref{sec:readout-reorganization-decodes} into readout features.

\subsection{Readout Learning Rate Shifts Readout Development}
\label{sec:recipe-control}

\begin{figure}[t]
\centering
\includegraphics[width=\linewidth]{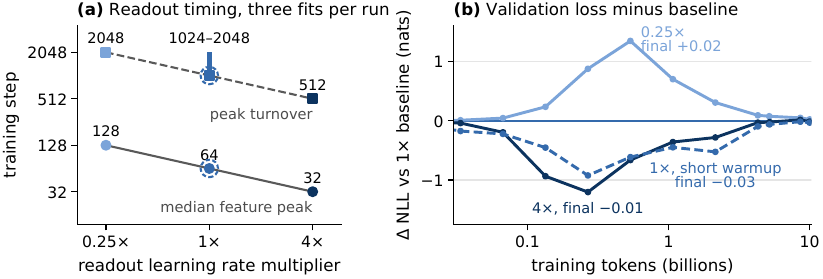}
\caption{A faster readout develops earlier and lowers early validation loss, while final losses agree within $0.06$ nats. (a) Median feature peak step and peak turnover step against the readout learning rate multiplier; bars span the three fits where they differ, and dashed circles mark the shorter warmup. (b) Validation loss of each 31M run minus that of the baseline run.}
\label{fig:main-recipe-dose-response}
\end{figure}

Early in training, the readout lags behind the hidden states, and in Pythia its reorganization window contains the end of warmup. Does the readout's timing come from how fast the readout itself learns, or from its warmup? We find that changing the readout's learning rate shifts when its features peak, while halving the warmup leaves this timing unchanged on our checkpoint grid. In the 31M runs of \cref{sec:experimental-setup}, raising the multiplier moves the median feature peak step from $128$ to $64$ to $32$ in every fit (\cref{fig:main-recipe-dose-response}), one checkpoint per fourfold increase. Halving the warmup leaves the median feature peak at its baseline value in every fit.

A faster readout also lets native expression reach availability sooner. In these runs, availability is a probe's accuracy on the frequency class of the next token, and native expression is the model's accuracy on the same classes. Along the multiplier axis, a slower readout gives a larger early gap between this probe and native expression, and the gaps close by step $1024$ in every run except $0.25\times$, which closes by step $2048$. Readout swaps show the same ordering. In the $0.25\times$ run, step-$256$ hidden states are decoded best by the step-$512$ readout, which lowers their next-token loss by $0.91$ nats and keeps this gain after orthogonal Procrustes alignment, whereas in the $4\times$ run each checkpoint's own readout is best (\cref{sec:app-recipe-control-concordance}).

Does developing earlier leave the model better off at the end of training? It lowers early loss but not final loss. At $0.27$B tokens, the validation loss of the $4\times$ run is $1.20$ nats below the baseline and $2.08$ nats below the $0.25\times$ run, whereas final validation losses agree within $0.06$ nats (\cref{fig:main-recipe-dose-response}b).

We therefore identify the readout's learning rate as one factor that shifts when the readout develops and when native expression reaches availability. Unlike the fixed-state swaps of \cref{sec:readout-reorganization-decodes}, this intervention acts during training and changes the hidden states as well as the readout.

\section{Related Work}
\label{sec:related-work}

\paragraph{Reading development through the readout.}
Developmental analyses track linguistic representations, circuits, and concepts across checkpoints \citep{olsson2022induction,nanda2023progress,inaba2025how,xu2024tracking,sawmya2025birth}, and the logit lens and tuned lens project hidden states into vocabulary scores \citep{nostalgebraist2020,belrose2023tuned}.
Analyses that read development through the outputs depend on a readout that is itself still being learned.
We add the readout to these comparisons, tracking its rows with shared sparse features and testing how changing it alters what fixed hidden states express (\cref{tab:method-comparison} compares these approaches).

\paragraph{Weight-space analyses.}
Analyses of output embeddings study weight tying \citep{inan2016tying,press2017output}, the softmax bottleneck \citep{yang2018softmax}, and embedding geometry and its evolution \citep{gao2019representation,machina2024anisotropy}.
Closest to our lexical analysis, \citet{papadimitriou2025vocabulary} correlate the geometry of input and output embeddings with semantic, syntactic, and frequency structure over training.
Parameter decomposition and weight-sparse models analyze mechanisms in weights, typically at one model state \citep{braun2025parameter,gao2025weightsparse}. As an alternative to reading logit lens scores in token units, \citet{he2026prism} explain them as sums of sparse feature contributions, decomposing a single readout from its weights alone.

\paragraph{Shared dictionaries over checkpoints.}
Sparse autoencoders decompose activations into sparse features \citep{bricken2023monosemanticity,huben2024sae,templeton2024scaling,gao2025scaling,rajamanoharan2024gated}, and crosscoders share feature indices across layers or models \citep{lindsey2024crosscoders}.
The closest methods are the checkpoint crosscoders of \citet{bayazit2025crosscoding} and \citet{ge2026crosscoder}, which learn shared features over activations sampled from a text corpus (400M and 800M tokens) and relate their evolution to behavior through interventions or attribution.
We adopt their shared codes and checkpoint-specific decoders, with the gate and sparsity penalty of \citet{ge2026crosscoder}, both of which use decoder norms, and apply them to token-aligned readout parameters.
Fitting then needs only checkpoint weights and covers every readout row rather than only the tokens a corpus elicits, while the behavioral tests still need text, forward passes, and task labels.

\section{Discussion}
\label{sec:discussion}

Taking our findings together, we argue that the readout is part of the measurement in any analysis that reads development off logits. A rise in output accuracy or in a logit margin can come from the hidden state or from the readout, even for continuous scores that avoid metric-induced emergence \citep{schaeffer2023mirage}. Reporting native expression, a probe on the same task with its initialization baseline, and a readout sweep on the same states would help distinguish the two.

Readout change concentrates in the same early period in which many developmental transitions are dated \citep{chen2024sudden}. The induction head phase change, for example, occurs roughly 1--2\% of the way through training in large models \citep{olsson2022induction}, close to the 0.4--1.1\% window in Pythia. A behavioral date in this period can therefore reflect readout development as well as hidden state development. What drives the different feature trajectories within this period remains open. Language models first predict from unigram frequencies and then loosely from bigram probabilities, before converging on richer predictions \citep{chang2022word}, and early-decaying features may record the first of these stages, with late-emerging features taking over as context use develops.

The same confound affects evaluations of early checkpoints and of small, briefly trained models, which inform forecasts of how larger runs will perform \citep{openai2023gpt4,snell2024predicting}. Such forecasts already contend with capabilities that appear abruptly with scale \citep{wei2022emergent,ganguli2022predictability}. While the readout still lags, outputs can underreport what the hidden states already encode, whereas a probe alone would overpredict expression. Internal progress measures can anticipate a behavioral transition \citep{nanda2023progress}. Testing whether native expression, probes, and readout sweeps sharpen such forecasts, also for safety-relevant capabilities, is a natural next step.

The lag is largest early in training. Once the readout has reorganized, each hidden state is decoded best by its own readout or a neighbor, and native expression catches up with availability on several tasks, so most implications concern analyses of early checkpoints. In the 31M runs, the readout's learning rate changes early loss and when a distinction becomes expressible, but leaves final losses within $0.06$ nats (\cref{sec:recipe-control}). Recipe comparisons made on partial runs can therefore reflect how fast the readout develops rather than where training ends. Overall, learning a representation and learning to express it can happen at different times in pretraining, and measuring both separates what a model encodes from what its logits express.

\paragraph{Limitations.}\label{sec:limitations}
First, limited compute restricted the evidence that the readout's learning rate shifts its timing to one 31M setting, with one run per condition stopped at 10B tokens. The readout holds about 42\% of the parameters there, against about 10\% in Pythia-1B, so larger models remain untested. Second, availability is measured with probes on two-token contrasts, so the gap describes when a binary distinction becomes expressible, not when an open-ended capability appears. \Cref{sec:app-result-scope} covers further limits, such as checkpoint density and dictionary settings.

\paragraph{Future work.}
We suggest three useful directions. First, recipe comparisons on early checkpoints can favor a faster readout even when final losses agree (\cref{sec:recipe-control}). Checking whether recipe rankings made with probes or swapped readouts match the final rankings would test a fix. Second, extending a pretrained model's vocabulary adds untrained readout rows, so training these rows with their own learning rate may speed up adaptation. Third, measuring weight drift in every layer (\cref{fig:app-dense-reorg-pythia}) would show whether each component, like the readout, settles on its own schedule.

\clearpage
\begin{ack}
\paragraph{Funding and competing interests.}
The authors report no third-party funding or in-kind support for this work and no additional competing interests beyond the affiliations listed above.
\end{ack}

\bibliographystyle{plainnat}
\bibliography{refs}
\appendix
\crefalias{section}{appendix}
\crefalias{subsection}{subappendix}
\crefalias{subsubsection}{subsubappendix}
\numberwithin{figure}{section}
\numberwithin{table}{section}
\renewcommand{\theHfigure}{\thesection.\arabic{figure}}
\renewcommand{\theHtable}{\thesection.\arabic{table}}

\clearpage
\startcontents[appendix]
\section*{Appendix Contents}
\begingroup
\footnotesize
\setlength{\parskip}{0pt}
\setcounter{tocdepth}{2}
\printcontents[appendix]{}{1}{}
\endgroup

\begingroup
\raggedbottom
\setlength{\textfloatsep}{8pt plus 2pt minus 2pt}
\setlength{\floatsep}{8pt plus 2pt minus 2pt}
\setlength{\intextsep}{8pt plus 2pt minus 2pt}
\setlength{\abovecaptionskip}{3pt}
\setlength{\belowcaptionskip}{0pt}
\setlength{\LTpre}{3pt}
\setlength{\LTpost}{5pt}
\section{Dictionary Normalization, Objective, and Evaluation Details}
\label{sec:app-method-details}

\subsection{Normalization and Reconstruction}
\label{sec:app-method-normalization}

At checkpoint $t$, we subtract the mean vocabulary row and scale the centered entries by their root mean square:
\begin{equation}
\begin{aligned}
 \mu^{(t)}&=\frac1V\sum_{v=1}^{V}\WU^{(t)}[v,:],\\
 s^{(t)}&=\left[\frac{1}{V\dmodel}\sum_{v=1}^{V}
 \left\|\WU^{(t)}[v,:]-\mu^{(t)}\right\|_2^2\right]^{1/2},\qquad
 x_v^{(t)}=\frac{\WU^{(t)}[v,:]-\mu^{(t)}}{s^{(t)}}.
\end{aligned}
\end{equation}
The quantity inside the square root is floored at $10^{-8}$. Apart from this floor, the normalized matrix has zero mean row and average squared row norm $\dmodel$. Centering removes a vocabulary-wide offset, and scaling keeps a checkpoint with larger entries from dominating the loss. \Cref{sec:app-mean-direction-preprocessing} measures these global effects.

Dictionary inputs keep every output row, including padding and special-token rows. GPT-NeoX uses \texttt{<|endoftext|>} (id 0) as both BOS and EOS and \texttt{<|padding|>} (id 1) as the pad token, and the probe predicates exclude both. The mean and scale are stored with the dictionary, so raw $\WU$ reconstructions can be recovered. Training uses no external token frequency table. The configuration name \texttt{frequency\_scale} is historical, and its value, $0.01$, is $\alpha$ in \cref{eq:crosscoder-objective}. Because the dictionaries are fit to every vocabulary row, EV is an in-sample metric and does not measure generalization to unseen rows. We therefore report it with mean $L_0$ (active features per row and checkpoint pair), the dead feature rate, per-checkpoint fidelity, stability checks, and baselines.

Given a normalized reconstruction $\widehat x_v^{(t)}$, its original-space reconstruction is
\[
 \widehat\WU^{(t)}[v,:]=s^{(t)}\widehat x_v^{(t)}+\mu^{(t)}.
\]
Feature $f$ therefore contributes $s^{(t)}a_{v,f}^{(t)}D_f^{(t)}$ in the original readout space. Readout interventions keep the stored mean and decoder bias unless stated otherwise.

The final normalization and the readout share one exact symmetry. For the learned gain $\gamma$, the LayerNorm bias $\beta$, and any invertible diagonal matrix $\Lambda$, replacing $(\gamma,\beta,\WU)$ with $(\Lambda\gamma,\Lambda\beta,\WU\Lambda^{-1})$ leaves every logit unchanged. Native expression, next-token loss, and the standardized logistic probes of \cref{sec:app-evaluation-details} are invariant to it. Readout rows, their norms and directions, the dictionary features, and readout swaps between checkpoints are not, because they depend on how each coordinate's scale is split between the gain and the readout. We measure readout geometry in the split that each released checkpoint stores, and centering and scaling remove only a vocabulary-wide offset and one global scale. \Cref{sec:app-recipe-control-gauge} discusses the scalar case in the 31M runs.

\subsection{Gate and Training Objective}
\label{sec:app-method-objective}

\begin{figure}[H]
\centering
\includegraphics[width=\linewidth]{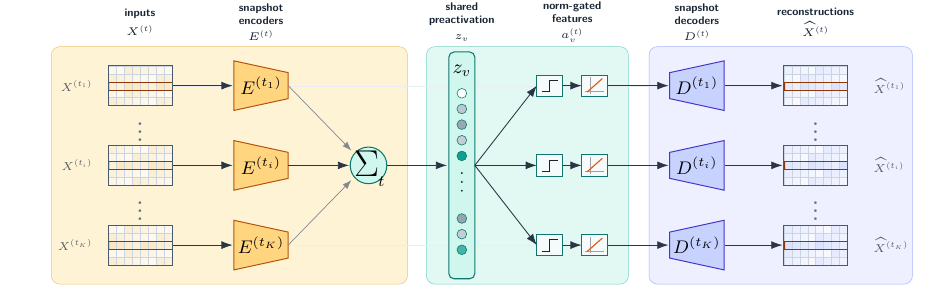}
\caption{The parameter-trajectory crosscoding forward pass (\cref{eq:crosscoder}) retains one feature index across checkpoints while allowing its contribution to change through training. One vocabulary row is collected across centered and scaled checkpoints and encoded into a shared preactivation, and a checkpoint-specific gate and decoder reconstruct each view.}
\label{fig:method-cross-snapshot-schematic}
\end{figure}

Training minimizes the objective of \cref{eq:crosscoder-objective}. The reconstruction term sums squared errors over hidden dimensions and averages over vocabulary rows and checkpoints, without dividing by $\dmodel$. Minibatch training uses the same reductions with the batch size in place of $V$ (\texttt{src/readout/crosscoder/wu\_adapter.py}). The separately logged per-entry reconstruction MSE also averages over hidden dimensions, so it is $\dmodel$ times smaller and cannot replace this term without rescaling $\lambda_1$.

The constant $c$ sets the saturation scale of the $\tanh$, $\alpha$ the extra cost of broad firing, and $\lambda_1$ the tradeoff with reconstruction. The effective $\lambda_1$ is warmed up linearly from zero over the first $5\%$ or $10\%$ of optimizer steps (\cref{tab:repro-dictionary-hparams}). The gate and the penalty are unchanged by the positive featurewise rescaling $z_{v,f}\mapsto z_{v,f}/r$ and $D_f^{(t)}\mapsto rD_f^{(t)}$ of \cref{sec:method-crosscoder}, but this invariance does not make the sparse factorization unique.

\subsection{Behavioral Probe Fitting and Reporting}
\label{sec:app-evaluation-details}

\paragraph{Hidden-state convention.}
The hidden state $h_p^{(t)}$ of \cref{sec:method-diagnostics} is the output of the final normalization, captured with a forward pre-hook on Pythia's \texttt{embed\_out} module. Native and swapped logits are computed directly as $h_p^{(t)}(\WU^{(s)})^\top$, without normalizing a second time.

\paragraph{Logistic probes for the 32-checkpoint SVA and IOI trajectories.}
The SVA target is subject number, and the role-balanced IOI target is whether the recipient is the first or second name. In each of five folds, a standardization fit on the training examples is followed by $L_2$-regularized logistic regression ($C=1$, \texttt{lbfgs}, at most $10{,}000$ iterations), with no tuning on test folds. We report mean held-out accuracy; fold standard deviations describe split variation, not independent training runs. Folds group SVA items by head noun lemma and IOI items by unordered name pair, so individual names can recur in other pairs across folds. The unbalanced recipient identity diagnostic is a separate multiclass probe, not the role-balanced binary result.

The prompt-token controls fit the same classifier to binary indicators or counts of prompt tokens. The random-feature control fits it to a fixed Gaussian matrix of the same shape as the hidden states, on the same folds. These controls test surface token statistics and fitting to unrelated features.

\paragraph{Broader binary task suite.}
The broader task suite reports a single shrinkage LDA direction, $A_1$, and a logistic probe, $A_{\mathrm{full}}$, both with cross-validation grouped by the family's lemma or pair key. Number pair groups keep both orderings of a pair together, but individual numbers can recur in other pairs. The shrinkage coefficient, fold count, and selection rule of the shrinkage LDA probe were not recovered from the archived runs. Both classifiers project onto one direction, but their supervision and optimization differ from the model's pretrained answer row difference, so they measure task-supervised decodability.

\paragraph{Readout sweeps and candidate scores.}
For the 32-checkpoint SVA and IOI trajectories and \cref{fig:main-sva-availability-expression}, the sweep $B_{\mathrm{env}}$ picks one readout per task and hidden state checkpoint after scoring all items, so it is a post hoc upper envelope over the evaluated readouts. In the broader task suite, the sweep $B_{\mathrm{ho}}$ picks the readout on a random half of the items and reports its accuracy on the other half (\cref{sec:app-availability-expression}). Neither sweep picks a readout per item. Both exclude the native and terminal readouts, while the corpus negative log-likelihood (NLL) grid uses the full readout schedule. Contrast accuracy counts strictly positive margins. Full-vocabulary rank-1 accuracy, whether the gold answer is the argmax over all tokens, is reported separately.

Swap effects cannot come from drift in the mean readout row. Adding a vector $u$ to every row of $\WU^{(s)}$ shifts every logit by $h_p^{(t)}\cdot u$, which cancels in each contrast margin and in the softmax, so contrast accuracy and next-token NLL are unchanged. Multiplying $\WU^{(s)}$ by a positive scalar scales every margin, which leaves contrast accuracy unchanged but changes NLL.

\paragraph{Tokenizer filters and localization splits.}
Both answers are single tokens with a leading space (templates in \cref{sec:app-task-construction}). Only the localization items pass the terminal row norm filter of \cref{sec:experimental-setup}, which bounds norm differences but does not equalize norms or remove directional effects. Selecting features on half the items and scoring the other half leaves 160 SVA, 1000 greater-than completion, 231 IOI, and 13 relational-fact scoring items (\cref{tab:task-protocol,tab:app-task-construction}).

\paragraph{31M recipe probe.}
The availability probe of the 31M runs predicts one of four next-token frequency classes, each holding about equal token mass in the evaluation slice. It uses standardized logistic regression ($C=1$, \texttt{lbfgs}, at most $10{,}000$ iterations) over five shuffled folds of sampled positions. The folds are stratified by frequency class and are not grouped by document, so positions from one $2048$-token block can fall in both training and test folds. Native expression maps the model's full-vocabulary argmax to its frequency class. Label-shuffle and random-label controls use the same folds, and the readout sweep covers each condition's reported checkpoints (\cref{sec:app-recipe-control}).

\begin{table}[H]
\centering\small
\caption{Controls and the comparison each supports. Each control addresses one alternative explanation.}
\label{tab:evaluation-controls}
\begin{tabular}{@{}p{0.25\linewidth}p{0.68\linewidth}@{}}
\toprule
Control & Comparison \\
\midrule
Step-0 hidden-state probe & Distinctions decodable before training versus gains in decodability during training. \\
Shuffled/random labels & Probe predictability under targets unrelated to the task assignment. \\
Prompt-token probe & Hidden-state predictions versus information in token presence/counts. \\
Gaussian-feature probe & Hidden-state predictions versus fitting an equally sized unrelated feature matrix. \\
WordNet matched controls & Lexical labels versus frequency, length, row norm, spacing, and token-class metadata or matched token sets. \\
Answer-row geometry & Differences in contrast scores attributable to candidate norms and token-frequency patterns. \\
Matched-feature ablations & Selected reconstructed contributions versus edits matched on decoder norm and firing rate, with an additional attribution-sign control. \\
\bottomrule
\end{tabular}
\end{table}

\section{Measurement Setup and Reproducibility}
\label{sec:app-reproducibility}

\subsection{Code and Data Release}
\label{sec:app-reproducibility-statement}

Code is available at \url{https://github.com/hematteo/learning-to-read-out}, the trained dictionaries for Pythia-160M, 1B, 6.9B, and OLMo-2-7B with derived aggregates and evaluation outputs at \url{https://huggingface.co/hematteo/parameter-trajectory-crosscoders}, and the checkpoints of the controlled 31M runs at \url{https://huggingface.co/hematteo/readout-recipe-control}. The code repository provides training and evaluation implementations and instructions for external model and data inputs.

\subsection{Models, Tokenizers, and Checkpoint Schedules}
\label{sec:app-models-checkpoints}
\label{sec:app-selected-checkpoint-schedules}
The primary trajectory is \href{https://huggingface.co/EleutherAI/pythia-1b}{\texttt{EleutherAI/pythia-1b}} ($\dmodel=2048$), with \href{https://huggingface.co/EleutherAI/pythia-160m}{\texttt{EleutherAI/pythia-160m}} ($\dmodel=768$) as the smaller setting, \href{https://huggingface.co/EleutherAI/pythia-6.9b}{\texttt{EleutherAI/pythia-6.9b}} ($\dmodel=4096$) as a larger-scale check, and \href{https://huggingface.co/allenai/OLMo-2-1124-7B}{\texttt{allenai/OLMo-2-1124-7B}} ($\dmodel=4096$) as a cross-architecture check \citep{biderman2023pythia,olmo2024olmo2}.

The Pythia models share the GPT-NeoX BPE tokenizer, fixed across checkpoints, with $V=50{,}304$ readout rows for Pythia-160M and Pythia-1B and $V=50{,}432$ padded rows for Pythia-6.9B.
OLMo-2-7B has its own tokenizer and $V=100{,}352$ readout rows.

Checkpoint selection is fixed before any dictionary is fit.
Pythia runs use $K=32$ checkpoints, loaded from Hugging Face revisions \texttt{step\{N\}}:
\[
\begin{gathered}
0, 1, 2, 4, 8, 16, 32, 64, 128, 256, 512,\\
1000, 2000, 3000, 4000, 5000, 6000, 7000, 8000, 9000,\\
14000, 21000, 27000, 34000,\\
47000, 61000, 75000, 89000, 102000, 116000, 130000, 143000.
\end{gathered}
\]
These are 20 checkpoints before step 10k, 4 from step 14k to 34k, and 8 from step 47k to 143k.
OLMo-2-7B runs use 32 stage-1 checkpoints, loaded from revisions named \texttt{stage1-stepXXX-tokensYYYB}:
\[
\begin{gathered}
150, 600, 700, 850, 900, 1000, 2000, 3000, 4000, 5000,\\
6000, 7000, 8000, 9000, 14000, 21000, 27000, 34000,\\
47000, 110000, 173000, 236000, 299000, 362000,\\
425000, 488000, 614000, 677000, 740000, 803000,\\
866000, 928000.
\end{gathered}
\]

\subsection{Checkpoint-Suite Selection}
\label{sec:app-checkpoint-suite-selection}
Tracking token rows needs a fixed tokenizer, accessible $\WU$ weights, and enough checkpoints around step 1000 to resolve the early readout changes.
Pythia is the controlled family because each model has 154 public checkpoints (step 0, log-spaced steps 1 to 512, then every 1000 steps to 143k) and all scales share one tokenizer, one data order, and an untied readout.
OLMo-2 is the cross-architecture check, with a public stage-1 trajectory that starts before step 1000 and shares many later steps with the Pythia grid.
Its architecture, tokenizer, data, optimizer, and schedule all differ from Pythia's, so it is not a point on the Pythia scaling axis.
\Cref{tab:checkpoint-suite-comparison} lists the other public suites we considered and why we did not use each.

\begin{table}[H]
\centering
\small
\caption{Other public checkpoint suites considered for $\WU$ trajectory analysis, and why each was not used.}
\label{tab:checkpoint-suite-comparison}
\begin{tabular}{@{}>{\raggedright\arraybackslash}p{0.3\linewidth}>{\raggedright\arraybackslash}p{0.66\linewidth}@{}}
\toprule
Suite & Reason not used \\
\midrule
\href{https://huggingface.co/allenai/OLMo-7B-0424-hf}{OLMo 7B, April 2024} \citep{groeneveld2024olmo} & Sparser coverage before step 1000 than OLMo-2 \\
\href{https://huggingface.co/allenai/Olmo-3-1025-7B}{Olmo 3} \citep{teamolmo2025olmo3} & No stage-1 checkpoint between steps 0 and 1000 \\
\href{https://huggingface.co/LLM360/Amber}{LLM360 Amber} \citep{liu2023llm360} & Checkpoints indexed \texttt{ckpt\_000} to \texttt{ckpt\_359} instead of by training step \\
\href{https://huggingface.co/LLM360/Crystal}{LLM360 Crystal} \citep{liu2023llm360} & Phase-specific revisions, custom loading, and a code and data mixture \\
Cerebras-GPT Intermediate \citep{dey2023cerebrasgpt} & No documented early step schedule, and some maximal update parametrization ($\mu$P) variants need custom loading \\
\href{https://huggingface.co/openlm-research/open_llama_3b_step_100000}{OpenLLaMA previews} \citep{openlm2023openllama} & Sparse previews that begin after step 1000 (3B from step 2.5k) \\
\href{https://huggingface.co/TinyLlama/tinyLlama-intermediate-checkpoints}{TinyLlama} \citep{zhang2024tinyllama} & First intermediate checkpoint at step 5k (10B tokens) \\
\href{https://huggingface.co/HuggingFaceTB/SmolLM3-3B-checkpoints}{SmolLM3} \citep{bakouch2025smollm3} & Checkpoints every 40k steps (about 94.4B tokens) \\
\href{https://huggingface.co/trillionlabs/Tri-7B-Intermediate-Checkpoints}{Tri family} \citep{trillionlabs2025tri} & Coarse token intervals, and a multilingual mix that needs new lexical controls \\
\bottomrule
\end{tabular}
\end{table}

\subsection{Unembedding Convention and Tied Embeddings}
\label{sec:app-unembedding-convention}
The Pythia unembedding is \texttt{model.embed\_out.weight}, a layer without bias that Hugging Face GPT-NeoX stores with shape $(V,\dmodel)$, so row $v$ is the output direction for token $v$.
Pythia's input embedding $\WE$ (\texttt{gpt\_neox.embed\_in.weight}) is a separate, untied parameter, and the readout analyses use only \texttt{embed\_out.weight}.
The OLMo-2-7B unembedding is \texttt{model.lm\_head.weight}.
Extracted unembeddings are stored as fp32 CPU tensors, one per checkpoint.

\subsection{Sparse Dictionary Architecture and Loss}
\label{sec:app-dictionary-architecture-loss}
The dictionaries follow the checkpoint-spanning crosscoder recipe of \citet{ge2026crosscoder}, applied to aligned $\WU$ rows instead of activations from a text corpus.
Each has $F=\dsae$ features, $K=32$ checkpoint-specific encoder and decoder heads with a decoder bias, and the JumpReLU gate of \cref{eq:crosscoder} \citep{rajamanoharan2024jumping}.
Each is trained on the objective of \cref{eq:crosscoder-objective} (\cref{sec:app-method-objective}) with $\alpha=0.01$ and $c=1.0$.
The inputs are already centered and scaled (\cref{sec:app-method-normalization}), so the dictionary library applies no further normalization to them.
The decoder is initialized uniformly on $[-1/\sqrt{\dsae},1/\sqrt{\dsae}]$, and the encoder is then set to the decoder transpose, which replaces its initial uniform draw on $[-1/\sqrt{\dmodel},1/\sqrt{\dmodel}]$.
JumpReLU thresholds are learned in log space, starting from the initial values in \cref{tab:repro-dictionary-hparams}.
The table gives the remaining settings of each main dictionary.
Near-inactive features can keep small nonzero decoder norms, so decoder norm trajectories are read together with firing rate and reconstruction checks.

\begin{table}[H]
\centering
\caption{
Sparse-dictionary training hyperparameters for every main dictionary setting.
Each column is a complete recipe for one model family and matrix setting. All dictionaries train in fp32 on centered-and-scaled vocabulary rows (\texttt{center\_scale}) for 300 epochs with batches of 1024 rows and no auxiliary loss on dead features (AuxK).
Adam-family optimizers use the adaptive-moment method of \citet{kingma2015adam}.
}
\label{tab:repro-dictionary-hparams}
\resizebox{\linewidth}{!}{%
\begin{tabular}{llllll}
\toprule
Hyperparameter & Pythia-1B & Pythia-160M & Pythia-6.9B $\lambda_1=0.6$ & Pythia-6.9B $\lambda_1=0.3$ & OLMo-2-7B \\
\midrule
$\dsae$ & 24,576 & 24,576 & 32,768 & 32,768 & 32,768 \\
Expansion factor & 12.0 & 32.0 & 8.0 & 8.0 & 8.0 \\
Sparsity coefficient $\lambda_1$ & 0.3 & 0.3 & 0.6 & 0.3 & 0.3 \\
Optimizer & Adam & Adam & Adam & Adam & SparseAdam \\
Learning rate & $2\times 10^{-5}$ & $5\times 10^{-5}$ & $1\times 10^{-5}$ & $1\times 10^{-5}$ & $1\times 10^{-5}$ \\
JumpReLU LR factor & 0.2 & 0.1 & 0.4 & 0.3 & 0.3 \\
JumpReLU init threshold & 0.1 & 0.1 & 0.15 & 0.1 & 0.1 \\
Optimizer steps & 15,000 & 15,000 & 15,000 & 15,000 & 29,400 \\
Sparsity warmup fraction & 0.05 & 0.05 & 0.05 & 0.10 & 0.10 \\
LR warmup / decay & 0 / 0 & 0 / 0 & 0 / 0 & 0.10 / 0.20 & 0.10 / 0.20 \\
Training strategy & single process & single process & 4-GPU head parallel & 4-GPU head parallel & 4-GPU head parallel \\
Compute & A100-class GPU(s) & A100-class GPU(s) & A100/H100-class GPUs & A100/H100-class GPUs & A100/H100-class GPUs \\
Seeds & 0 & 0, 1, 2 & 0 & 0 & 0 \\
\bottomrule
\end{tabular}
}
\end{table}

Each epoch visits all vocabulary rows in a fresh permutation from a CPU \texttt{torch.Generator} seeded by the run seed, so a run takes $\lceil V/1024\rceil \times 300$ optimizer steps.
We keep the final epoch without early stopping, and rolling checkpoints save the generator state.
EV, mean $L_0$, and dead feature counts (\cref{sec:app-instrument-validation}) are computed afterwards over the full vocabulary and play no role in selecting checkpoints.
The head-parallel strategy follows \citet{ge2026crosscoder}, and the Pythia-6.9B $\lambda_1=0.3$ run is a denser capacity point.

\begin{table}[H]
\centering
\caption{
Compute for reproducing the main dictionary fits and the controlled 31M pretraining runs, by hardware class rather than specific cloud offers.
The 31M wall clock is estimated from the $10$B-token budget and the recorded throughput of about $131$k tokens/s. The 31M row covers the four compared runs and leaves out the excluded fifth run (\cref{sec:app-recipe-control-setup}).
}
\label{tab:compute-resources}
\resizebox{\linewidth}{!}{%
\begin{tabular}{lllll}
\toprule
Run family & Worker class & GPU memory & Wall clock & GPU-hours \\
\midrule
Pythia-160M $\WU/\WE$ checkpoint-spanning fits & 1--3 A100-class GPUs & 40--80 GB/GPU & 4--8 h & 4--24 \\
Pythia-1B $\WU$ capacity sweep & 1 A100-80GB GPU & 80 GB & up to 12 h & up to 12 \\
Pythia-6.9B $\WU$ checkpoint-spanning fit, per $\lambda_1$ setting & 4 A100/H100-class GPUs & 40--80 GB/GPU & 4--8 h & 16--32 \\
OLMo-2-7B $\WU$ checkpoint-spanning fit & 4 A100/H100-class GPUs & 40--80 GB/GPU & 4--8 h & 16--32 \\
31M pretraining runs (four compared) & 1 A40 GPU per run & 48 GB & about 21 h per run & about 85 in total \\
Probe, null, baseline, and figure jobs & CPU workstation or 1 GPU & 24--80 GB if GPU & minutes--4 h & up to 4 \\
\bottomrule
\end{tabular}
}
\end{table}

The four-GPU fits need batch size 1024 on 40 GB GPUs and can use 2048 on 80 GB GPUs.
The wall-clock ranges exclude checkpoint extraction.
The Pythia-1B width sweep uses $\dsae \in \{8192,16384,24576\}$, expansion factors of about $\{4.0,8.0,12.0\}$.

\subsection{Existing Assets, Licenses, and Terms}
\label{sec:app-assets-licenses}

The Pythia checkpoints and tokenizers are released by EleutherAI under Apache-2.0, as stated on the Hugging Face model cards for the Pythia suite \citep{pythia160mHf,pythia1bHf,pythia69bHf}.
The OLMo-2-1124-7B checkpoints and tokenizer are released by Ai2 under Apache-2.0 \citep{olmo2Hf}.
The controlled 31M runs train on Pythia's preshuffled Pile \citep{gao2020pile,biderman2023pythia}, distributed by EleutherAI on Hugging Face as \texttt{EleutherAI/pile-standard-pythia-preshuffled}, and use it for research; we do not redistribute the data.
WordNet is used under the Princeton WordNet 3.0 license, including its notice, disclaimer, and non-endorsement terms \citep{wordnetLicense}.
The BLiMP families are used under CC BY 4.0 \citep{warstadt2020blimp}, and the function vector datasets under the MIT license \citep{todd2024function}.
The Wikipedia slices used for the multilingual readout-swap corpus come from Wikimedia database dumps; Wikimedia text contributions are available under CC BY-SA and GFDL terms under the Wikimedia Terms of Use \citep{wikimedia2023wikipedia,wikimediaTerms}.
The experiments use standard Python scientific and machine-learning dependencies, including PyTorch, Transformers, Datasets, NumPy, SciPy, Pandas, Matplotlib, and scikit-learn, under their respective open-source licenses \citep{paszke2019pytorch,wolf2020transformers,lhoest2021datasets,harris2020array,virtanen2020scipy,mckinney2010pandas,hunter2007matplotlib,pedregosa2011scikit}.
The code repository, the dictionaries, and the 31M checkpoints are released under the MIT license.
We do not redistribute the original Pythia or OLMo checkpoints or raw Wikipedia dumps; the released dictionaries and derived measurements identify the upstream assets and should be used subject to the original licenses and terms.

\section{Dictionary Validation and Quality}
\label{sec:app-instrument-validation}

This appendix validates the checkpoint-spanning sparse dictionaries of \cref{sec:method}, with the Pythia-1B dictionary of the main analyses as the primary case.
\Cref{fig:main-multimodel-validation} compares the selected dictionaries across models and, on Pythia-160M, compares a checkpoint-spanning dictionary with independent per-snapshot SAEs; \cref{tab:full-inventory} lists every fit.
Explained variance (EV) is one minus the mean squared reconstruction error divided by the variance of the centered and scaled rows, computed in sample on the rows the dictionary was fit to (\cref{sec:app-method-normalization}). Crosscoder $L_0$ is the number of features with nonzero activation at one row and checkpoint pair, averaged over all pairs, and active density is $100\cdot L_0/\dsae$. A feature is active if its maximum decoder norm over checkpoints, $\max_t\|D_f^{(t)}\|_2$, exceeds $0.01$, and dead if its firing rate over all row and checkpoint pairs is at most $10^{-6}$.
Several checks use the Pythia-160M $\dsae=8{,}192$ dictionary, a narrower fit than the selected Pythia-160M $\dsae=24{,}576$ dictionary.

\begin{figure}[H]
    \centering
    \setlength{\abovecaptionskip}{3pt}
    \includegraphics[width=\linewidth,trim=7pt 9pt 7pt 8pt,clip]{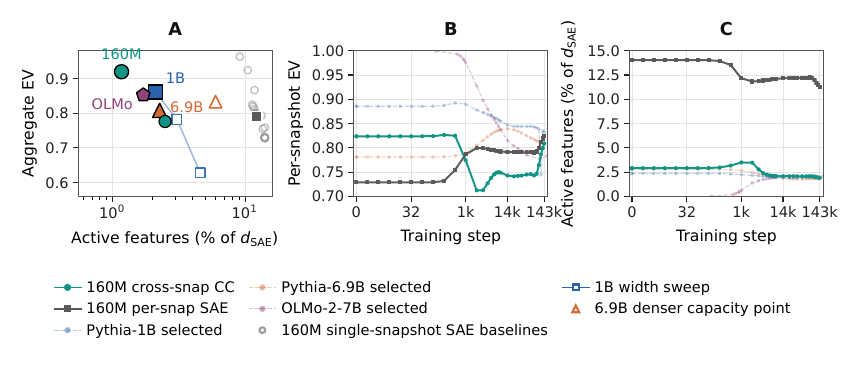}
    \caption{
        The selected dictionaries reconstruct the readout trajectories with aggregate EV $0.808$--$0.920$ while activating $1.17$--$2.26\%$ of their features per row, and on Pythia-160M a checkpoint-spanning dictionary stays within $0.1$ EV of independent per-snapshot SAEs at every checkpoint while activating $1.9$--$3.5\%$ of its features against their $11$--$14\%$.
        (A) Aggregate EV against active density $100\cdot L_0/\dsae$ for the four selected dictionaries (large labeled markers), the Pythia-1B width sweep, the denser Pythia-6.9B $\lambda_1=0.3$ fit, the Pythia-160M $\dsae=8{,}192$ dictionary, and the Pythia-160M single-snapshot SAE baselines (\cref{tab:full-inventory}).
        (B) Per-checkpoint EV and (C) per-checkpoint active density.
        Solid lines compare the Pythia-160M $\dsae=8{,}192$ seed-0 dictionary (``160M cross-snap CC''), not the selected $\dsae=24{,}576$ dictionary, with the 32 per-snapshot SAEs of the same width; dashed lines show the selected Pythia-1B, Pythia-6.9B, and OLMo-2-7B dictionaries for reference.
        OLMo-2-7B is plotted on its own stage-1 step axis, which panels B and C cut at step $180$k, so its last $11$ checkpoints are not shown.
    }
    \label{fig:main-multimodel-validation}
    \vspace{-0.5em}
\end{figure}

\subsection{Mean-Direction and Preprocessing Check}
\label{sec:app-mean-direction-preprocessing}

Raw $\WU$ geometry is dominated by a global offset shared across vocabulary rows.
After early training, the mean row $\mu^{(t)} = V^{-1}\sum_{v=1}^{V} \WU^{(t)}[v,:]$ aligns with the top right singular vector of raw $\WU$, with absolute cosine close to 1 at the final checkpoint of Pythia-1B, Pythia-6.9B, and OLMo-2-7B (\cref{fig:app-mean-direction-preprocessing}), and it inflates raw spectral and cosine geometry.
Centering removes the positive floor in mean pairwise row cosine and lowers the top spectral gap $\sigma_1/\sigma_2$, estimated by block power iteration, at late checkpoints in all three models, so all sparse dictionaries and dense baselines use centered rows with a per-snapshot scalar rescaling (\texttt{center\_scale}).

	\begin{figure}[H]
	    \centering
	    \includegraphics[width=\linewidth]{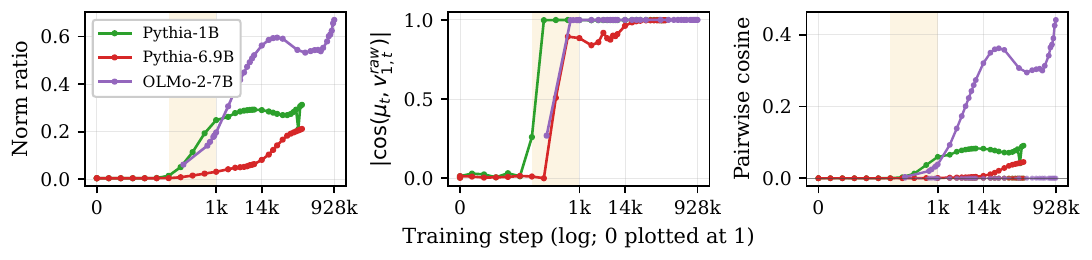}
	    \caption{
	        The mean row of checkpointed $\WU$ grows and aligns with the raw top right singular vector after early training, inducing a positive row-cosine floor, so later analyses use centered rows.
	        The norm ratio is the mean-row norm relative to the median row norm, and the right panel compares raw and centered mean pairwise row cosine.
	        The shaded band marks steps $64$--$1$k; OLMo-2-7B is plotted on its own stage-1 schedule.
	        Pythia-160M is not shown for readability.
	    }
	    \label{fig:app-mean-direction-preprocessing}
	\end{figure}

\subsection{Next-Token Distribution Fidelity of the Reconstructed Readout}
\label{sec:app-functional-fidelity}

Matrix EV measures agreement row by row, so we also decode the hidden states of a fixed held-out shard through the original and the reconstructed $\WU$ at each checkpoint.
The shard is the first 32 sequences of 512 tokens ($16{,}352$ predictions in English, Russian, and Chinese) from the multilingual corpus of the readout swaps (\cref{sec:app-readout-swap}), and $\Delta\mathrm{NLL}$ is the mean increase in target NLL when the reconstructed readout replaces the original.

Fidelity varies non-monotonically over training and is least stable before step $2$k (\cref{fig:app-fidelity-curves}).
At the terminal checkpoint, the selected Pythia-1B and Pythia-6.9B reconstructions agree with the original readout on $0.70$ and $0.69$ of top-1 predictions, with $\Delta\mathrm{NLL}$ of $0.23$ and $0.35$~nats.
The narrower Pythia-160M $\dsae=8{,}192$ dictionary keeps logit shape at the final checkpoint but loses scale calibration ($\Delta\mathrm{NLL}$ of $3.758$ nats/token), so matrix EV alone is not sufficient evidence at small widths.

\begin{figure}[H]
    \centering
    \includegraphics[width=0.85\linewidth,trim=7pt 7pt 6pt 6pt,clip]{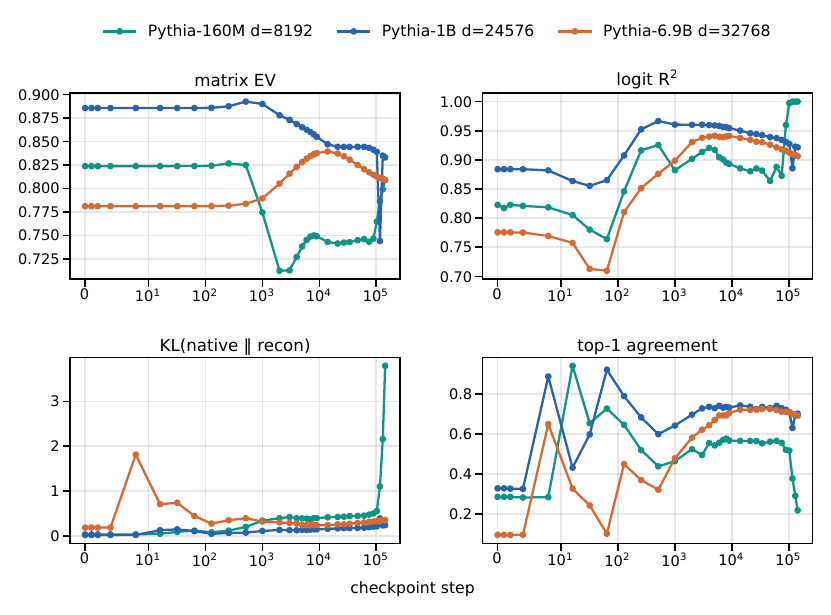}
    \caption{
        Next-token fidelity of the reconstructed readouts varies over training, and the Pythia-160M $\dsae=8{,}192$ dictionary keeps logit shape but loses scale calibration late in training.
        At each of the 32 checkpoints, hidden states from the fixed held-out shard are decoded through both the original (native) and reconstructed readouts.
        The third panel plots $\mathrm{KL}(\text{native}\,\|\,\text{reconstructed})$, whereas the text reports $\Delta\mathrm{NLL}$.
    }
    \label{fig:app-fidelity-curves}
\end{figure}

\paragraph{Token-stratified check.}
Within token-frequency deciles, visible-length bins, and coarse token classes, the primary Pythia-1B reconstruction stays faithful at step $1000$, with matrix EV of at least $0.868$ in every stratum and target-conditioned $\Delta\mathrm{NLL}$ of at most $0.42$ nats (for ASCII punctuation).
No single stratum explains the step-$116\text{k}$ dip in aggregate EV (\cref{sec:app-late-checkpoint-anomaly}).

\subsection{Sparse Autoencoder Baselines}

On Pythia-160M, we compare checkpoint-spanning $\WU$ dictionaries with final-snapshot SAEs and 32 independent per-snapshot SAEs.
Reconstruction barely changes across fit seeds and rises with width (\cref{fig:main-multimodel-validation,tab:full-inventory}).
Per-snapshot SAEs reconstruct individual checkpoints but are much denser, and the final-snapshot SAE fits the terminal readout but fails on early checkpoints (\cref{fig:main-multimodel-validation,fig:app-snapshot-fidelity}).
Neither gives a feature index shared across training at low $L_0$.
Checkpoint-spanning dictionaries are sparse in active-code density but not parameter-efficient, because they spend parameters on checkpoint-specific encoder and decoder heads to preserve a shared feature index (Params column of \cref{tab:full-inventory}).

	\begin{figure}[H]
	    \centering
	    \includegraphics[width=\linewidth]{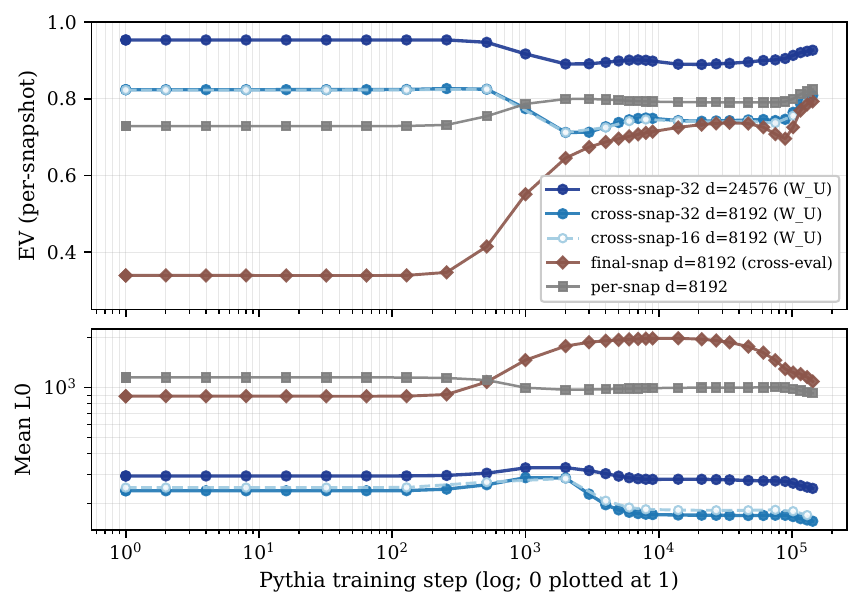}
	    \caption{
        On Pythia-160M, the checkpoint-spanning dictionaries ($\dsae=8{,}192$ and the selected $\dsae=24{,}576$) reconstruct early and late checkpoints with one feature index, while the SAE baselines either fail early or require far higher $L_0$.
	        The final-snapshot SAE (cross-eval) is trained on the final checkpoint and evaluated backward; per-snapshot SAEs are trained independently at each checkpoint.
	    }
	    \label{fig:app-snapshot-fidelity}
	\end{figure}

\subsection{Linear and Trajectory Baselines}
\label{sec:app-linear-baselines}

The dense baselines ask how much of the centered $\WU$ trajectory can be explained without a sparse shared feature basis, at ranks and active-code budgets that differ from the dictionaries', so they do not rank the methods at equal budget.

	\begin{figure}[H]
	    \centering
	    \includegraphics[width=\linewidth]{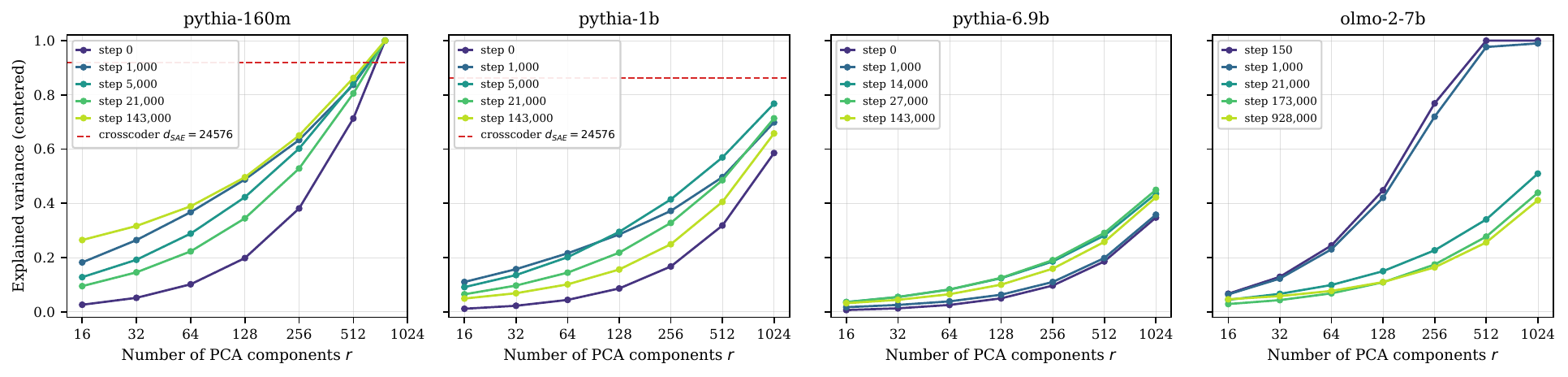}
	    \caption{
        Per-snapshot PCA of centered $\WU$ needs many dense components to approach the EV of the selected sparse dictionaries, and at rank $1024$ it falls short of them at every plotted Pythia-1B and Pythia-6.9B checkpoint and at the plotted OLMo-2-7B checkpoints from step $21$k onward.
        Each checkpoint is approximated independently by its best rank-$r$ PCA reconstruction.
        The dictionary reference line is drawn only for Pythia-160M and Pythia-1B; the selected Pythia-6.9B and OLMo-2-7B dictionaries reach aggregate EV $0.808$ and $0.853$ (\cref{tab:full-inventory}).
        Pythia-160M has $\dmodel=768$, so its PCA is exact at rank $768$.
	    }
	    \label{fig:app-pca-static}
	\end{figure}

Per-snapshot PCA needs many dense components to approach the selected dictionaries.
At the final checkpoint, rank $512$ explains $0.256$--$0.405$ of the centered variance, and rank $1024$ still falls short of the dictionaries at every Pythia-1B and Pythia-6.9B checkpoint and at OLMo-2-7B checkpoints from step $4$k onward (\cref{fig:app-pca-static}).
Pythia-1B would need roughly $1.36$k--$1.66$k components to match its dictionary.
The exception is OLMo-2-7B's earliest checkpoints (steps $150$--$3$k), where rank-$1024$ PCA exceeds the dictionary's aggregate EV and, at steps $150$ and $1$k, rank $512$ already explains nearly all centered variance.
A single dense subspace fitted to each model's centered token rows concatenated across checkpoints ($45$ for Pythia-160M, $32$ for Pythia-1B and OLMo-2-7B, and $15$ for Pythia-6.9B) does worse at every tested rank from $16$ to $256$, reaching EV $0.41$ in Pythia-160M and $0.23$ in Pythia-1B at rank $256$, against $0.92$ and $0.86$ for their dictionaries.

In Pythia, the dominant centered subspaces move most just before the $512$--$1.6$k reorganization window of \cref{sec:app-temporal-localization} and then stabilize (\cref{fig:app-dense-reorg-pythia}).
Top-$10$ drift peaks on the checkpoint pair ending at step $256$ in Pythia-160M and 1B and at step $512$ in Pythia-6.9B, one pair later at $k=50$.
This dense check locates change in time without identifying sparse features or token supports.

	\begin{figure}[H]
	    \centering
	    \includegraphics[width=\linewidth]{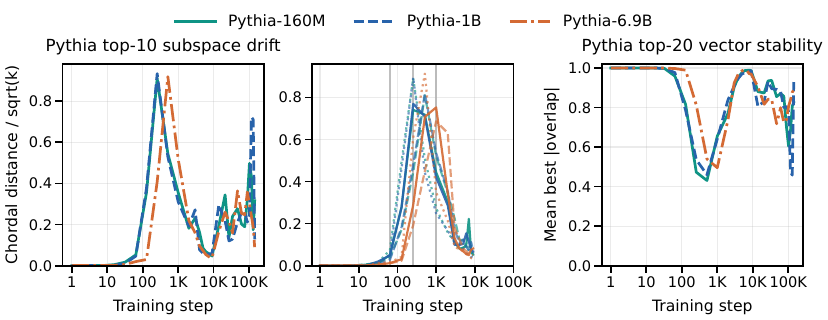}
	    \caption{
        Dominant centered $\WU$ subspaces in Pythia move most during early training, with the largest top-$10$ movement between steps $128$ and $512$; the later Pythia-1B step-$116{,}000$ anomaly is examined in \cref{sec:app-late-checkpoint-anomaly}.
        For each checkpoint, we center $\WU$, estimate top right-singular directions, and track adjacent-checkpoint chordal Grassmann distance (divided by $\sqrt{k}$ for a rank-$k$ subspace) and top-vector stability, plotted at the later checkpoint of each pair.
        In the left and right panels, color and line style mark the model.
        The middle panel restricts the left panel's distance to steps up to $9$k for subspace ranks $k=5$ (dotted), $20$ (solid), and $50$ (dashed), with colors as in the left panel; gray verticals mark steps $64$, $256$, and $1$k.
	    }
	    \label{fig:app-dense-reorg-pythia}
	\end{figure}

\subsection{Held-out Checkpoint Reconstruction}
\label{sec:app-heldout-checkpoints}

Checkpoint-spanning dictionaries are trained on 32 anchor checkpoints, so held-out reconstruction tests whether the fitted heads merely memorize those anchors.
Pythia has $154$ released checkpoints; the $122$ absent from the anchor schedule are unseen by construction.
We evaluate eight of them, at steps $10$k, $23$k, $33$k, $43$k, $73$k, $93$k, $113$k, and $123$k, so held-out coverage starts at step $10$k and misses the $512$--$1.6$k reorganization window and all earlier steps.
For a held-out step $t^\star$ with trained neighbors $t_-, t_+$ and log-spaced interpolation weight $\eta = (\log(1{+}t^\star) - \log(1{+}t_-))/(\log(1{+}t_+) - \log(1{+}t_-))$, we build an interpolated synthetic head from the neighboring encoder and decoder parameters,
\[
E^\eta = (1-\eta)\,E^{(t_-)} + \eta\,E^{(t_+)},
\qquad
b_{\mathrm{enc}}^\eta = (1-\eta)\,b_{\mathrm{enc}}^{(t_-)} + \eta\,b_{\mathrm{enc}}^{(t_+)},
\]
and analogously for the decoder parameters, leaving the shared JumpReLU threshold unchanged.
The shared code $z_v$ of token $v$ keeps the 32 anchor terms of \cref{eq:crosscoder}, computed from $v$'s rows at the anchor checkpoints, and adds the synthetic head's term $x_v^{(t^\star)}E^\eta+b_{\mathrm{enc}}^\eta$ on its held-out row; the gate then uses the synthetic decoder norms.
Only the reconstruction of the held-out row is scored, after applying the held-out checkpoint's \texttt{center\_scale} statistics, with nearest-head, raw-space interpolation, endpoint-line interpolation, and rank-$50$ PCA as references.

From step $10$k onward, neither tested dictionary shows a sharp anchor-memorization failure, although smoother trajectory interpolation remains possible.
For the Pythia-160M $\dsae{=}8{,}192$ seed-0 crosscoder, synthetic heads track the in-sample ceiling within $0.011$ EV at all eight held-out checkpoints, well above rank-$50$ PCA and endpoint interpolation and below direct raw-space interpolation, which has no sparsity bottleneck.
For the primary Pythia-1B crosscoder, the only deviations bracket the step-$116\text{k}$ anomaly, which the in-sample ceiling and direct $\WU$ interpolation also show (\cref{fig:app-heldout-checkpoints-1b}).

	\begin{figure}[H]
	    \centering
	    \includegraphics[width=0.85\linewidth]{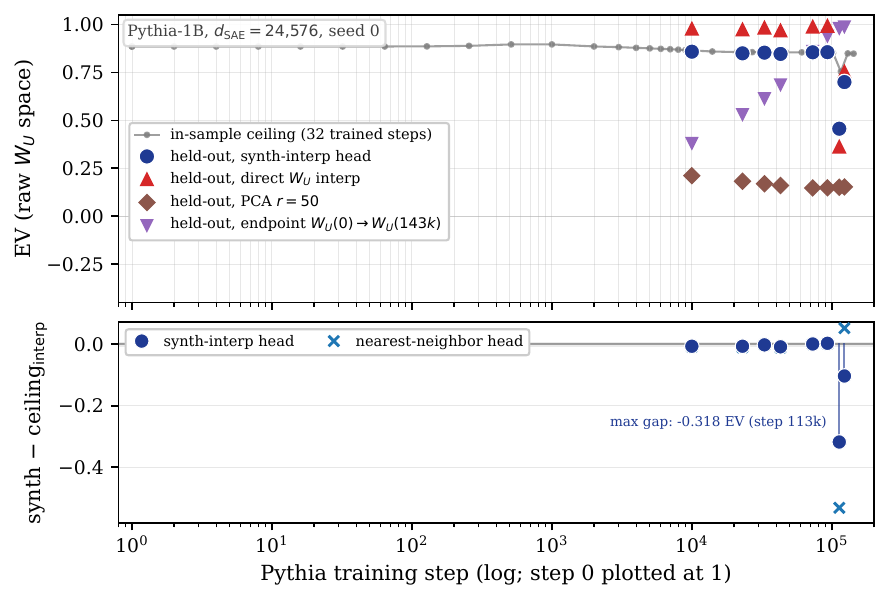}
	    \caption{
        Interpolated heads of the Pythia-1B $\dsae{=}24{,}576$ seed-0 crosscoder track the in-sample ceiling within $0.01$ EV at six held-out steps ($10\text{k}$--$93\text{k}$) and deviate only at $113\text{k}$ and $123\text{k}$.
        These deviations bracket the step-$116\text{k}$ anomaly (\cref{sec:app-late-checkpoint-anomaly}), where the in-sample ceiling dips to $0.755$ EV and direct $\WU$ interpolation drops to $0.36$ EV at $113\text{k}$.
	    }
	    \label{fig:app-heldout-checkpoints-1b}
	\end{figure}

\subsection{Sparsity-Penalty Sensitivity}

Retraining the Pythia-160M $\dsae=8{,}192$ dictionary at five sparsity weights from $\lambda_1=0.4$ to $1.8$ (seven fits, three of them at $\lambda_1=1.35$) traces a single reconstruction--sparsity tradeoff at fixed width, and the $\lambda_1=0.3$ fits of the same width lie at its high-EV end (\cref{fig:app-lambda-sweep}).
The sweep keeps the rest of the recipe but adds an auxiliary loss on dead features, which the $\lambda_1=0.3$ fits do not use.

	\begin{figure}[H]
	    \centering
	    \includegraphics[width=0.8\linewidth]{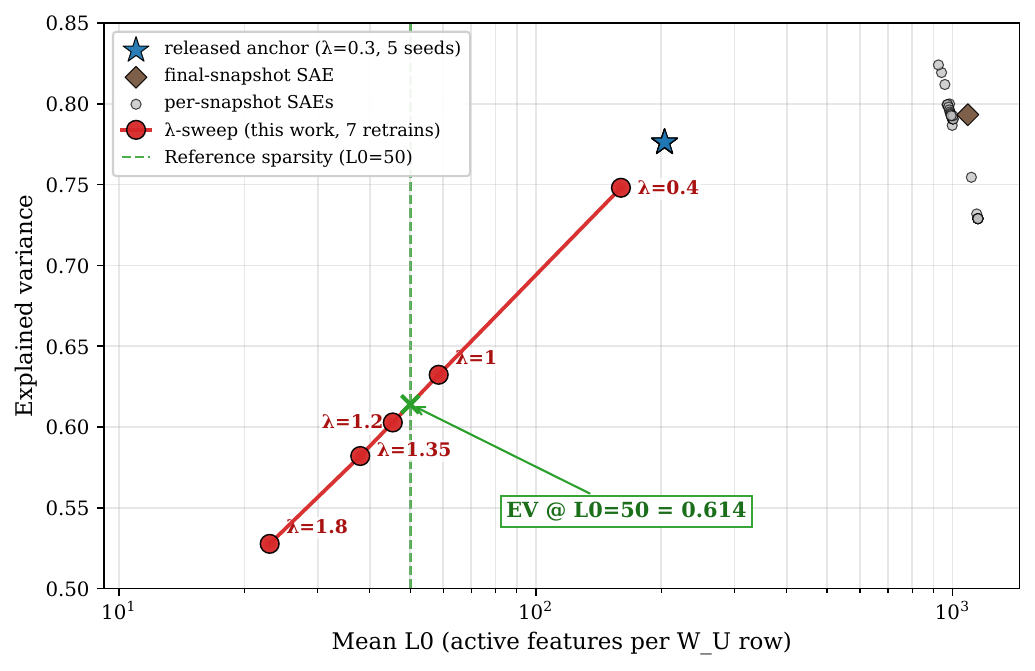}
	    \caption{
        For the Pythia-160M $\dsae=8{,}192$ dictionary, raising the sparsity weight $\lambda_1$ from $0.4$ to $1.8$ lowers mean $L_0$ from about $160$ to about $23$ and EV from $0.75$ to $0.53$ along a single tradeoff curve.
        Red points are the sweep fits (\cref{tab:app-lifecycle-stability}).
        The star (``released anchor'') marks the five-seed Pythia-160M $\dsae=8{,}192$ fits at $\lambda_1=0.3$, and the gray and brown markers are the same-width SAE baselines of \cref{tab:full-inventory}.
        The dashed green line marks $L_0=50$, near the low end of the sparsities of the Pythia-160M activation crosscoders of \citet{ge2026crosscoder} ($L_0\approx47$--$64$ in their Figure 2), and ``EV @ L0 = 50'' is the sweep's EV at that $L_0$, interpolated linearly in $\log L_0$ between adjacent sweep points.
	    }
	    \label{fig:app-lambda-sweep}
	\end{figure}

\subsection{Full Dictionary Inventory}

\Cref{tab:full-inventory} lists every dictionary and baseline behind a validation figure or numeric result, with selected dictionaries marked by $\star$.
The 160M, 1B, and OLMo checkpoint-spanning rows use the main $\lambda=0.3$ recipe; final-snapshot SAE baselines use $\lambda=0.05$ and 100 epochs; per-snapshot SAE baselines use $\lambda=0.3$ and 100 epochs.
$\WU$ crosscoders use penalty saturation scale $1.0$, except the Pythia-160M width-$16\,384$ fit marked $^{d}$, which uses $4.0$.
The final-snapshot SAE baselines also use saturation scale $4.0$. The init threshold is $0.1$ except for the selected Pythia-6.9B $\lambda_1=0.6$ retrain.

Raw mean $L_0$ is not comparable across dictionary widths, so the table also reports active density $100\cdot L_0/\dsae$, on which the selected dictionaries are similarly sparse and the single-snapshot SAE baselines several times denser.
In the selected Pythia-6.9B and OLMo-2-7B dictionaries, per-token active counts cluster near their medians ($734$ and $599$), so a small dense-token tail does not drive the averages.
Every feature of the four selected dictionaries, and of every Pythia fit in \cref{tab:app-lifecycle-stability}, is active, so the lifecycle analyses use all features of each dictionary.

{\footnotesize
\setlength{\tabcolsep}{2pt}
\begin{longtable}{@{}c l l r@{\hspace{6pt}}r@{\hspace{8pt}}l r r r r r@{}}
\caption{
Full dictionary inventory used across the validation figures.
$\star$ marks the selected dictionaries analyzed in \cref{sec:experimental-setup,sec:results}, each the seed-0 fit of its row, and rows with several fit seeds give ranges over seeds.
The \textbf{Exp.} column reports the expansion ratio $\dsae/\dmodel$ (Pythia-160M: $\dmodel{=}768$; Pythia-1B: $2\,048$; Pythia-6.9B and OLMo-2-7B: $4\,096$).
The \textbf{\% Act.} column reports $100\cdot L_0/\dsae$, i.e.\ the fraction of the dictionary active per token row.
For the distributed 6.9B and OLMo fits, EV and mean $L_0$ are recomputed over the full vocabulary after training.
$^{a}$ per-snap SAE family aggregates 32 per-checkpoint SAEs; $^{b}$ Pythia-6.9B retrain with $\lambda_1=0.6$ (selected dictionary; recipe in \cref{tab:repro-dictionary-hparams}); $^{c}$ Pythia-6.9B run trained at $\lambda_1=0.3$ (denser capacity-point); $^{d}$ fit with saturation scale $4.0$ instead of $1.0$.
}
\label{tab:full-inventory} \\
\toprule
$\star$ & \textbf{Model} & \textbf{Matrix / Family} & $\dsae$ & \textbf{Exp.} & \textbf{Seed} & \textbf{Params} & \textbf{EV} & \textbf{Mean $L_0$} & \textbf{\% Act.} & \textbf{Dead} \\
\midrule
\endfirsthead
\toprule
$\star$ & \textbf{Model} & \textbf{Matrix / Family} & $\dsae$ & \textbf{Exp.} & \textbf{Seed} & \textbf{Params} & \textbf{EV} & \textbf{Mean $L_0$} & \textbf{\% Act.} & \textbf{Dead} \\
\midrule
\endhead
\midrule
\multicolumn{11}{r}{\emph{continued on next page}} \\
\endfoot
\bottomrule
\endlastfoot
 & 160M & $\WU$ cross-snap-32 & 8\,192  & 10.7$\times$ & 0--4 & 403\,M  & 0.776 & 203.6--203.9 & 2.49 & 0.000 \\
 & 160M & $\WU$ cross-snap-32$^{d}$ & 16\,384 & 21.3$\times$ & 0 & 806\,M  & 0.780 & 104.9 & 0.64 & 0.000 \\
$\star$ & 160M & $\WU$ cross-snap-32 & 24\,576 & 32$\times$ & 0--2 & 1.21\,B & 0.920 & 286.4--287.2 & 1.17 & 0.000 \\
\midrule
 & 160M & $\WU$ final-snap SAE & 6\,144  & 8$\times$    & --- & 9.5\,M   & 0.758 & 864.2  & 14.07 & 0.000 \\
 & 160M & $\WU$ final-snap SAE & 8\,192  & 10.7$\times$ & --- & 12.6\,M  & 0.793 & 1\,088 & 13.28 & 0.000 \\
 & 160M & $\WU$ final-snap SAE & 16\,384 & 21.3$\times$ & --- & 25.2\,M  & 0.867 & 1\,921 & 11.72 & 0.000 \\
 & 160M & $\WU$ final-snap SAE & 32\,768 & 42.7$\times$ & --- & 50.4\,M  & 0.924 & 3\,424 & 10.45 & 0.000 \\
 & 160M & $\WU$ final-snap SAE & 65\,536 & 85.3$\times$ & --- & 100.8\,M & 0.963 & 5\,967 &  9.10 & 0.000 \\
\midrule
 & 160M & $\WU$ per-snap SAE$^{a}$ & 8\,192 & 10.7$\times$ & per-step & 12.6\,M & 0.729--0.824 & 925--1\,150 & 11.29--14.04 & 0.000 \\
\midrule
 & 1B   & $\WU$ cross-snap-32 & 8\,192  & 4$\times$  & 0 & 1.07\,B & 0.628 & 374.8 & 4.58 & 0.000 \\
 & 1B   & $\WU$ cross-snap-32 & 16\,384 & 8$\times$  & 0 & 2.15\,B & 0.781 & 500.2 & 3.05 & 0.000 \\
$\star$ & 1B & $\WU$ cross-snap-32 & 24\,576 & 12$\times$ & 0 & 3.22\,B & 0.861 & 516.6 & 2.10 & 0.000 \\
\midrule
$\star$ & 6.9B & $\WU$ cross-snap-32$^{b}$ & 32\,768 & 8$\times$ & 0 & 8.59\,B & 0.808 & 741.8   & 2.26 & 0.000 \\
        & 6.9B & $\WU$ cross-snap-32$^{c}$ & 32\,768 & 8$\times$ & 0 & 8.59\,B & 0.833 & 1\,956.6 & 5.97 & 0.000 \\
\midrule
$\star$ & OLMo-2-7B & $\WU$ cross-snap-32 & 32\,768 & 8$\times$ & 0 & 8.59\,B & 0.853 & 557.7 & 1.70 & 0.000 \\
\end{longtable}
}

\section{Sparse Lifecycle and Reorganization Timing}
\label{sec:app-lifecycle-diagnostics}

This appendix supports \cref{sec:readout-lifecycle} with the rule-based lifecycle profiles, aggregate timing summaries, the reorganization score, and the stability of these statistics across dictionary fits.

\subsection{Rule-Based Lifecycle Profiles}

For each feature, let $\rho_f(t)=\|D_f^{(t)}\|_2/\max_{t'}\|D_f^{(t')}\|_2$ be its decoder norm relative to its own maximum, and map checkpoints to normalized log-step time $\tau=(\log_{10}(t+1)-\log_{10}(t_{\min}+1))/(\log_{10}(t_{\max}+1)-\log_{10}(t_{\min}+1))\in[0,1]$ over each dictionary's checkpoint range $[t_{\min},t_{\max}]$.
The $+1$ offset keeps step $0$ finite, so Pythia's step $0$ sits at $\tau=0$.
In Pythia, whose range is steps $0$--$143$k, the early bin $\tau\le0.35$ covers steps up to $62$ (checkpoints $0$--$32$), and the late bin $\tau\ge0.65$ covers steps from $2{,}243$ (checkpoints $3$k--$143$k).
In OLMo-2-7B, whose range starts at step $150$ and ends at step $928$k, the early bin covers steps up to $3{,}197$ (checkpoints $150$--$3$k), and the late bin covers steps from $43{,}807$ (checkpoints $47$k--$928$k).
We summarize each trajectory by early strength $r_f^{\mathrm{early}}=\mathrm{mean}_{\tau\le0.35}\rho_f(t)$, late strength $r_f^{\mathrm{late}}=\mathrm{mean}_{\tau\ge0.65}\rho_f(t)$, mid-training peak $r_f^{\mathrm{mid,max}}=\max_{0.35<\tau<0.65}\rho_f(t)$, mid-training mean $r_f^{\mathrm{mid,mean}}=\mathrm{mean}_{0.35<\tau<0.65}\rho_f(t)$, peak time $\tau_f^{\mathrm{peak}}$, alive fraction $\phi_f^{\mathrm{alive}}=\Pr[\rho_f(t)\ge0.5]$, alive span $\Delta\tau_f^{\mathrm{alive}}=\tau_{\mathrm{last\,alive}}-\tau_{\mathrm{first\,alive}}$, and trajectory range $r_f^{\mathrm{range}}=\max_t \rho_f(t)-\min_t \rho_f(t)$.
Features are assigned in the order shown in \cref{tab:lifecycle-profile-rules}; later rows apply only if no earlier row matched, and \cref{fig:app-lifecycle-profile-composition} aggregates the assignments.

\begin{table}[H]
    \centering
    \small
    \setlength{\tabcolsep}{4pt}
    \renewcommand{\arraystretch}{1.12}
    \caption{
        Rule-based lifecycle profile assignment used in \cref{fig:app-lifecycle-profile-composition}.
        Rules are applied top to bottom, so each active feature receives exactly one descriptive label.
        The labels ``persistent'', ``early-decaying'', ``late-emerging'', ``transitional'', and ``mixed/ambiguous'' are thresholded trajectory descriptions.
    }
    \label{tab:lifecycle-profile-rules}
    \begin{tabular}{@{}p{0.20\linewidth}p{0.53\linewidth}p{0.19\linewidth}@{}}
        \toprule
        \textbf{Profile} & \textbf{Rule} & \textbf{Meaning} \\
        \midrule
        persistent &
        $(\phi_f^{\mathrm{alive}}\ge0.70\ \mathrm{and}\ r_f^{\mathrm{range}}\le0.35)$ or $\min(r_f^{\mathrm{early}},r_f^{\mathrm{mid,mean}},r_f^{\mathrm{late}})\ge0.45$ &
        sustained across training \\
        transitional &
        $r_f^{\mathrm{mid,max}}\ge0.75$, $r_f^{\mathrm{mid,max}}-\max(r_f^{\mathrm{early}},r_f^{\mathrm{late}})\ge0.35$, $r_f^{\mathrm{late}}\le0.30$, $\Delta\tau_f^{\mathrm{alive}}\le0.35$, and $0.35<\tau_f^{\mathrm{peak}}<0.65$ &
        rare localized mid-training peak \\
        early-decaying &
        $r_f^{\mathrm{early}}\ge0.50$, $r_f^{\mathrm{early}}-r_f^{\mathrm{late}}\ge0.30$, and $\tau_f^{\mathrm{peak}}\le0.45$ &
        strong early, weaker late \\
        late-emerging &
        $r_f^{\mathrm{late}}\ge0.50$, $r_f^{\mathrm{late}}-r_f^{\mathrm{early}}\ge0.30$, and $\tau_f^{\mathrm{peak}}\ge0.55$ &
        weak early, stronger late \\
        mixed/ambiguous &
        no previous rule matched &
        residual category \\
        \bottomrule
    \end{tabular}
\end{table}

\subsection{Aggregate Timing Summaries}

In the Pythia dictionaries, the $512$--$1.6$k reorganization window is where early-peaking features decay rather than where features peak.
Peaks cluster at steps $16$--$128$ and, in Pythia-160M and 1B, again late in training, with at most $2\%$ inside the window.
Features that peak by step $128$ keep a mean $\rho_f$ of at least $0.98$ through step $256$, which falls by step $3$k to $0.29$ in Pythia-160M and $0.63$ in Pythia-1B as total decoder norm mass also falls.

\begin{figure}[H]
    \centering
    \includegraphics[width=0.9\linewidth]{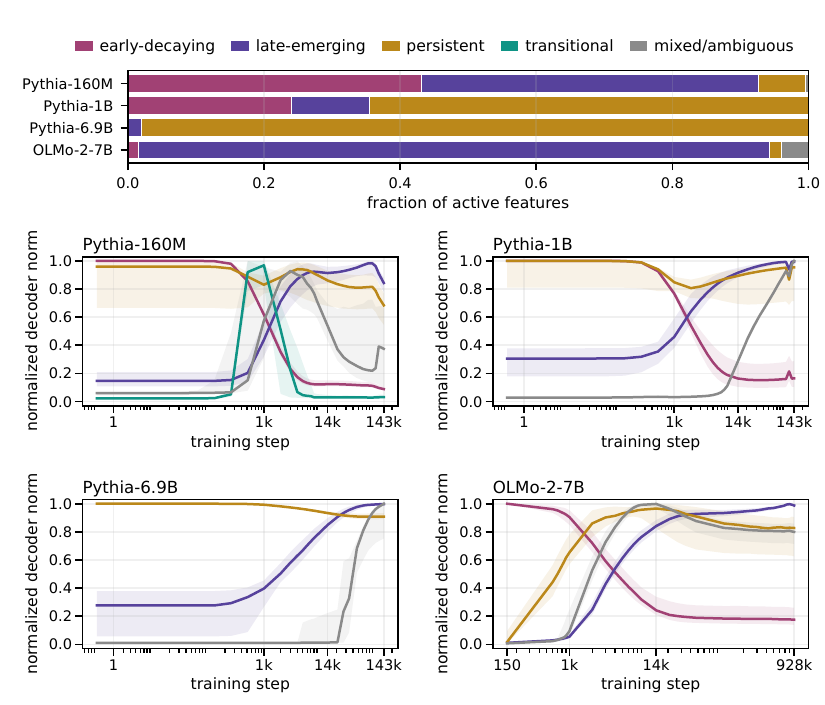}
    \caption{
        The four selected cross-snapshot sparse dictionaries contain persistent and late-emerging features, with substantial early-decaying shares in Pythia-160M and Pythia-1B.
        Sharply transitional mid-training peaks are rare, so readout features do not follow a single monotone growth pattern.
        Bars give the share of active features in each rule-based profile, and lines give each profile's median trajectory with interquartile bands.
    }
    \label{fig:app-lifecycle-profile-composition}
\end{figure}

\subsection{Late-Start Control for OLMo}
\label{sec:app-olmo-checkpoint-window-control}

Few OLMo-2-7B features peak early and then decay, unlike those of the selected Pythia-160M and Pythia-1B dictionaries.
Possible explanations include the missing OLMo-2-7B checkpoints before step $150$, more monotone readout development, underrepresentation of weak early-only directions by a checkpoint-spanning dictionary, and differences in tokenizer, data mixture, optimizer, and checkpoint spacing.
To test the first, we fit a Pythia-1B $\WU$ crosscoder on a late-starting 32-checkpoint schedule beginning at step $256$, which removes the very early checkpoints but none of the other Pythia--OLMo differences.

Without the early checkpoints, the early-decaying profile almost vanishes, at $0.02\%$ of late-start features ($\tau$ normalized over steps $256$--$143$k) against $24.1\%$ in the selected Pythia-1B dictionary, whose own trajectories restricted to steps from $256$ onward still give $25.4\%$.
The profile therefore arises when the dictionary is fitted together with checkpoints near initialization, with the underlying Pythia model unchanged.
Activation-rate definitions agree.
Features whose activation rate peaks by step $1{,}000$ and ends below half of its peak make up $0.012\%$ of late-start and $0.015\%$ of OLMo-2-7B features against $6.5\%$ in the selected dictionary, and a looser early-dominant score, $\mathrm{mean}_{t\le1\mathrm{k}} r_f(t)-\mathrm{mean}_{t\ge14\mathrm{k}} r_f(t)>0.2$ for peak-normalized activation rate $r_f$, gives $0.057\%$ and $0.595\%$ against $41.0\%$.
OLMo's missing early-decaying features are therefore better read as an effect of the checkpoint range than as evidence for a qualitatively different readout lifecycle.
The two trajectory distributions still differ, since $91\%$ of the late-start features peak at the first checkpoint without decaying below half of their peak, whereas OLMo-2-7B features mostly peak late.

\FloatBarrier

\subsection{Temporal Localization and Null Models}
\label{sec:app-temporal-localization}

This subsection defines the reorganization score and window of \cref{sec:readout-lifecycle}, which ask whether several non-identical views of readout change in sparse coordinates concentrate in the same training window.
The window is a multi-metric co-occurrence peak, and individual change measures can reach their maxima outside it.

For each adjacent checkpoint pair, we define six structural components over the active features (\cref{sec:app-instrument-validation}): decoder norm turnover, activation rate turnover, weighted adjacent decoder rotation, movement toward the terminal decoder direction, instability of the top feature set, and peak time density.
Norm and activation turnover are weighted mean absolute changes in max-normalized trajectories, with pair weight $\max(\|D_f^{(t_i)}\|_2,\|D_f^{(t_{i+1})}\|_2)$.
Adjacent rotation and movement toward the terminal decoder direction use weight $\sqrt{\|D_f^{(t_i)}\|_2\|D_f^{(t_{i+1})}\|_2}$.
Movement toward the terminal decoder direction is the absolute change in cosine between $D_f^{(t)}$ and the feature's final decoder direction.
Instability of the top feature set is one minus the Jaccard overlap of the top features by decoder norm, using the larger of $10\%$ of active features or 256 features, and peak time density assigns each feature's peak mass to the adjacent pair ending at its peak checkpoint.

We integrate each component over candidate windows of width $0.5$ in $\log_{10}(\mathrm{step}+1)$ by converting adjacent-pair values to rates per log-step interval and summing the overlap with each window.
Candidate windows start at each checkpoint whose window ends by the final checkpoint, so their stride follows the checkpoint schedule; the Pythia window starting at step $512$ ends at step $1{,}621$.
Within each model, each integrated component is min--max normalized, and we define the reorganization score as the geometric mean of the six normalized components.
Aggregate decoder-norm mass growth is integrated and plotted as a control, but is excluded from the score.

For nulls, each model uses 200 circular-shift draws.
In each draw, the six adjacent-pair component trajectories are independently shifted by nonzero offsets before integration and aggregation.
The null-standardized score subtracts the mean of the null scores for each window and divides by their standard deviation, and the selected window maximizes it; empirical one-sided $p$ values use a plus-one correction.
To test the choice of window, we resample active features with replacement 400 times per model, recompute the metrics, use 50 circular-shift null draws per resample, and count how often each window is selected.

\Cref{tab:app-reorganization-window-stats} gives the selected windows, their null probabilities, and how often feature resamples select each window, which shows the Pythia-6.9B window as the least stable of the three.

\begin{table}[H]
\centering
\footnotesize
\caption{Windows selected by the reorganization score, in training steps. $p$ is the one-sided circular-shift probability with a plus-one correction over 200 draws, so $0.005$ is the smallest attainable value. The last two columns count how many of 400 resamples of active features select the same window and each other window.}
\label{tab:app-reorganization-window-stats}
\begin{tabular}{@{}lcccp{0.38\linewidth}@{}}
\toprule
Model & Selected window & $p$ & Same window & Other windows (resamples) \\
\midrule
Pythia-160M & 512--1.6k & 0.005 & 370 & 1k--3.2k (30) \\
Pythia-1B & 512--1.6k & 0.005 & 283 & 1k--3.2k (117) \\
Pythia-6.9B & 512--1.6k & 0.04 & 222 & \mbox{9k--28k (134)}, \mbox{14k--44k (33)}, \mbox{1k--3.2k (10)}, \mbox{8k--25k (1)} \\
OLMo-2-7B & 9k--28k & 0.12 & 399 & 900--2.8k (1) \\
\bottomrule
\end{tabular}
\end{table}

\begin{figure}[H]
    \centering
    \includegraphics[width=\linewidth]{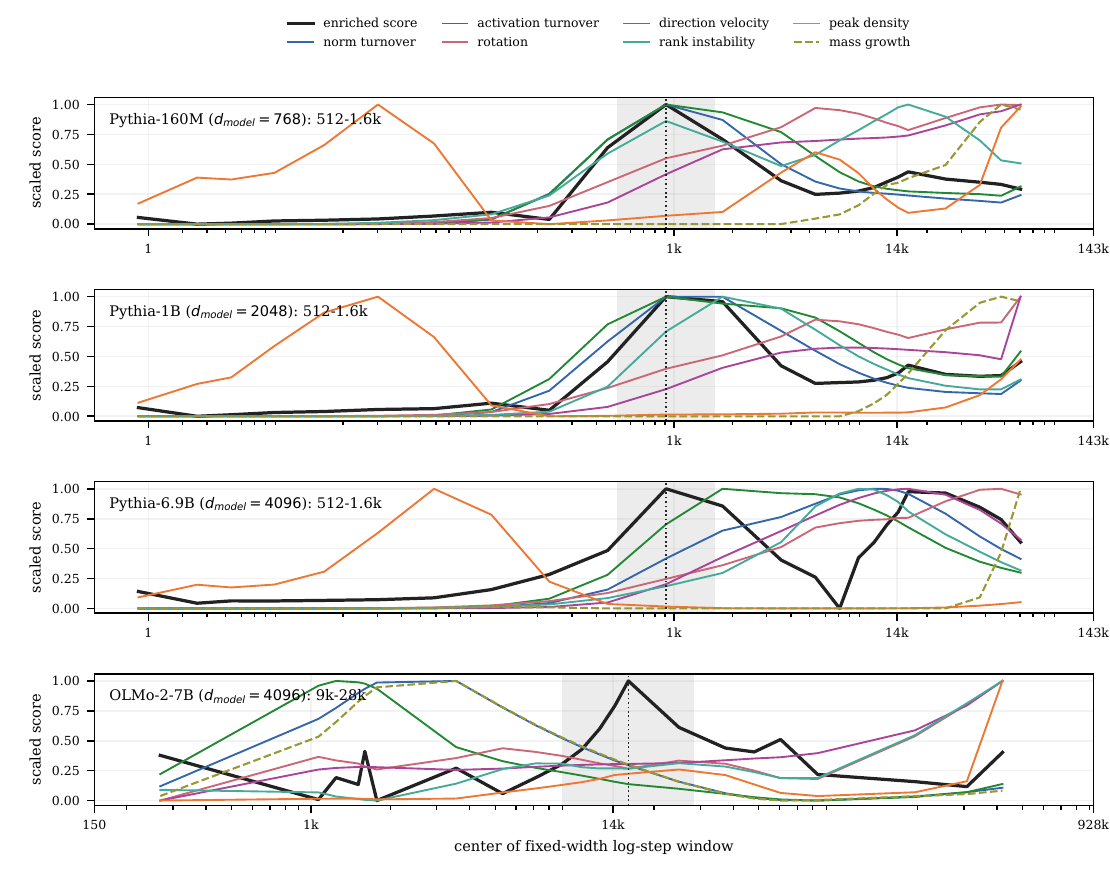}
    \caption{
        The three Pythia dictionaries select the same reorganization window, steps 512--1.6k, where several feature change metrics co-occur more than expected under shifted-component nulls.
        OLMo-2-7B's highest-scoring window, 9k--28k stage-1 steps, is not separated from the nulls (\cref{tab:app-reorganization-window-stats}).
        Each curve is plotted at its window's center and min--max scaled within each model.
        Black curves (``enriched score'') show the null-standardized reorganization score, gray bands mark the selected windows, and dotted verticals mark their centers.
        In the legend, ``direction velocity'' is movement toward the terminal decoder direction and ``rank instability'' is instability of the top feature set; the other four components match the text.
        Mass growth is plotted as a control and is excluded from the score.
    }
    \label{fig:app-reorganization-window-metrics}
\end{figure}

\subsection{Stability Across Dictionary Fits}
\label{sec:app-lifecycle-stability}

To test whether the lifecycle statistics describe the readout or one dictionary fit, we recompute the profile fractions of \cref{tab:lifecycle-profile-rules} and the reorganization window, with the same code, rules, and null draws, for all 21 Pythia dictionaries fitted with this architecture on the 32-checkpoint schedule (\cref{tab:app-lifecycle-stability}); OLMo-2-7B has one fit.

The reorganization window is stable across fits.
Twenty of the 21 fits, covering 12 of the 13 configurations, select steps 512--1.6k, and the exception, Pythia-1B at width 16{,}384, selects steps 1k--3.2k, the same alternative that feature resamples of the selected Pythia-1B dictionary pick (\cref{tab:app-reorganization-window-stats}).

The profile fractions and the median feature peak step agree across fit seeds, within about one percentage point and one checkpoint, but move with dictionary settings.
Larger sparsity weights, and in Pythia-1B larger widths, assign more features to the persistent profile, and across settings the median peak step ranges from step 0 to step 8k.
Every Pythia-160M and Pythia-1B dictionary contains all three profiles, although one of them can fall below $1\%$, and Pythia-6.9B is almost entirely persistent at both sparsity weights.
We therefore read the profile fractions qualitatively, do not attribute differences in them to the models alone, and compare the four 31M dictionaries of \cref{sec:recipe-control} only at their shared width and sparsity weight.

\begin{table}[H]
\centering
\footnotesize
\setlength{\tabcolsep}{3.5pt}
\caption{Lifecycle statistics across dictionary fits. Profile shares are percentages of active features; rows with several fit seeds give the range over seeds. The window is the one selected by the reorganization score, and $p$ is its circular-shift probability (200 draws). $^{*}$Selected dictionaries of the main text. $^{\dagger}$Fit with penalty saturation scale $c=4$ instead of $c=1$. The rows with $\lambda_1$ from $0.4$ to $1.8$ come from a separate sweep that also used an auxiliary loss on dead features.}
\label{tab:app-lifecycle-stability}
\begin{tabular}{@{}lrrrrrrll@{}}
\toprule
Model & Width & $\lambda_1$ & Seeds & Early-decaying & Persistent & Late-emerging & Window & $p$ \\
\midrule
Pythia-160M & 8{,}192 & 0.3 & 5 & 45.6--46.1 & 0.5--0.9 & 52.5--53.0 & 512--1.6k & 0.005 \\
Pythia-160M$^{\dagger}$ & 16{,}384 & 0.3 & 1 & 7.8 & 80.9 & 11.2 & 512--1.6k & 0.005 \\
Pythia-160M$^{*}$ & 24{,}576 & 0.3 & 3 & 42.9--43.1 & 6.9 & 49.5--49.7 & 512--1.6k & 0.005 \\
Pythia-160M & 8{,}192 & 0.4 & 1 & 43.2 & 1.2 & 55.0 & 512--1.6k & 0.005 \\
Pythia-160M & 8{,}192 & 1.0 & 1 & 27.8 & 43.4 & 28.3 & 512--1.6k & 0.005 \\
Pythia-160M & 8{,}192 & 1.2 & 1 & 22.3 & 57.5 & 19.7 & 512--1.6k & 0.005 \\
Pythia-160M & 8{,}192 & 1.35 & 3 & 17.8--18.1 & 66.2--67.3 & 14.4--15.2 & 512--1.6k & 0.005 \\
Pythia-160M & 8{,}192 & 1.8 & 1 & 6.8 & 82.7 & 10.0 & 512--1.6k & 0.005 \\
Pythia-1B & 8{,}192 & 0.3 & 1 & 41.4 & 21.7 & 36.9 & 512--1.6k & 0.010 \\
Pythia-1B & 16{,}384 & 0.3 & 1 & 34.0 & 44.8 & 21.2 & 1k--3.2k & 0.005 \\
Pythia-1B$^{*}$ & 24{,}576 & 0.3 & 1 & 24.1 & 64.4 & 11.4 & 512--1.6k & 0.005 \\
Pythia-6.9B & 32{,}768 & 0.3 & 1 & 0.0 & 98.1 & 1.9 & 512--1.6k & 0.04 \\
Pythia-6.9B$^{*}$ & 32{,}768 & 0.6 & 1 & 0.0 & 98.0 & 2.0 & 512--1.6k & 0.04 \\
\bottomrule
\end{tabular}

\end{table}

\FloatBarrier

\section{WordNet Supersense Probe Analyses}
\label{sec:app-vocabulary-family-probes}

This appendix gives the protocol and results of the Pythia-160M WordNet analysis of \cref{sec:vocabulary-family-readout} and its Pythia-1B replication.

\subsection{Probe Hyperparameters}
\label{sec:app-probe-hyperparameters}
Each WordNet category is probed one-vs-rest on the mean-centered $\WU$ rows. Its positives are the category's clean token ids, and its negatives are all other vocabulary rows, including the tokens of other categories, with class weights that balance the two classes.
Each category uses one stratified split of vocabulary rows into two-thirds train and one-third held-out test, sampled with \texttt{StratifiedShuffleSplit} and run seed $0$ \citep{pedregosa2011scikit}. The split depends only on the labels and the seed, so each category keeps the same train and test rows at every checkpoint.
This row split assigns token ids independently, so case and spacing variants of one lemma (\texttt{\textvisiblespace dog}, \texttt{dog}, \texttt{\textvisiblespace Dog}) can fall on both sides of it. \Cref{sec:app-wordnet-lemma-split} repeats the probe with the split grouped by lemma. All other probe results in this appendix, including every figure and table, use the row split.
Within the training split, an inner 3-fold stratified cross-validation selects $C \in \{0.1,1,10\}$.
The selected classifier is refit on the full training split and evaluated once on the held-out split.
The reported WordNet analysis uses a batched GPU implementation that matches \texttt{LogisticRegression(solver='liblinear', class\_weight='balanced')} \citep{pedregosa2011scikit} to within a few thousandths in agreement checks, with \texttt{max\_iter=80}.
All reported probe values are held-out balanced accuracy.

\subsection{WordNet Supersense Inventory}

For each WordNet synset \citep{miller1995wordnet,fellbaum1998wordnet}, each one-word ASCII alphabetic lemma with length at least three is assigned to a lexicographer file only if that file accounts for at least $60\%$ of the lemma's synset occurrences.
Each surviving lemma is mapped through five surface variants (lowercase and capitalized, each with and without a leading space, and all capitals), and a variant is kept only if it is a single Pythia token. A kept token is clean if, after its leading space is stripped, it decodes to at least three ASCII alphabetic characters and it has a leading space, an initial capital, or all capitals, which excludes word-internal fragments. A category is probed only if at least 20 clean token ids survive.
This leaves 35 probed categories out of 45 WordNet lexicographer files.
The skipped support-limited files are: \texttt{adj.ppl} (15 clean token ids), \texttt{noun.Tops} (13), \texttt{noun.motive} (3), \texttt{noun.shape} (9), \texttt{verb.body} (8), \texttt{verb.competition} (8), \texttt{verb.consumption} (11), \texttt{verb.emotion} (12), \texttt{verb.perception} (16), and \texttt{verb.weather} (1).

\begin{figure}[H]
    \centering
    \includegraphics[width=\linewidth]{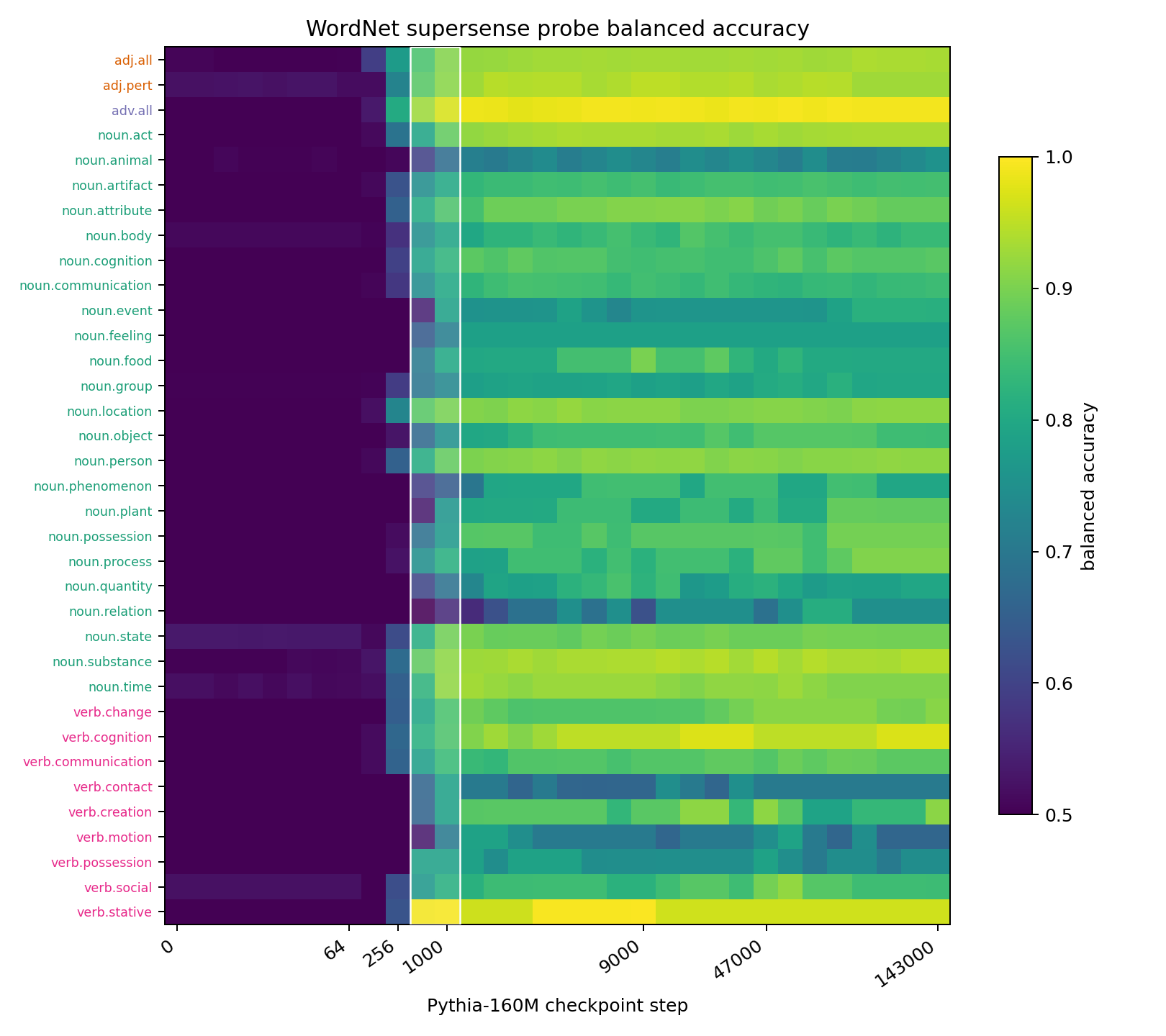}
    \caption{
        In Pythia-160M, all 35 WordNet supersense categories reach sustained decodability in the readout rows between steps $128$ and $1000$, and 33 have their largest adjacent-checkpoint gain in that interval.
        Rows are the 35 lexicographer-file categories with at least 20 clean token ids, in alphabetical order, and color is the held-out balanced accuracy of a one-vs-rest linear probe on centered $\WU$ rows under the row split of \cref{sec:app-probe-hyperparameters}. Columns are the 32 Pythia checkpoints, equally spaced by index. The white box marks steps $512$ and $1000$, the checkpoints inside the $512$--$1.6$k reorganization window.
    }
    \label{fig:app-wordnet-supersense-probes-160m-inventory}
\end{figure}

\subsection{Matched-Control Analysis}

A matched-control analysis repeats the same probe procedure at all 32 Pythia-160M checkpoints.
It compares observed WordNet labels with three controls. The metadata-only baseline is a probe on token features alone. These are the log count of the token in a held-out evaluation token set (not its training-corpus frequency), its character length, the norm of its centered row at the same checkpoint, spacing (whether the token begins with a space), special-token status, and a coarse token class (special, whitespace, lowercase, uppercase, or mixed-case alphabetic, digit, punctuation, other ASCII, or non-ASCII). The matched-random control replaces the category with tokens drawn from the same frequency, length, row norm, spacing, special-token, and token class bins, relaxing the match when a bin is empty, and the label-shuffle control shuffles labels within these bins.
The null probes draw new token sets and splits at every checkpoint.
The matched-random and label-shuffle controls use 20 permutations per category and null type, so they are interpreted as a confound check for this analysis. With 20 permutations the smallest attainable permutation $p$-value is $1/21\approx0.048$, so $p\le0.05$ means that the observed accuracy beats all 20 null values. The $95$th-percentile threshold used below interpolates between the two largest null values and is slightly less strict.

\subsection{Probe Timing Trajectories}

A category's onset is the first checkpoint at which it reaches sustained decodability, meaning balanced accuracy of at least $\max(0.55,\mathrm{acc}_{0}+0.05)$ at that checkpoint and the next two.
In Pythia-160M, lexical organization appears within a short interval (\cref{fig:app-wordnet-supersense-probes-160m-inventory}).
Onsets fall between steps $128$ and $1000$, 19 of them at step $256$, and 33 of 35 categories have their largest adjacent-checkpoint gain in that interval, whose three adjacent pairs a category's largest gain would hit with probability $0.10$ if it fell on a uniformly random pair of the grid.
Median probe accuracy starts near the matched-random and label-shuffle nulls at step $0$, and with each category's threshold set to the larger of its two $95$th-percentile nulls, it exceeds the median threshold by $0.170$ at step $512$ and $0.272$ at the final checkpoint.

\subsection{Pythia-1B Replication}

The timing replicates on the Pythia-1B $\WU$ trajectory, the model of the localization analysis in \cref{sec:sparse-coordinate-localization}, with the same 35 categories and probe settings under the row split.
Onsets range from step $128$ to step $2000$ with a median of $512$, and 33 of 35 categories have their largest gain in the four adjacent pairs from $128{\to}256$ through $1000{\to}2000$, against $0.13$ expected if each category's largest gain fell on a uniformly random pair.
Neither the metadata-only baseline nor the matched-random and label-shuffle nulls were run on Pythia-1B, so these are observed probe accuracies only.

\subsection{Lemma-Grouped Splits}
\label{sec:app-wordnet-lemma-split}

Under the row split, 1199 of the 3249 positive test tokens ($37\%$) have a variant of the same lemma in the training split. We therefore rerun the probe with every vocabulary row grouped by its lowercased, space-stripped decoded string. The outer split is \texttt{StratifiedGroupKFold} with three folds and seed $0$, holding out one fold as the test set, and the inner selection of $C$ uses the same grouped scheme on the training split. No positive lemma then appears on both sides. The categories, token ids, probe settings, and 32 checkpoints are unchanged, and each category keeps a test support within one token of the row split.

In Pythia-160M, median balanced accuracy rises from $0.500$ at step $0$ to $0.841$ at the final checkpoint, against $0.874$ under the row split. Of the 35 categories, 34 reach sustained decodability by step $1000$, 33 still have their largest adjacent-checkpoint gain between steps $128$ and $1000$, and the median onset stays at step $256$.
The category that never separates is \texttt{noun.relation}, whose 24 token ids come from 14 lemmas. It stays at $0.500$ at every checkpoint under the grouped split, so its accuracy under the row split came from variants of the same lemma falling on both sides of the split.
Across split seeds $1$ to $4$, the median per-category difference between accuracy under the lemma-grouped split and under the row split is about $-0.04$ at the final checkpoint and lies within $\pm0.005$ at steps $256$ and $512$.
In Pythia-1B, the median onset stays at step $512$, onsets range from step $128$ to step $1000$, and one category never separates. The final median is $0.060$ lower than under the row split.
The matched-random and label-shuffle nulls and the metadata-only baseline were not rerun under the grouped split. Against the Pythia-160M nulls computed under the row split, 32 of 35 categories under the lemma-grouped split exceed both $95$th percentiles at step $512$, and 34 do from step $1000$ to the final checkpoint.

\FloatBarrier

\section{Availability and Expression Across Contrastive Task Families}
\label{sec:app-availability-expression}
This appendix extends the probe and readout comparison to a broader suite of 20 task families over 24 checkpoints of Pythia-1B and Pythia-6.9B. The 32-checkpoint logistic probe trajectories for SVA and IOI are in \cref{fig:app-contrastive-readout-lag}.

\paragraph{Measurements.}
All measures read the final normalized hidden state. The 32-checkpoint SVA and IOI trajectories use the sweep $B_{\mathrm{env}}$ and the broader task suite uses $B_{\mathrm{ho}}$ (\cref{sec:app-evaluation-details}), and both exclude the native and terminal readouts, so neither need exceed $E$.
Most open-vocabulary families have no shared binary label, so \cref{tab:app-ae-openvocab} reports them by expression only. The BLiMP (Benchmark of Linguistic Minimal Pairs) anaphor and determiner--noun agreement families do have one, and \cref{sec:app-ae-timing} uses the anaphor family as a contrast case.

\begin{figure}[H]
\centering
\includegraphics[width=\linewidth]{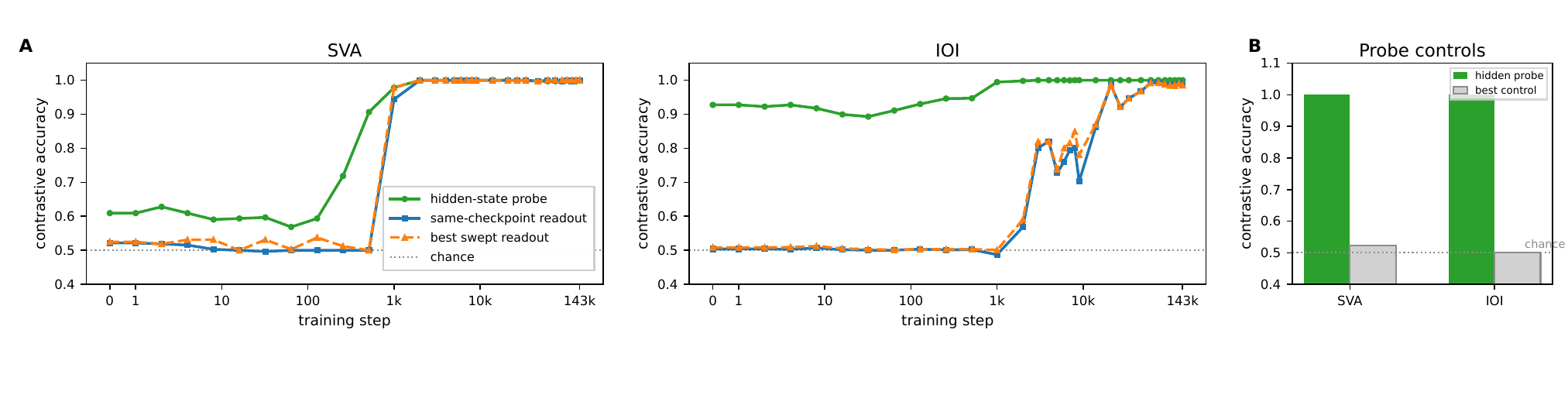}
\caption{In the 32-checkpoint Pythia-6.9B trajectories for SVA ($n=320$) and role-balanced IOI ($n=2{,}000$), the logistic hidden state probe reaches high accuracy before native contrast accuracy rises, and on SVA the best swept readout tracks the native one. The best swept readout is the post hoc envelope $B_{\mathrm{env}}$. IOI probe accuracy is already $0.928$ at initialization, so the entire early probe--expression gap cannot be attributed to learning. All three curves in panel A use the same task inventory, and panel B compares the probe with the strongest prompt-token or Gaussian-feature control.}
\label{fig:app-contrastive-readout-lag}
\end{figure}

\subsection{Prompt Construction}
\label{sec:app-task-construction}

The tasks of \cref{tab:task-protocol} and the other binary families of the broader task suite are built from fixed templates and word lists, and every item has a correct answer and a foil that are both single tokens with a leading space. The prompt stops where the answer would begin, and we score the margin at its last token. \Cref{tab:app-task-construction} gives the templates. The availability task suite and the localization items were built separately, with different templates for IOI and numeric comparison.

\paragraph{Availability task suite.}
Simple agreement crosses 20 noun lemmas (\texttt{key}, \texttt{book}, \texttt{student}, and others) in singular and plural form with eight fixed prepositional phrases whose nouns are singular (\texttt{to the cabinet}, \texttt{near the wall}, and six more), so every plural subject is followed by a singular attractor. The across-PP set pairs each singular item with a plural item that has the same head lemma and verb pair. Its heads are 30 person nouns, its prepositions are \texttt{by}, \texttt{near}, \texttt{beside}, \texttt{behind}, \texttt{next to}, \texttt{across from}, and \texttt{in front of}, its attractors are singular place and object nouns or plural person nouns, and its answers come from 12 verb pairs; head and attractor number differ in 966 of 2,000 items. The relative clause sets follow the constructions of \citet{marvin2018targeted}. They use 20 person nouns as heads and attractors (never the same noun twice in one item), six relative clause verbs, and five main verb pairs (\texttt{likes}/\texttt{like}, \texttt{runs}/\texttt{run}, \texttt{knows}/\texttt{know}, \texttt{writes}/\texttt{write}, \texttt{reads}/\texttt{read}). The two sets contain the same 1,200 combinations of head, attractor, and verbs and differ in clause structure, so the clause verb agrees with the head in a subject relative clause and with the attractor in an object relative clause. Role-balanced IOI uses two of 22 single-token first names and one of eight objects. For each name pair and object it produces both role orders, with a filler sentence naming the recipient (\texttt{Jane waited.}), so the two prompts of a pair contain the same tokens and only their order marks the answer. The three numeric sets use the same 1,000 unordered pairs of the numbers 2 to 99, each in both orders, so the correct answer comes first in half of the comparison items and \texttt{True} is correct in half of the true/false items.

\paragraph{Localization items.}
A separate builder produced the localization items with the Pythia tokenizer and the terminal row norm filter of \cref{sec:app-evaluation-details}. Simple agreement uses the same 320 prompts. IOI takes every ordered pair of the 22 names once, with a random object, in the template without the filler sentence, so the recipient is always the second name, and all 462 pairs pass the norm filter. In place of numeric comparison, localization uses a greater-than completion task that asks for a number smaller than the one in the prompt (\texttt{84 is greater than}, with answers \texttt{64}/\texttt{84}), so it contrasts a smaller number with the prompt's own number. The builder draws 2,000 ordered pairs of numbers from 2 to 99 without repeating a draw, but both orders of a pair give the same item, so an item can occur twice, and each of the at most 97 prompts recurs with different smaller numbers. Relational facts keep the 26 of 30 country and capital pairs that pass the single-token and norm filters. The foil is the first other capital in a fixed list that passes the filters, which makes \texttt{Paris} the foil for most items, so this family largely contrasts each capital with one token.

\paragraph{Splits.}
Availability probes use the grouped folds of \cref{sec:app-evaluation-details}. Localization splits each task's items into halves at random at the item level (split seed 0), so the selection and scoring halves share answer tokens and, for greater-than completion, prompts and occasionally whole items, since that builder can produce an item twice. The held-out half therefore tests whether the selected features carry the margin on other items built from the same answer tokens, and it does not test transfer to new tokens.

\begin{table}[H]
\centering
\footnotesize
\setlength{\tabcolsep}{3pt}
\caption{Prompt templates. Braces mark slots filled from the word lists in the text, and answers are the correct token and the foil, each with a leading space. In role-balanced IOI, \texttt{\{R\}} is the recipient and \texttt{\{S\}} the giver, one each of \texttt{\{A\}} and \texttt{\{B\}}. Set A is the availability task suite and set L the localization items.}
\label{tab:app-task-construction}
\begin{tabular}{@{}>{\raggedright\arraybackslash}p{0.15\linewidth}>{\raggedright\arraybackslash}p{0.45\linewidth}>{\raggedright\arraybackslash}p{0.22\linewidth}cr@{}}
\toprule
Task & Template & Answers & Set & $n$ \\
\midrule
Simple SVA & \texttt{The \{noun\} \{prepositional phrase\}} & \texttt{are}/\texttt{is} & A, L & 320 \\
Across-PP SVA & \texttt{The \{head\} \{preposition\} the \{attractor\}} & agreeing verb / other form & A & 2000 \\
Subject-RC SVA & \texttt{The \{head\} that \{clause verb\} the \{attractor\}} & agreeing main verb / other form & A & 1200 \\
Object-RC SVA & \texttt{The \{head\} that the \{attractor\} \{clause verb\}} & agreeing main verb / other form & A & 1200 \\
IOI & \texttt{\{A\} and \{B\} went to the store. \{R\} waited. \{S\} gave the \{object\} to} & \texttt{\{R\}}/\texttt{\{S\}} & A & 2000 \\
IOI & \texttt{\{A\} and \{B\} went to the store. \{A\} gave the \{object\} to} & \texttt{\{B\}}/\texttt{\{A\}} & L & 462 \\
Numeric $>$ & \texttt{Which is greater, \{x\} or \{y\}? Answer:} & larger / smaller number & A & 2000 \\
Numeric $<$ & \texttt{Which is smaller, \{x\} or \{y\}? Answer:} & smaller / larger number & A & 2000 \\
Numeric T/F & \texttt{\{x\} is greater than \{y\}. True or False? Answer:} & \texttt{True}/\texttt{False} by truth value & A & 2000 \\
$>$ completion & \texttt{\{x\} is greater than} & smaller number / \texttt{\{x\}} & L & 2000 \\
Relational facts & \texttt{The capital of \{country\} is} & capital / another capital & L & 26 \\
\bottomrule
\end{tabular}
\end{table}

\subsection{Task Items and Final Checkpoint Results}
\label{sec:app-ae-tasks}
\label{sec:app-ae-gap}
\Cref{tab:app-ae-binary} reports the binary families at the final checkpoint, and \cref{tab:app-ae-openvocab} reports the open-vocabulary and control families, which reach high contrast accuracy. The BLiMP anaphor family combines the BLiMP anaphor gender and anaphor number agreement sets, 1,000 items each. Its probe label is the gender of the reflexive, so $A_1$ is fit only on the 1,455 items whose correct token is \texttt{ herself} or \texttt{ himself}, while $E$, $B_{\mathrm{ho}}$, and the rank-1 rate use all 2,000 items.

\begin{table}[H]
  \centering
  \small
  \caption{Binary families at the final checkpoint
  (Pythia-6.9B), with the 1B gap and regime. $A_1@0$ is the step-0 initialization
  baseline, and the across-PP value of $0.43$ lies below chance for a reason the stored runs do not record. $B$ is $B_{\mathrm{ho}}$, and ``unreach'' $= (A_1 - B)/(A_1 - E)$ is the fraction of the gap that the evaluated intermediate readout does not reach, computed from unrounded values. The Pythia-1B numeric probes use stratified folds without number pair grouping. A gap is persistent when
  $A_1-E$ exceeds $0.05$ at the final checkpoint and transient when it ends at
  or below $0.05$ after exceeding $0.30$ while $E$ was below $0.7$.}
  \label{tab:app-ae-binary}
  \setlength{\tabcolsep}{3.4pt}
  \begin{tabular}{@{}lrcccccccl@{}}
    \toprule
    family & $n$ & $A_1@0$ & $A_1$ & $E$ & $B$ & $A_1\!-\!E$ & unreach & $A_1\!-\!E$(1B) & regime \\
    \midrule
    SVA simple & 320 & 0.60 & 1.00 & 1.00 & 1.00 & 0.00 & -- & 0.00 & transient \\
    SVA across-PP & 2000 & 0.43 & 1.00 & 0.97 & 0.98 & 0.03 & 63\% & 0.03 & transient \\
    SVA subject-RC & 1200 & 0.73 & 1.00 & 0.75 & 0.79 & 0.25 & 85\% & 0.21 & persistent \\
    SVA object-RC & 1200 & 0.73 & 1.00 & 0.91 & 0.93 & 0.10 & 72\% & 0.13 & persistent \\
    IOI & 2000 & 0.92 & 1.00 & 1.00 & 1.00 & 0.00 & -- & 0.00 & transient \\
    numeric $>$ & 2000 & 0.86 & 1.00 & 0.63 & 0.82 & 0.37 & 47\% & 0.55 & persistent \\
    numeric $<$ & 2000 & 0.86 & 1.00 & 0.45 & 0.66 & 0.55 & 62\% & 0.40 & persistent \\
    numeric T/F & 2000 & 0.89 & 0.99 & 0.51 & 0.66 & 0.48 & 70\% & 0.48 & persistent \\
    \bottomrule
  \end{tabular}
\end{table}

\begin{table}[H]
  \centering
  \footnotesize
  \caption{Open-vocabulary and control
  families (Pythia-6.9B), with a truncated example prompt and the contrast tokens $y^+/y^-$ (leading space shown). The BLiMP families
  come from \citet{warstadt2020blimp} and the function vector (FV) families, which are in-context prompts, from
  \citet{todd2024function}; BLiMP probe folds are grouped by antecedent name for anaphor agreement and by singular noun lemma for determiner--noun agreement. Apart from these two BLiMP agreement families, these carry no binary feature, so $A_1$ is undefined;
  expression is reported by contrast accuracy $E$, the swept accuracy $B=B_{\mathrm{ho}}$, and the
  full vocabulary rank-1 rate of $y^+$. NPI is negative polarity item licensing, and for BLiMP anaphor $n$ counts the gender-labeled items of \cref{sec:app-ae-tasks}. The control families are built from single-token nouns drawn at random. Fixed-token control prompts list two to five distinct nouns, and the answer is always \texttt{ the} against \texttt{ ,}, so the context carries no information about the answer. Copy control prompts have the form $a\,a\,b\,b\,c$, with answer $c$ and a foil absent from the prompt. Induction prompts have the form $a\,b\,a\,b\,a$, with answer $b$ and a foil absent from the prompt. Relational (bal.) gives each of $24$ countries five foils drawn at random from the other capitals, whereas relational facts uses the fixed foil list of the localization items (\cref{sec:app-task-construction}).}
  \label{tab:app-ae-openvocab}
  \setlength{\tabcolsep}{2.5pt}
  \begin{tabular}{@{}l>{\raggedright\arraybackslash}p{4.2cm}lrccc@{}}
    \toprule
    family & example prompt (truncated) & $y^+$/$y^-$ & $n$ & $E$ & $B$ & rank1 \\
    \midrule
    BLiMP anaphor & Katherine can't help & \texttt{ herself}\,/\,\texttt{ himself} & 1455 & 0.99 & 0.99 & 0.02 \\
    BLiMP det-noun & Raymond is selling this & \texttt{ sketch}\,/\,\texttt{ sketches} & 838 & 0.92 & 0.94 & 0.03 \\
    BLiMP NPI & Teresa had & \texttt{ not}\,/\,\texttt{ probably} & 656 & 0.98 & 0.97 & 0.15 \\
    copy control &  valley valley anchor anchor zebra & \texttt{ zebra}\,/\,\texttt{ jacket} & 2000 & 0.98 & 0.98 & 0.01 \\
    fixed-token control &  mirror thunder cookie music island & \texttt{ the}\,/\,\texttt{ ,} & 2000 & 0.98 & 0.90 & 0.00 \\
    FV antonym & selfish:altruistic / note:silence / known\dots & \texttt{ perfect}\,/\,\texttt{ socialist} & 2000 & 1.00 & 1.00 & 0.00 \\
    FV country-capital & Monaco:Monaco-Ville / Senegal:Dakar / Slo\dots & \texttt{ Vienna}\,/\,\texttt{ Moscow} & 36 & 1.00 & 0.94 & 0.00 \\
    FV en-fr & vector:vecteur / indonesia:Indon\'esie / ga\dots & \texttt{ ici}\,/\,\texttt{ tous} & 435 & 1.00 & 1.00 & 0.00 \\
    FV present-past & demonstrate:demonstrated / treat:treated \dots & \texttt{ accelerated}\,/\,\texttt{ fastened} & 288 & 1.00 & 1.00 & 0.00 \\
    induction &  lemon burner lemon burner lemon & \texttt{ burner}\,/\,\texttt{ basket} & 2000 & 0.98 & 0.98 & 0.70 \\
    relational facts & The capital of France is & \texttt{ Paris}\,/\,\texttt{ Berlin} & 26 & 1.00 & 1.00 & 0.96 \\
    relational (bal.) & The capital of France is & \texttt{ Paris}\,/\,\texttt{ Ottawa} & 120 & 1.00 & 0.98 & 0.96 \\
    \bottomrule
  \end{tabular}
\end{table}

\subsection{Initialization, Probe, and Answer-Token Controls}
\label{sec:app-ae-controls}
The step-0 probe ($A_1@0$ in \cref{tab:app-ae-binary}) separates information that becomes decodable during training from information already decodable at initialization, so each gap is read against it. On the three numeric families it already reaches $0.86$ to $0.89$, consistent with the probe learning a value for each number from its token identity, since folds grouped by number pair still share individual numbers. The numeric rows therefore measure a persistent expression gap and do not show that the model encodes the comparison.

Shuffled-label fits stay below $0.56$ on the agreement, IOI, and numeric families of \cref{tab:task-protocol} and reach $0.585$ on BLiMP anaphor agreement at Pythia-6.9B, and prompt-token and matched-dimension Gaussian-feature probes control for simple task cues in the 32-checkpoint trajectories. The availability result holds under both fitting choices, since at step $1000$ in Pythia-6.9B the shrinkage-LDA fit $A_1$ and the logistic fit $A_{\mathrm{full}}$ (\cref{sec:app-evaluation-details}) differ by at most $0.03$ on the eight binary families. Both fit a single direction, so this comparison checks the fitting method, and neither is trained with the model's next-token objective.

Answer-token norm and rarity percentiles correlate with the final gap at $+0.23$ and $+0.36$ across the eight families, a sample too small to show that geometry or frequency is irrelevant. The subject and object relative clause sets give a more direct check, because they share their verb pairs. At the end of training, native expression is higher for object than for subject relative clauses while both probes are at ceiling, so the answer tokens alone do not set the gap.

\subsection{Transient and Persistent Gaps}
\label{sec:app-ae-timing}
In Pythia-6.9B, simple and across-PP agreement and IOI close their gaps during training, while the relative clause and numeric gaps persist to the final checkpoint (\cref{fig:app-ae-grid}).

Among the families whose step-$0$ $A_1$ lies within $0.1$ of chance, so that any gap reflects information learned during training (\cref{tab:app-ae-early}), every subject--verb agreement row has $A_1$ near ceiling at step $512$ while native and swept expression are at or below chance, whereas in BLiMP anaphor agreement $A_1$ and native expression rise together. For simple and across-PP agreement, native expression has risen by step $1000$, the next checkpoint of this grid, so the lead spans a single checkpoint that the grid cannot resolve.

\begin{figure}[H]
\centering
\includegraphics[width=\linewidth]{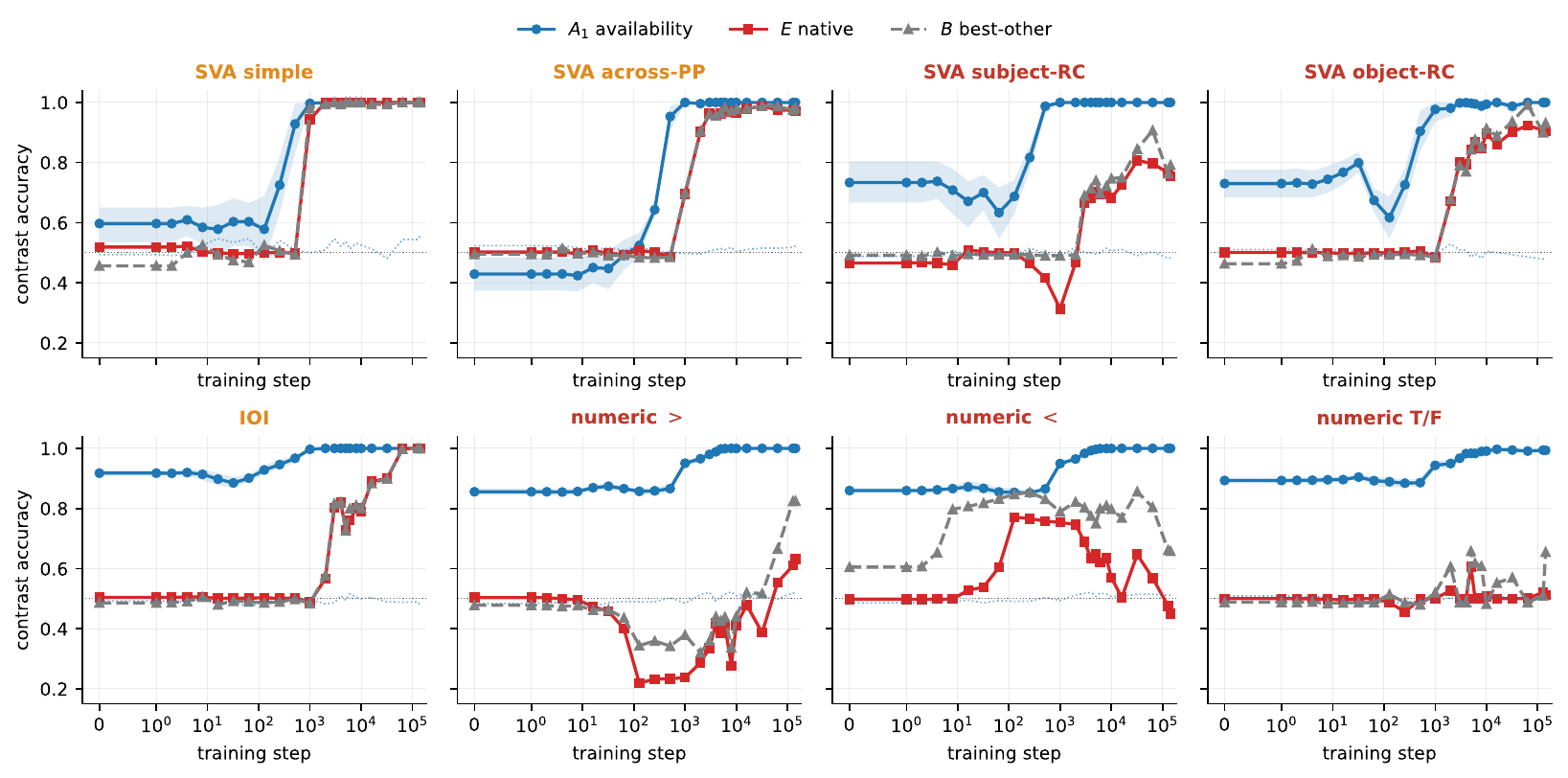}
\caption{Across the eight binary families in Pythia-6.9B, agreement and numeric tasks differ in both initialization baselines and final checkpoint gaps. Panel titles are colored by the regime of \cref{tab:app-ae-binary}, orange for a transient gap and red for a persistent one. $B$ (best-other in the legend) is $B_{\mathrm{ho}}$. Shaded bands are $95\%$ percentile bootstrap intervals for $A_1$ over its five per-fold accuracies, and dotted curves show the label-shuffle control (blue) and chance (black). Native expression on numeric $<$ rises to about $0.75$ between steps $128$ and $2000$ and falls to $0.45$ at the final checkpoint, so this family's final gap is larger than its gap in mid-training.}
\label{fig:app-ae-grid}
\end{figure}

\begin{table}[H]
  \centering
  \small
  \caption{Families whose probe starts within $0.1$ of chance, from the broader task suite ($24$ checkpoints; $A_1$ is the shrinkage-LDA probe and $B$ is $B_{\mathrm{ho}}$). For BLiMP anaphor, $n$ counts the items with a gender label used by $A_1$, and $E$ and $B$ use all $2{,}000$ items.}
  \label{tab:app-ae-early}
  \setlength{\tabcolsep}{4pt}
  \begin{tabular}{@{}llrccccccc@{}}
    \toprule
    & & & & \multicolumn{3}{c}{step $512$} & \multicolumn{2}{c}{step $2000$} & final \\
    \cmidrule(lr){5-7}\cmidrule(lr){8-9}
    model & family & $n$ & $A_1@0$ & $A_1$ & $E$ & $B$ & $A_1$ & $E$ & $E$ \\
    \midrule
    6.9B & SVA simple & 320 & 0.60 & 0.93 & 0.50 & 0.49 & 1.00 & 1.00 & 1.00 \\
    6.9B & SVA across-PP & 2000 & 0.43 & 0.95 & 0.49 & 0.48 & 1.00 & 0.90 & 0.97 \\
    6.9B & BLiMP anaphor & 1455 & 0.60 & 0.63 & 0.48 & 0.48 & 0.85 & 0.90 & 0.99 \\
    1B & SVA simple & 320 & 0.45 & 0.97 & 0.50 & 0.49 & 1.00 & 1.00 & 1.00 \\
    1B & SVA across-PP & 2000 & 0.53 & 0.98 & 0.48 & 0.48 & 1.00 & 0.93 & 0.97 \\
    1B & SVA subject-RC & 1200 & 0.56 & 1.00 & 0.22 & 0.39 & 1.00 & 0.55 & 0.79 \\
    \bottomrule
  \end{tabular}
\end{table}

\subsection{Contrast Accuracy and Full-Vocabulary Prediction}
\label{sec:app-ae-bottleneck}
A two-token contrast does not measure whether either candidate is the most likely token over the full vocabulary. At the final Pythia-6.9B checkpoint, native contrast accuracy on subject-relative-clause agreement is $0.75$, but the gold verb's rank-1 rate is zero. Numeric greater-than, by contrast, has a rank-1 rate near its contrast accuracy. The rank-1 rates in \cref{tab:app-ae-openvocab} likewise separate strong contrast scores from generation, most clearly in the function vector families, which reach $E=1.00$ with rank-1 rate $0.00$.
\FloatBarrier

\section{Behavioral and Intervention Readout Checks}
\label{sec:app-readout-swap}

This appendix documents the corpus-level readout swaps and the sparse feature interventions behind \cref{sec:sparse-coordinate-localization}.
Swaps exchange $\WU$ while keeping hidden states and targets fixed, and feature ablations remove selected reconstructed contributions to $\WU$ rows; neither intervenes on upstream circuits.

\subsection{Swap Grid Setup and Aggregate Measures}

The readout-swap grid evaluates Pythia-1B on a fixed held-out corpus of $173{,}229$ next-token predictions.
The corpus consists of nine $20{,}000$-character language slices from the 2023-11-01 Wikimedia Wikipedia dumps \citep{wikimedia2023wikipedia}, tokenized with the Pythia tokenizer and covering English, Russian, Chinese, Japanese, Thai, Arabic, Hindi, Korean, and Bengali.
Each slice concatenates articles of at least $200$ characters in dump order until it reaches $20{,}000$ characters, from at most $50$ articles, and is then truncated to $20{,}000$ characters; the number of articles in each slice was not recorded.
For each hidden-state checkpoint $t \in \{256,512,1000,2000,3000,14000,47000,143000\}$, we record the normalized input $h_p^{(t)}$ to the output layer at every held-out position $p$ and score it through the readouts $\WU^{(s)}$ of 32 released checkpoints from step 0 through step $143\text{k}$, with logits $h_p^{(t)}\left(\WU^{(s)}\right)^\top$ and no sparse dictionary or feature edit (\cref{fig:app-readout-swap-nll-kl-grid}).
A replication on Pythia-160M uses the same corpus and hidden-state checkpoints $t \in \{256, 512, 1000, 2000, 143000\}$ with the full 32-step readout schedule. It shows the same qualitative pattern, with many later readouts decoding early hidden states without large NLL spikes and later hidden states preferring nearby readouts, and it lacks the step-$116\text{k}$ spike (\cref{sec:app-late-checkpoint-anomaly}).

\subsection{Corpus Composition and Per-Language Swaps}
\label{sec:app-readout-swap-languages}

The swap corpus is mostly non-English, whereas Pythia was trained mostly on English. English supplies only $2.3\%$ of the scored targets (\cref{tab:app-readout-swap-languages}), while the Pile is about $97.4\%$ English by its authors' fastText estimate and includes a EuroParl component covering 21 European languages \citep{gao2020pile}. Pythia's BPE tokenizer, trained on the Pile \citep{black2022gptneox}, operates on bytes and so can encode text in any script \citep{radford2019language}, and small amounts of non-English text in nominally English pretraining corpora correlate with cross-lingual ability, most strongly in RoBERTa \citep{blevins2022language}, so next-token prediction on these languages is measurable in Pythia. Because the aggregate NLL in \cref{fig:app-readout-swap-nll-kl-grid} describes mainly non-English text, we also report it per language.

\begin{table}[H]
    \centering
    \footnotesize
    \setlength{\tabcolsep}{3pt}
    \caption{
        Readout swap gains by language for Pythia-1B.
        For hidden states from checkpoint $t$, $s^\ast$ is the readout checkpoint with the lowest NLL, and the gain is native NLL minus the NLL under $\WU^{(s^\ast)}$ in nats, so a gain of $0$ means that the native readout is best.
        Targets are scored next-token positions and share is their percentage of all targets; the aggregate is the token-weighted mean over all targets, including $8$ targets whose context token belongs to the preceding language slice.
        Brackets give $95\%$ paired bootstrap intervals from $2{,}000$ resamples of $256$-token blocks within each slice, which do not measure variation across documents, and $^\dagger$ marks a bootstrap probability below $0.95$ that $s^\ast$ is the minimum.
    }
    \label{tab:app-readout-swap-languages}
\begin{tabular}{@{}lrrrlrlrl@{}}
    \toprule
    & & & \multicolumn{2}{c}{$t=256$} & \multicolumn{2}{c}{$t=512$} & \multicolumn{2}{c}{$t=1000$} \\
    \cmidrule(lr){4-5}\cmidrule(lr){6-7}\cmidrule(lr){8-9}
    Language & Targets & Share (\%) & $s^\ast$ & Gain & $s^\ast$ & Gain & $s^\ast$ & Gain \\
    \midrule
        Aggregate & 173{,}229 & 100.0 & 1000 & 1.503 [1.455, 1.548] & 1000 & 0.147 [0.141, 0.153] & 1000 & 0 \\
        \midrule
        English & 3{,}939 & 2.3 & 512 & 0.048 [0.026, 0.070] & 1000 & 0.064 [0.039, 0.089] & 2000$^\dagger$ & 0.002 [$-$0.021, 0.026] \\
        Russian & 10{,}639 & 6.1 & 512 & 1.033 [1.007, 1.060] & 1000 & 0.114 [0.090, 0.139] & 1000$^\dagger$ & 0 \\
        Chinese & 25{,}330 & 14.6 & 1000 & 1.251 [1.176, 1.305] & 1000 & 0.134 [0.123, 0.146] & 1000 & 0 \\
        Japanese & 20{,}698 & 11.9 & 2000 & 1.148 [1.090, 1.195] & 1000 & 0.192 [0.178, 0.207] & 2000 & 0.013 [0.002, 0.024] \\
        Thai & 22{,}192 & 12.8 & 2000 & 2.542 [2.492, 2.585] & 1000 & 0.223 [0.212, 0.233] & 1000 & 0 \\
        Arabic & 14{,}477 & 8.4 & 1000 & 1.768 [1.732, 1.802] & 1000 & 0.149 [0.139, 0.158] & 1000 & 0 \\
        Hindi & 19{,}853 & 11.5 & 1000 & 2.042 [2.008, 2.078] & 1000 & 0.139 [0.124, 0.152] & 2000$^\dagger$ & 0.003 [$-$0.009, 0.014] \\
        Korean & 29{,}118 & 16.8 & 512 & 1.474 [1.441, 1.505] & 1000 & 0.189 [0.174, 0.203] & 1000 & 0 \\
        Bengali & 26{,}975 & 15.6 & 2000 & 1.976 [1.944, 2.007] & 1000 & 0.045 [0.037, 0.054] & 1000 & 0 \\
    \bottomrule
\end{tabular}

\end{table}

English gains far less from a swap than the aggregate (\cref{tab:app-readout-swap-languages}). For step-$256$ hidden states, English is best decoded by the step-$512$ readout, and the aggregate's best readout, from step $1000$, raises English NLL by $0.726$ nats $[0.677, 0.779]$ over the native readout. For step-$512$ states both reach their best at the step-$1000$ readout, and from step $1000$ onward no swap improves English NLL beyond its interval of zero.

\begin{figure}[H]
    \centering
    \includegraphics[width=0.85\linewidth]{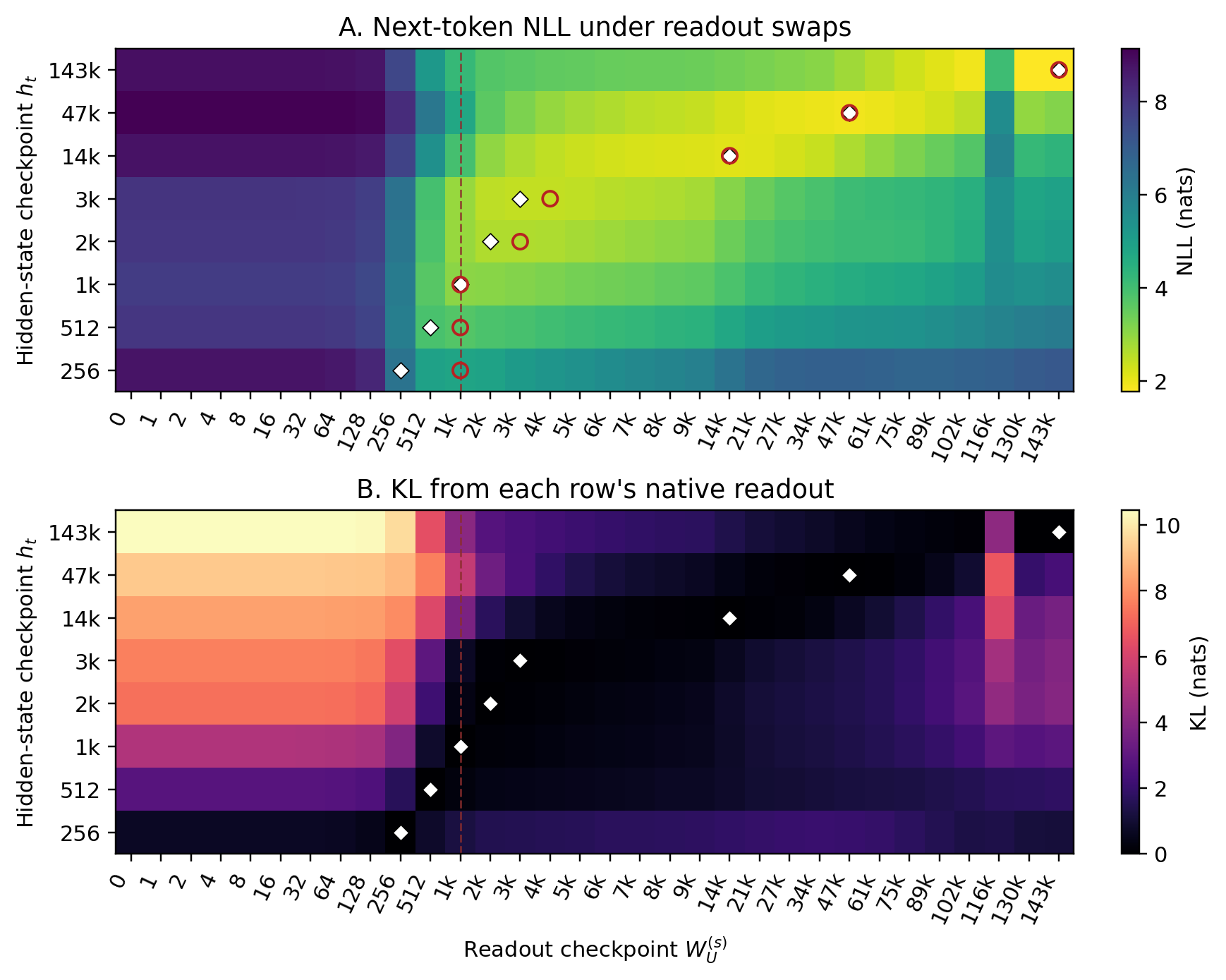}
    \caption{
        In the token-weighted aggregate for Pythia-1B, the step-$1000$ readout is the NLL optimum for hidden states from steps $256$, $512$, and $1000$, while later hidden states prefer nearby or same-checkpoint readouts and become more readout-specific.
        Per language, step $1000$ is the shared optimum only for step-$512$ states (\cref{sec:app-readout-swap-languages}).
        Rows are hidden state checkpoints $t$ (the axis label $h_t$ denotes $h^{(t)}$), and columns are readout checkpoints $s$.
        White diamonds mark same-checkpoint readouts, red circles mark the row-wise minimum-NLL readout, and the dashed vertical line marks the step-$1000$ readout column.
        Panel B shows $\mathrm{KL}(\mathrm{softmax}(h_p^{(t)}\left(\WU^{(s)}\right)^\top)\,\|\,\mathrm{softmax}(h_p^{(t)}\left(\WU^{(t)}\right)^\top))$, so dark cells are close to the same-checkpoint predictive distribution for that hidden-state row.
        The bright column near $116\text{k}$ is the released-checkpoint anomaly of \cref{sec:app-late-checkpoint-anomaly}.
    }
    \label{fig:app-readout-swap-nll-kl-grid}
\end{figure}

The size of the gain depends on the language (\cref{tab:app-readout-swap-languages,fig:app-readout-swap-languages}). For step-$256$ hidden states, the seven languages written in neither Latin nor Cyrillic script supply $97.2\%$ of the aggregate gain at the step-$1000$ readout. With one Latin-script language in the corpus, language, script, training share, and tokenization cannot be separated.

The location of the best readout agrees across languages only in a weaker form (\cref{tab:app-readout-swap-languages}). Every language prefers an intermediate readout for step-$256$ and step-$512$ states, but the step-$256$ optima differ by language, each with bootstrap probability at least $0.997$ of being the minimum, and from step $1000$ onward each language prefers its native readout or the next checkpoint, with gains of at most $0.109$ nats.

\begin{figure}[H]
    \centering
    \includegraphics[width=\linewidth]{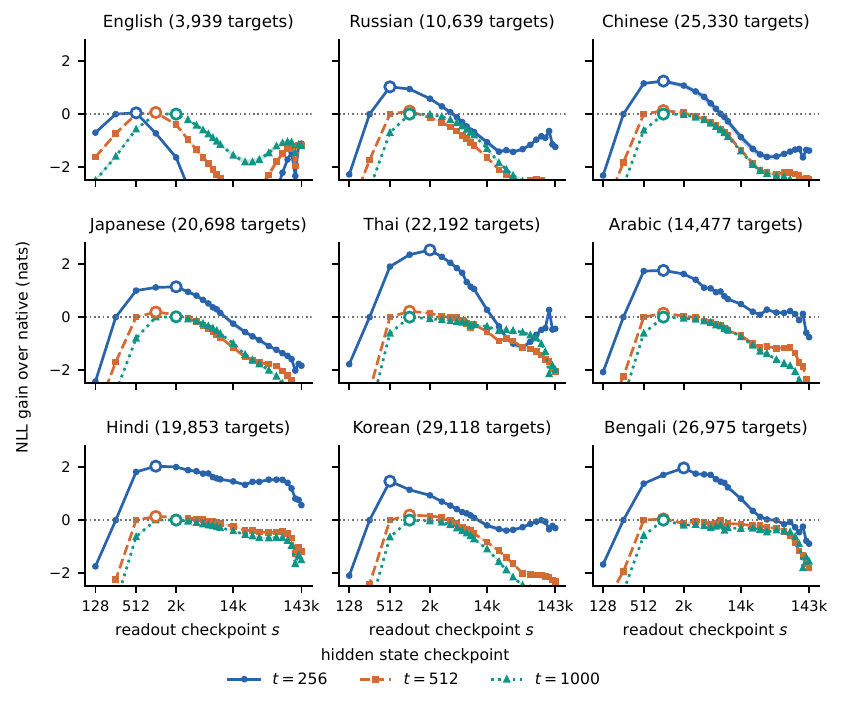}
    \caption{
        In Pythia-1B, the readout swap NLL gain and the best readout checkpoint for step-$256$ hidden states vary by language, while every language prefers an intermediate readout to its native and terminal readouts; for step-$512$ states, all nine languages share the step-$1000$ optimum.
        The gain is native NLL minus swapped NLL, so positive values favor the swapped readout.
        Open circles mark each curve's best readout, bands give $95\%$ block-bootstrap intervals, and values below $-2.5$ nats are clipped.
    }
    \label{fig:app-readout-swap-languages}
\end{figure}

\subsection{Step-1000 Swap Optimum and Late Checkpoint Anomaly}
\label{sec:app-late-checkpoint-anomaly}

Later hidden states become readout-specific (\cref{fig:app-readout-swap-nll-kl-grid}). States from steps $2000$ and $3000$ narrowly prefer the next checkpoint's readout in the aggregate, and in three and five of the nine languages, and late states stay close to their same-checkpoint distribution only under nearby late readouts, whereas early states stay close under many later ones. Terminal $\WU$ is therefore not a neutral logit lens decoder for early checkpoints.

The late spike is the Pythia-1B step-$116\text{k}$ readout anomaly. For fixed step-$47\text{k}$ hidden states, NLL rises from $2.50$ nats under $\WU^{(102\text{k})}$ to $5.55$ under $\WU^{(116\text{k})}$ and returns to $2.95$ under $\WU^{(130\text{k})}$; for step-$143\text{k}$ states, the values are $1.93$, $4.06$, and $1.78$. The spike therefore comes from the step-$116\text{k}$ readout itself, since it appears for fixed hidden states from two checkpoints and disappears under the next readout, and the Pythia-160M replication lacks it, consistent with an anomaly specific to the released Pythia-1B checkpoint. The same checkpoint stands out in reconstruction. A baseline that interpolates each row along the line between its first and last checkpoints drops to EV $0.13$ there, while the dictionary loses only about $0.10$ EV. The held-out heads deviate only at the neighboring steps $113\text{k}$ and $123\text{k}$ (\cref{fig:app-heldout-checkpoints-1b}), and no single token stratum explains the dip in aggregate EV (\cref{sec:app-functional-fidelity}).

\begin{figure}[H]
    \centering
    \includegraphics[width=\linewidth,trim=0 0 0 21,clip]{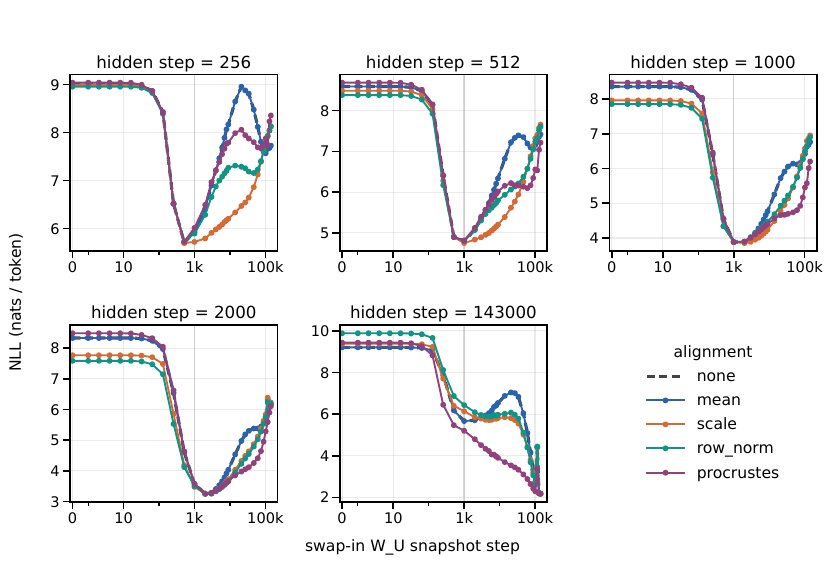}
    \caption{
        In Pythia-1B, each hidden-state checkpoint keeps its minimum-NLL readout across four alignments of the swapped readout (step $512$ for $t=256$ and step $1000$ for $t=512$), except for a one-checkpoint shift under scale alignment for hidden states from checkpoint $1000$ with a small NLL gap.
        Panels are hidden-state checkpoints, the horizontal axis is the readout checkpoint $s$, and the thin vertical line marks $s=1000$.
        ``None'' is the raw swap. Its curve (dashed) lies under the ``mean'' curve, which is identical by construction.
    }
    \label{fig:app-aligned-readout-swap}
\end{figure}

\paragraph{Readout-alignment check.}
The location of the best readout for early hidden states does not depend on the swapped readout's mean, scale, row norms, or orthogonal frame. Scoring each alternate readout raw or after matching the native readout's column mean, centered Frobenius scale, per-row norm, or centered orthogonal Procrustes frame leaves the row-wise optimum unchanged, apart from a one-checkpoint shift under Frobenius scaling for step-$1000$ states with an NLL gap of $0.031$ nats (\cref{fig:app-aligned-readout-swap}).
The check uses the first $16{,}352$ targets of the corpus, mostly English and Russian (about $3.9$k and $10.6$k) with some Chinese, so its NLL values are not comparable with the nine-language aggregate. Its optimum for $t=256$ is the step-$512$ readout rather than the aggregate's step $1000$, consistent with both English and Russian preferring step $512$ there (\cref{tab:app-readout-swap-languages}).

\FloatBarrier

\subsection{Contrastive Task Margin Localization}
\label{sec:app-contrastive-localisation}

The localization of the four task
margins (\cref{sec:sparse-coordinate-localization}, \cref{tab:app-contrastive-localisation-ledger})
uses the released Pythia-1B $\dsae{=}24{,}576$ seed-0 trajectory crosscoder, the norm-filtered localization items of \cref{sec:app-task-construction},
and features selected by attribution to a \emph{task} margin.

\paragraph{Evaluation configuration.}

All interventions on the readout side use one fixed configuration, step-$1000$ hidden states
decoded through the reconstructed terminal readout
$\widehat{\WU}^{(143\text{k})}$. The swap grid above shows that the terminal readout is not a neutral decoder for early hidden states, so the localization results are specific to this configuration. The hidden state projection (below) re-scores
under the native $\WU^{(1000)}$, so its baseline differs from the
reconstruction baseline.

For greater-than completion, IOI, and relational facts, the recon accuracy is at or below chance (\cref{tab:app-contrastive-localisation-ledger}), so the reconstructed terminal readout does not express these margins on step-$1000$ hidden states. Ablation can still lower accuracy when the selected features carry a margin that other features cancel, as for greater-than completion, where it drops accuracy from $0.512$ to $0.084$. Two caveats bound these three rows. First, the reconstruction agrees with the original terminal readout on $0.70$ of held-out top-1 decisions (\cref{sec:app-functional-fidelity}), so a recon value measures the reconstructed readout and may in part reflect reconstruction error. Second, a margin absent at this configuration can be present at the native terminal checkpoint, where the IOI gap closes by the final checkpoint (\cref{tab:app-ae-binary}). The IOI result here is therefore that no compact set carries the margin on early hidden states, and ablation where IOI is expressed remains to be run. We read the ablation and preserve-only deltas against each family's recon baseline, which replaces $0.5$, and the sign-matched control gives the effect of a non-selective edit of the same size.

\paragraph{Terminal readout swap and alignment.}
\label{sec:app-terminal-swap-alignment}
The task sweeps of \cref{sec:readout-reorganization-decodes} use intermediate readouts; the terminal readout behaves differently on early states. On all 320 SVA items of the swap grid at this cell, Pythia-1B's native step-$1000$ readout scores $0.800$, and the terminal readout scores $0.925$ without alignment, $0.925$ after scale alignment, and $0.922$ after row norm alignment. After an orthogonal Procrustes alignment of the terminal readout to the native one, accuracy falls to $0.566$. The terminal readout's gain on these early states therefore depends on its coordinate frame. The intervention run reports the same native and terminal accuracies on the 160 held-out items used in \cref{tab:app-contrastive-localisation-ledger}. \Cref{fig:overview} uses one of these items under the unaligned swap to show that the prediction depends on the readout.

\paragraph{Attribution and interventions.}
Features are attributed to the reconstructed contrastive margin on a random
$50\%$ split of the items (split~A), and all interventions are scored on the
held-out half (split~B), which shares answer tokens with split~A
(\cref{sec:app-task-construction}),
\[
  \operatorname{attr}_f =
  \mathbb{E}_{n\in A}\Bigl[
    \bigl(a_{y^+,f}^{(143\text{k})}-a_{y^-,f}^{(143\text{k})}\bigr)\,
    \bigl(h^{(1000)}_n \cdot D_f^{(143\text{k})}\bigr)\,
    \sigma^{(143\text{k})}
  \Bigr],
\]
where $a_{y^\pm,f}^{(143\text{k})}$ is feature $f$'s reconstructed coefficient
on the answer token row $y^\pm$, $D_f^{(143\text{k})}$ is its decoder row at the terminal checkpoint, and $\sigma^{(143\text{k})}$ is the terminal normalization scale of \cref{sec:app-method-normalization} (written $s^{(t)}$ there), so that $\sigma^{(143\text{k})}a_{v,f}^{(143\text{k})}D_f^{(143\text{k})}$ is the feature's contribution to row $v$ in the original readout space. The \emph{ablate} intervention removes the
reconstructed contributions of the top eight positive attribution features,
$\widehat{\WU}\leftarrow\widehat{\WU}-\sum_{f\in S}
\sigma^{(143\text{k})}a_{:,f}^{(143\text{k})}D_f^{(143\text{k})}$; \emph{preserve-only} keeps only
those eight; and the \emph{hidden state projection} sets $h^{(1000)}\leftarrow(I-QQ^\top)h^{(1000)}$, where the columns of $Q$ are an orthonormal basis, computed by QR, of the eight decoder
directions $\{D_f^{(143\text{k})}\}_{f\in S}$,
and re-scores under the model's native readout $\WU^{(1000)}$. Each intervention is
compared against the same edit applied to eight random features matched on decoder norm, firing rate, and attribution sign.
Each control feature is the nearest unused neighbor of one selected feature in the plane of log decoder norm and log firing rate, the fraction of vocabulary rows on which the feature fires at the terminal snapshot. The sign-matched control draws only from features with positive attribution, falling back to all features if too few qualify.
Random features matched only on decoder norm and firing rate leave accuracy at the recon value in $11$ of the $12$ stored family and model rows of this run, as expected if they are inactive on the answer rows, since an ablation changes only the rows where a feature is active; the table therefore reports the sign-matched control.

\begin{table}[H]
  \centering
  \footnotesize
  \caption{Per-family top eight intervention accuracies at the fixed
  evaluation cell (Pythia-1B, held-out half; relational facts, scored on $13$
  items, are exploratory). \emph{orig.} is the original terminal readout and \emph{recon} the full
  reconstructed terminal
  readout; \emph{ablate} and \emph{preserve-only} act on the top eight
  positive attribution features, and \emph{sign} is the same edit applied to eight random features matched on decoder norm, firing rate, and attribution sign. The \emph{hidden projection} columns score step-$1000$ states under the native $\WU^{(1000)}$ before (\emph{native}) and after (\emph{proj.}) the top eight decoder directions are projected out. $^\ast$Scored on all localization items ($2{,}000$, $462$, and $26$), not the held-out half.}
  \label{tab:app-contrastive-localisation-ledger}
  \setlength{\tabcolsep}{4pt}
  \begin{tabular}{@{}lcccccccc@{}}
    \toprule
    & & & \multicolumn{2}{c}{ablate} & \multicolumn{2}{c}{preserve-only} & \multicolumn{2}{c}{hidden projection} \\
    \cmidrule(lr){4-5}\cmidrule(lr){6-7}\cmidrule(lr){8-9}
    Family & orig. & recon & top-8 & sign & top-8 & sign & native & proj. \\
    \midrule
    Subject--verb agreement     & 0.925 & 0.900 & 0.487 & 0.931 & 0.938 & 0.463 & 0.781 & 0.500 \\
    Greater-than completion     & 0.529$^\ast$ & 0.512 & 0.084 & 0.500 & 0.894 & 0.507 & 0.601 & 0.497 \\
    Indirect object id.\ (IOI)  & 0.455$^\ast$ & 0.429 & 0.394 & 0.424 & 0.481 & 0.420 & 0.377 & 0.338 \\
    Relational facts            & 0.269$^\ast$ & 0.231 & 0.154 & 0.231 & 0.923 & 0.308 & 0.308 & 0.154 \\
    \bottomrule
  \end{tabular}
\end{table}

\paragraph{Reading the table.}
Subject--verb agreement and greater-than completion localize, with ablation far below the sign-matched control and preserve-only high on the eight directions alone, and relational facts show a strong but exploratory preserve-only effect on $12$ of $13$ scoring items (\cref{tab:app-contrastive-localisation-ledger}). Indirect object identification shows no selective positive-margin effect from a compact set at this configuration. Its ablation stays close to the sign-matched control, positive attribution is spread across $4{,}387$ of $24{,}576$ features, and projecting the eight directions out of $h^{(1000)}$ shifts the $y^+$ and $y^-$ logits almost equally ($-0.44$ and $-0.45$), consistent with generic token suppression at the top eight budget.

Projecting the eight directions out of $h^{(1000)}$ removes at most $1\%$ of the hidden state norm, yet it costs subject--verb agreement far more accuracy than greater-than completion. Because the projection is scored under the unmodified native readout $\WU^{(1000)}$, the agreement effect depends on hidden state directions spanned by the eight decoders, which the native readout uses to separate $y^+$ from $y^-$, so the change in the readout between steps $1000$ and $143\text{k}$ is not its only source. The projection run's native SVA baseline of $0.781$ differs from the $0.800$ of the swap grid and the intervention run.

\FloatBarrier

\section{Recipe Control: Readout Learning Rate and Developmental Timing}
\label{sec:app-recipe-control}

This appendix details the controlled training experiment summarized in \cref{sec:recipe-control}.

\subsection{Setup and Conditions}
\label{sec:app-recipe-control-setup}

The control trains four Pythia-architecture models of $31$M parameters that share initialization seed and sequential data order, on $10$B tokens.
The conditions differ only in one of two recipe factors.
Three vary the learning rate multiplier applied to the output readout $\WU$ ($\texttt{embed\_out.weight}$): \texttt{wu\_lr\_0p25} ($0.25\times$), \texttt{baseline} ($1\times$), and \texttt{wu\_lr\_4x} ($4\times$).
The fourth, \texttt{warmup\_short}, shortens the learning rate warmup (from the baseline $1430$ steps to $715$) at the baseline $1\times$ readout learning rate.
Each condition has one pretraining run and three crosscoder fits (seeds $0$, $1$, and $2$). Crosscoder tables and figures show seed $0$, and the timing statistics are reported for all three fits.
A fifth run with a fourfold longer warmup ($5720$ steps) was also trained. Its warmup exceeds the $4769$-step budget, so it never reaches the peak learning rate, and we leave it out of the comparison; its checkpoints are released with the others.
The models use the Pythia architecture at $\dmodel=256$ with six layers and the Pythia tokenizer. Training follows Pythia's optimizer recipe: AdamW \citep{loshchilov2019decoupled} with $(\beta_1,\beta_2)=(0.9,0.95)$, $\epsilon=10^{-8}$, weight decay $0.1$ (excluding LayerNorm parameters and biases), gradient clipping at $1.0$, fp16 mixed precision, and a global batch of $1024$ sequences of $2048$ tokens ($2.1$M tokens per step). The learning rate warms up linearly to a peak of $10^{-3}$ over $1430$ steps and follows Pythia's cosine schedule over $143$k steps, stopped at the $10$B-token budget ($4769$ steps). At that point the learning rate is still within $0.2\%$ of its peak, so the reported final losses precede any decay of the learning rate. The data are the first $10$B tokens of Pythia's preshuffled, pretokenized Pile \citep{gao2020pile,biderman2023pythia} (\texttt{EleutherAI/pile-standard-pythia-preshuffled}), in $2048$-token blocks read once, in Pythia's released order, by all conditions; the last $256$ blocks (about $0.5$M tokens) are held out for validation loss.
The control keeps $K=16$ checkpoints at steps $\{0,1,2,4,8,16,32,64,128,256,512,1024,2048,2479,4096,4769\}$, that is, step $0$, the powers of two up to $4096$, the final step, and one checkpoint that the trainer saved when the first $11$-hour job slot ended. That slot checkpoint falls at step $2479$ for \texttt{baseline} and at steps $2464$ to $2481$ across the four conditions, and we write it as $2479$ throughout.
The hidden state views of \cref{sec:app-recipe-control-concordance} use held-out Pile tokens that directly follow the $10$B-token training prefix in the same preshuffled order. The readout swap scores all $118{,}726$ next-token positions of a $120$k-token slice, and the frequency class probe uses $20{,}000$ positions sampled (seed $0$) from a $400$k-token slice.

Every $\WU$ trajectory crosscoder is fitted using the main Pythia-160M recipe of \cref{tab:repro-dictionary-hparams} (mapped to $\dmodel=256$): JumpReLU, $\dsae=8192$, $\lambda_1=0.3$, learning rate $5\times10^{-5}$, centered and scaled inputs, $300$ epochs.
The four seed-$0$ crosscoders reconstruct their trajectories with comparable explained variance and have no dead features (\cref{tab:app-recipe-control-summary}).
Decoder norm trajectories $\rho_f(t)$, the alive threshold $\rho_f(t)\ge0.5$, peak step, and the rule-based profile assignment are computed exactly as in \cref{sec:app-lifecycle-diagnostics}.

\subsection{Lifecycle Response to the Readout Learning Rate}
\label{sec:app-recipe-control-doseresponse}

A faster readout shifts the per-feature trajectories from flat, persistent curves toward curves that rise and fall with earlier peaks, so the profile shares move from mostly persistent at $0.25\times$ to mostly early-decaying at $4\times$ (\cref{tab:app-recipe-control-summary}).

\begin{table}[H]
    \centering
    \small
    \setlength{\tabcolsep}{3.5pt}
    \renewcommand{\arraystretch}{1.12}
    \caption{
        Per-condition crosscoder quality and lifecycle summary for the recipe control ($\dsae=8192$, $K=16$, $8192$ active features each, seed $0$).
        EV is explained variance; $L_0$ is mean active features per row; ``val.\ loss'' is the condition's final validation loss in nats; the profile columns are fractions of active features; ``med.\ peak'' is the median peak step; ``turnover step'' is the peak turnover step.
        The validation loss at the peak turnover step is $3.72$ nats for $0.25\times$ (step $2048$) and $5.13$ nats for $4\times$ (step $512$). At $1\times$, turnover at steps $1024$ and $2048$ differs by less than $0.002$ in every fit, and seed $1$ selects step $2048$.
        The two timing statistics are the primary measures; the rule-based profile fractions are discrete summaries sensitive to the profile rule boundaries (\cref{sec:app-recipe-control-warmup}).
    }
    \label{tab:app-recipe-control-summary}
    \begin{tabular}{lccccccccc}
        \toprule
        Condition & EV & $L_0$ & val.\ loss & persist. & early-dec. & late-em. & mixed & med.\ peak & turnover step \\
        \midrule
        \texttt{wu\_lr\_0p25}  & 0.905 & 193.9 & 3.405 & 0.875 & 0.000 & 0.093 & 0.032 & 128 & 2048 \\
        \texttt{baseline}      & 0.891 & 177.2 & 3.381 & 0.403 & 0.219 & 0.125 & 0.252 & 64  & 1024 \\
        \texttt{warmup\_short} & 0.890 & 175.5 & 3.350 & 0.011 & 0.442 & 0.153 & 0.394 & 64  & 1024 \\
        \texttt{wu\_lr\_4x}    & 0.874 & 168.5 & 3.369 & 0.003 & 0.675 & 0.131 & 0.191 & 32  & 512  \\
        \bottomrule
    \end{tabular}
\end{table}

\begin{figure}[H]
    \centering
    \includegraphics[width=\linewidth]{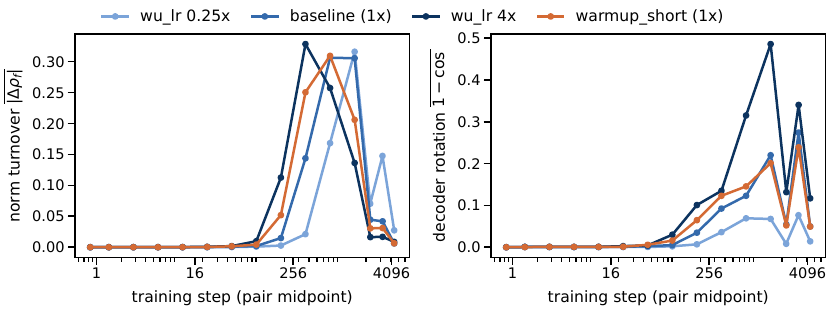}
    \caption{
        Higher readout learning rate shifts the peak of decoder norm turnover (left) earlier, and \texttt{warmup\_short} tracks \texttt{baseline}.
        Decoder rotation (right) peaks at $1024{\to}2048$ only for $4\times$. Its late spike, and the late rise in norm turnover for $0.25\times$, come from the pair from the slot checkpoint near step $2479$ to step $4096$, which spans about $1{,}600$ steps, the longest interval of the schedule, against fewer than $450$ steps for the pair before it.
        Each point averages the change over active features of the seed-$0$ crosscoder for one consecutive checkpoint pair, plotted at the pair midpoint.
    }
    \label{fig:app-recipe-control-reorg}
\end{figure}

Decoder norm turnover $\overline{|\Delta\rho_f|}$ peaks in a localized window that moves earlier as the readout learning rate rises, whereas decoder rotation $\overline{1-\cos(W_{\mathrm{dec},f}^{(t)},W_{\mathrm{dec},f}^{(t+1)})}$ peaks late in three of the four conditions, at the long interval after the slot checkpoint, so the timing claim rests on norm turnover (\cref{fig:app-recipe-control-reorg}). \Cref{fig:main-recipe-dose-response} reports the peak turnover step and the median peak step for all three fits.

\subsection{Warmup Length Leaves the Crosscoder Timing Statistics Unchanged}
\label{sec:app-recipe-control-warmup}

Halving the warmup ($715$ steps against $1430$) at the baseline multiplier raises the effective readout learning rate early in training, and both schedules reach the same peak rate after step $1430$.
The \texttt{warmup\_short} run matches \texttt{baseline} on both crosscoder timing statistics and on maximum turnover (\cref{tab:app-recipe-control-summary,fig:app-recipe-control-reorg}), although its validation loss at the peak turnover step is $0.52$ nats lower.
Its discrete profile fractions differ even though its peak-step distribution does not, and because these classifications are sensitive to the thresholds for near-flat trajectories in \cref{tab:lifecycle-profile-rules}, we read them separately from the timing statistics.
On the hidden state views of \cref{sec:app-recipe-control-concordance}, it lies near \texttt{baseline} with a small shift toward the faster readout, with a smaller early frequency class gap and a swap optimum one checkpoint later than native at $t=256$ and $512$ whose gains stay within $0.05$ nats, against $0.91$ nats in the $0.25\times$ run.
Warmup length therefore leaves the crosscoder timing unchanged on this grid and acts on the frequency class gap as a small partial step along the multiplier axis.

\subsection{Agreement Across Views on the Same Checkpoints}
\label{sec:app-recipe-control-concordance}

We compute geometry, lifecycle, swap, and probe diagnostics on the same four runs, and they agree along the readout learning rate axis.

The aggregate $\WU$ geometry shifts monotonically with the readout learning rate, with final mean row norms of $1.85$ at $4\times$ and $0.93$ at $0.25\times$ and the final LayerNorm gain moving in the opposite direction (\cref{fig:app-recipe-control-geometry}).

\begin{figure}[H]
    \centering
    \includegraphics[width=0.9\linewidth]{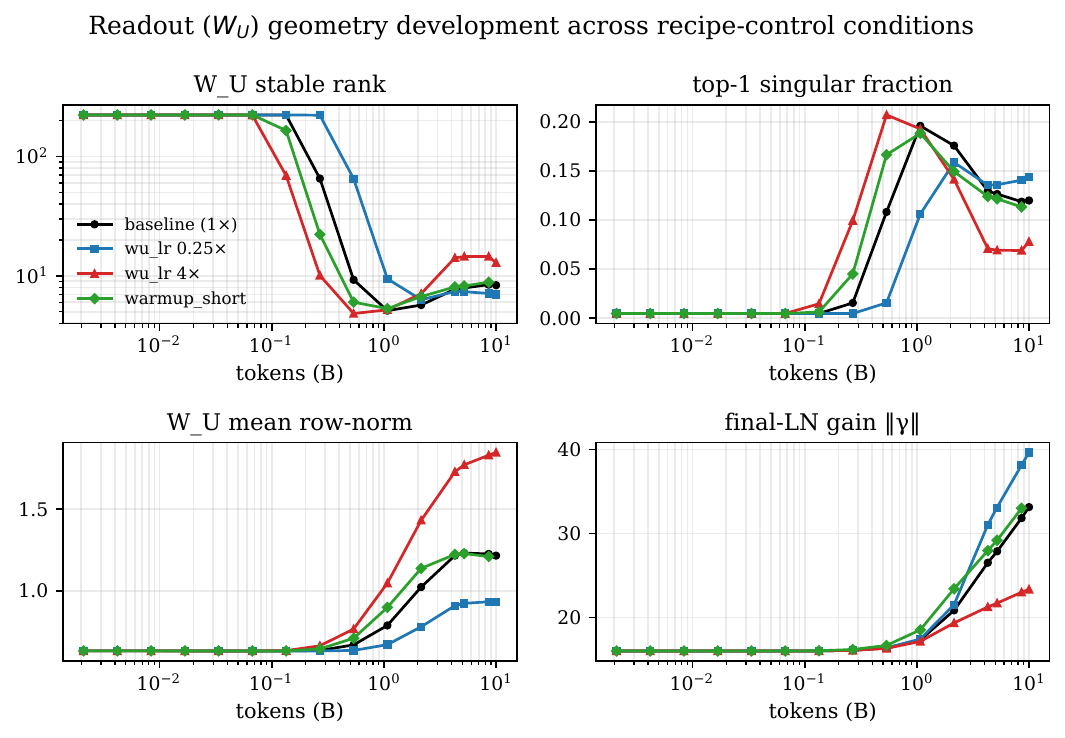}
    \caption{
        By the end of training, higher readout learning rate gives larger $\WU$ row norms, a less concentrated spectrum, and a smaller final LayerNorm gain, and \texttt{warmup\_short} tracks \texttt{baseline}.
        The stable rank of $\WU$ is $\|\WU\|_F^2/\sigma_1^2$, where $\sigma_1$ is its largest singular value.
        The top-1 singular fraction is $\sigma_1^2/\|\WU\|_F^2$, the share of squared Frobenius norm in the top singular direction.
        The final LayerNorm gain is the $\ell_2$ norm of its gain vector $\gamma$, and the \texttt{warmup\_short} curves end at step $4096$ because its final geometry measurement is missing.
    }
    \label{fig:app-recipe-control-geometry}
\end{figure}

The readout swap of \cref{sec:app-readout-swap}, applied per condition, follows the same axis (\cref{fig:app-recipe-control-basin}). At $4\times$ each hidden state is best decoded by its own readout, whereas at $0.25\times$ early states are best decoded by a later readout, and for $t=256$ the step-$512$ readout gives a gain of $0.91$ nats that stays positive under the mean, scale, row norm, and Procrustes alignments ($0.91$ to $1.07$ nats).

The probe-versus-native comparison of \cref{sec:app-availability-expression}, applied per condition, uses the frequency class probe of \cref{sec:app-evaluation-details} in place of a task label, so it measures availability of this coarse next-token property rather than of any task in \cref{tab:task-protocol}.
The early frequency class gap, probe minus native accuracy, is ordered $0.25\times > 1\times > 4\times$ at $t=256$ and $512$, and \texttt{warmup\_short} lies between \texttt{baseline} and $4\times$ at $t=256$ and below $4\times$ at $t=512$ (\cref{fig:app-recipe-control-explag}). Because the native readout exceeds the probe in every condition from $t=2048$ on, the early gap is read for its ordering across conditions rather than for its absolute size.

\begin{figure}[H]
    \centering
    \includegraphics[width=0.68\linewidth,trim=0 0 0 32,clip]{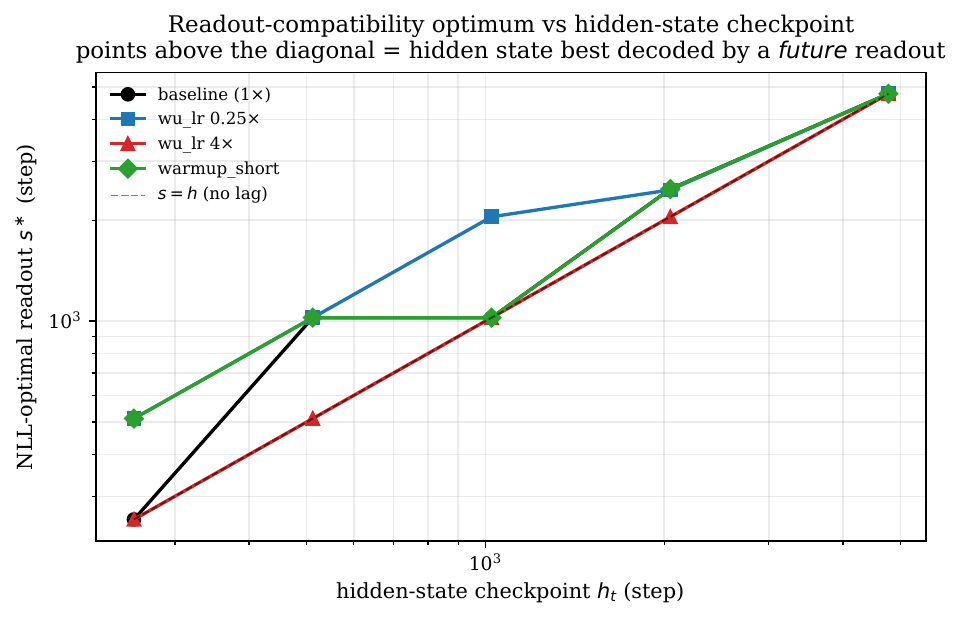}
    \caption{
        The swap optimum $s^\ast$, the readout checkpoint with the lowest NLL for hidden states from checkpoint $t$, tracks the readout learning rate. Points above the diagonal are hidden states best decoded by a later readout. \texttt{wu\_lr\_4x} follows the diagonal and \texttt{wu\_lr\_0p25} lies above it, while for \texttt{baseline} and \texttt{warmup\_short} the gain over the native readout is at most $0.05$ nats.
    }
    \label{fig:app-recipe-control-basin}
\end{figure}

\begin{figure}[H]
    \centering
    \includegraphics[width=\linewidth]{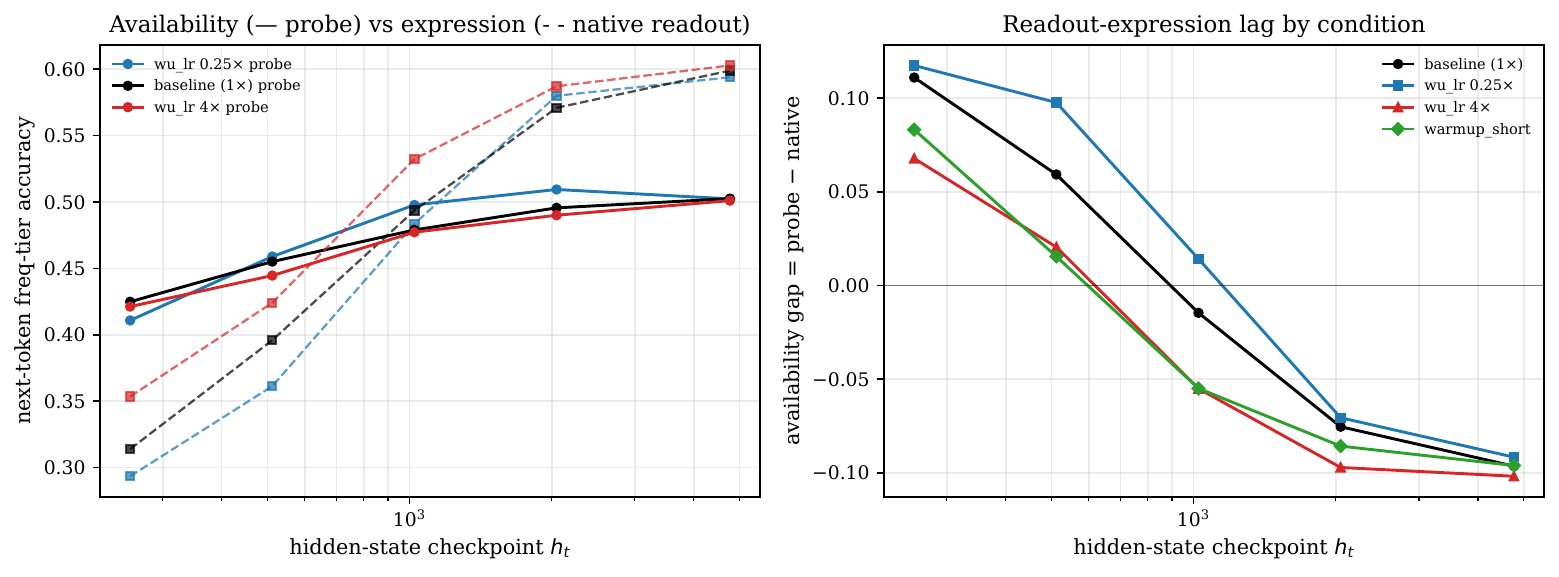}
    \caption{
        The frequency class gap (probe accuracy minus native accuracy) in the early window is ordered $0.25\times > 1\times > 4\times$ and closes by step $1024$ in every run except $0.25\times$, which closes by step $2048$.
        Accuracy is on four next-token frequency classes of equal corpus occurrence mass.
        From step $2048$ on the native readout exceeds the probe in every condition, so the probe is not a ceiling late in training.
    }
    \label{fig:app-recipe-control-explag}
\end{figure}

\subsection{Early Loss Differences and Similar Final Loss}
\label{sec:app-recipe-control-gauge}
The early loss comparison of \cref{sec:recipe-control} is made at step $128$ ($268{,}435{,}456$ training tokens), where the validation loss of the $4\times$ run is $1.202$ nats below baseline and $2.079$ nats below $0.25\times$, and \cref{tab:app-recipe-control-summary} gives the final losses.

Similar final losses need not imply identical models. Because of the scalar symmetry between the final normalization and the readout (\cref{sec:app-method-normalization}), readout row norms cannot be interpreted independently of the final normalization. The symmetry does not establish that the independently trained models differ only along it.

\section{Related Work, Limitations, and Broader Impact}
\label{sec:app-evidence-scope}

\subsection{Extended Related Work}
\label{sec:app-extended-related-work}

\paragraph{Encoded but unexpressed information.}
Probes test what is decodable from representations \citep{alain2016understanding,conneau2018cram,hewitt2019structural,tenney2019bert}, and control tasks separate what the probe itself learns \citep{hewitt2019designing,belinkov2022probing}.
Trained models can encode answers that their outputs do not express \citep{burns2023discovering,orgad2025llms,gekhman2025inside}, and training can make progress that loss and error metrics do not reveal \citep{barak2022hidden}.
We follow this gap across pretraining checkpoints and use readout swaps on fixed hidden states to attribute its expression side to the readout.

\paragraph{Feature correspondence.}
Within a fixed tokenizer, token identity gives our dictionary its row correspondence, but it does not remove hidden space reparameterizations or establish semantic identity across independently trained models \citep{ainsworth2023git}, and shared indices do not establish that the features are canonical units \citep{leask2025canonical,minder2025sparsity}.

\begin{table}[H]
  \centering
  \footnotesize
  \setlength{\tabcolsep}{3pt}
  \renewcommand{\arraystretch}{1.12}
  \caption{Analysis families and the comparisons they support. Temporal correspondence describes the unit compared across checkpoints. The last column distinguishes dictionary fitting from the interventions used to assess behavior.}
  \label{tab:method-comparison}
  \begin{tabular}{@{}>{\raggedright\arraybackslash}p{1.85cm}>{\raggedright\arraybackslash}p{2.1cm}>{\raggedright\arraybackslash}p{2.15cm}>{\raggedright\arraybackslash}p{2.9cm}>{\raggedright\arraybackslash}p{3.65cm}@{}}
    \toprule
    Family & Analyzed object & Input data & Temporal correspondence & What the comparison isolates \\
    \midrule
    Lens methods & Hidden states in vocabulary space & Text and forward passes & Vocabulary coordinates; native or learned readout & Predictions under the chosen readout \\
    Vocabulary geometry & Input/output embedding rows & Weights; lexical annotations & Token identities or geometric statistics & Changes in geometry and lexical association \\
    Independent SAEs & Activations per checkpoint & Activation corpus & Separate indices; post-hoc matching & Features within each fit \\
    Activation crosscoders & Activations across checkpoints & Activation corpus & Shared sparse indices & Representation change; feature use via interventions \\
    Parameter methods & Decomposed or sparse weights & Weights and inputs & Typically one model state & Parameter components and circuits \\
    This work & Readout row trajectories & Weights; text for tests & Token identities and shared sparse indices & Readout change; swaps and ablations at fixed states \\
    \bottomrule
  \end{tabular}
\end{table}

\subsection{Extended Limitations}
\label{sec:app-result-scope}
Beyond those of \cref{sec:limitations}, four limits apply.
The checkpoint schedules bound temporal resolution, since they sample Pythia densely only before step 10k and each 31M run at 16 checkpoints.
Profile shares and the median feature peak step depend on dictionary width and sparsity (\cref{sec:app-lifecycle-stability}), so we read the shares qualitatively and compare timelines only at matched settings, whereas the reorganization window holds in 20 of 21 fits.
The probes are linear and read only the final hidden state at the answer position, so they can miss distinctions decodable only nonlinearly or at other positions and layers.
The comparisons have narrow scope. OLMo-2-7B is the only cross-family check, the sparse interventions use one configuration (\cref{sec:sparse-coordinate-localization}), and tokenization, preprocessing, and optimization shape all comparisons.

\subsection{Implications}
\label{sec:app-implications}

\paragraph{Open questions.}
It remains open how task directions become coupled to the readout late in training, since relative clause agreement and numeric comparison keep gaps that no evaluated readout closes, and how hidden state and readout updates are coordinated, since a readout swap isolates only one of the three terms of \cref{eq:logit-change}.

\paragraph{Control through the readout.}
The directions that carry the agreement margin at one configuration (\cref{sec:sparse-coordinate-localization}) make the readout a compact intervention site, as steering generation through linear transformations of the output embeddings suggests \citep{han2024word}, and the side effects and persistence of such edits remain to be measured (\cref{sec:app-social-impact}).

\subsection{Social Impact}
\label{sec:app-social-impact}

The work deploys no language model and does not propose a validated safety evaluation procedure.
The main positive impact is methodological transparency, because treating the learned readout as part of the measured system makes developmental claims in logit space easier to audit, and the analysis code is released to support auditing of the reported checkpoint comparisons.
The main negative-use concern is more targeted readout manipulation, since sparse readout features can identify token-family directions that affect logits.

\endgroup

\end{document}